\documentclass[letterpaper, 10 pt, journal, twoside]{IEEEtran}

\usepackage{url}
\usepackage{textcomp,gensymb}
\usepackage{cite}
\usepackage{amsmath,amssymb,amsfonts}
\usepackage{graphicx}
\usepackage{textcomp}
\usepackage{xcolor}
\usepackage{multirow}
\usepackage{booktabs}
\def\BibTeX{{\rm B\kern-.05em{\sc i\kern-.025em b}\kern-.08em
    T\kern-.1667em\lower.7ex\hbox{E}\kern-.125emX}}
\usepackage{xcolor}
\usepackage{algorithm}
\usepackage{algpseudocode}
\DeclareMathSymbol{\shortminus}{\mathbin}{AMSa}{"39}
\usepackage{hhline,caption}
\title{
Mapping Mid-air Haptics\\ with a Low-cost Tactile Robot
}

\begin{document}

\author{Noor Alakhawand$^{1}$, William Frier$^{2}$, and Nathan F. Lepora$^{1}$
\thanks{%
%
This work was supported in part by an \mbox{EPSRC} CASE award to Noor Alakhawand sponsored by Ultraleap and in part by an award from the Leverhulme Trust on ‘A biomimetic forebrain for robot touch’ (RL-2016-39).}
\thanks{$^{1}$Noor Alakhawand and Nathan F. Lepora are with the Department of Engineering Mathematics and Bristol Robotics Laboratory, University of Bristol, Bristol BS8 1UB, U.K. (email: noor.alakhawand@bristol.ac.uk; n.lepora@bristol.ac.uk).}%
\thanks{$^{2}$William Frier is with Ultraleap Ltd., Bristol BS2 0EL, U.K. (email: william.frier@ultraleap.com).}%
}


\maketitle

\begin{abstract}
Mid-air haptics create a new mode of feedback to allow people to feel tactile sensations in the air. Ultrasonic arrays focus acoustic radiation pressure in space, to induce tactile sensation from the resulting skin deflection.
In this work, we present a low-cost tactile robot to test mid-air haptics. By combining a desktop robot arm with a 3D-printed biomimetic tactile sensor, we developed a system that can sense, map, and visualize mid-air haptic sensations created by an ultrasonic transducer array.
We evaluate our tactile robot by testing it on a variety of mid-air haptic stimuli, including unmodulated and modulated focal points that create a range of haptic shapes.
We compare the mapping of the stimuli to another method used to test mid-air haptics: Laser Doppler Vibrometry, highlighting the advantages of the tactile robot including far lower cost, a small lightweight form-factor, and ease-of-use. Overall, these findings indicate our method has multiple benefits for sensing mid-air haptics and opens up new possibilities for expanding the testing to better emulate human haptic perception. 
\end{abstract}


\section{Introduction}
\IEEEPARstart{M}{id-air} haptics provide a new mode of sensory feedback for humans, creating ``virtual touch” that allows people to feel tactile sensations in the space above a transducer. The primary method to produce mid-air haptics is with ultrasonic arrays, which focus acoustic radiation pressure to induce tactile sensation by microscopically deflecting the skin~\cite{carterUltraHapticsMultipointMidair2013}.     
This technology is opening up new possibilities for contactless interactions: since there are no wearables needed, distinct from other haptic feedback devices, mid-air haptics can facilitate spontaneous interaction with haptic displays. Removing the need for physical contact can also be beneficial in situations needing sterile conditions, such as medical applications.
Mid-air haptics can also provide an extra mode of interaction in mixed-reality interfaces, additional to the commonly-used visual and auditory forms of feedback \cite{rakkolainenSurveyMidAirUltrasound2021}. 

\begin{figure}[t]
	\centering
	\begin{tabular}[b] {p{0.46\linewidth}p{0.46\linewidth}}
		\includegraphics[width=0.45\columnwidth,trim={0 0 0 0},clip]{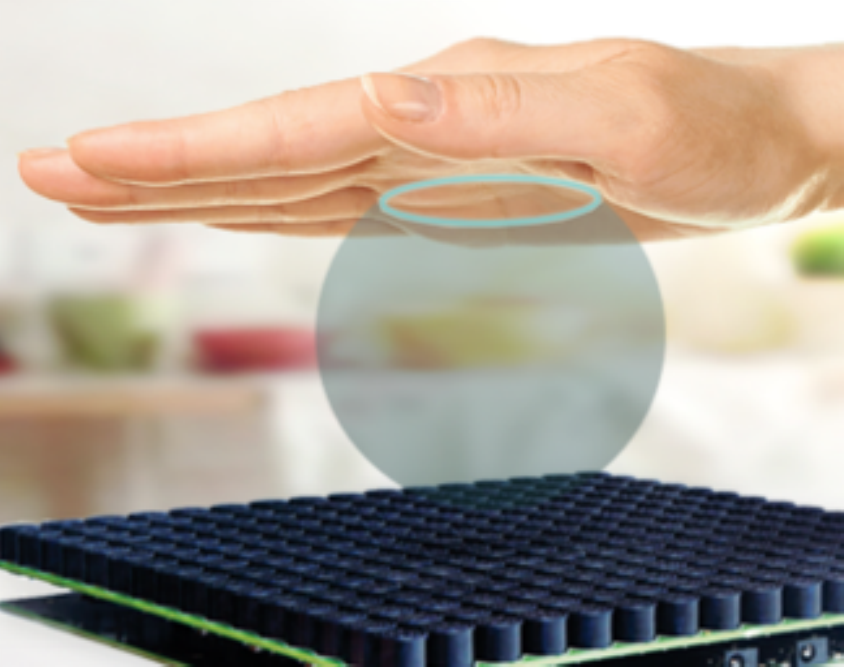} &
    	\includegraphics[width=0.45\columnwidth,trim={0 20 0 60},clip]{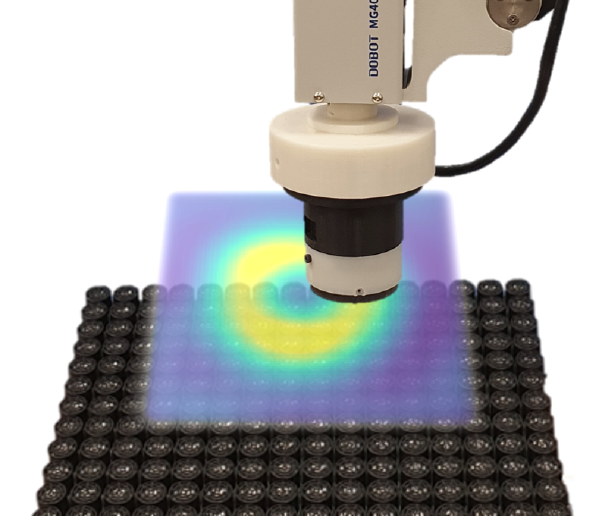} \\
		{\bf (a) Mid-air haptics as sensed by a hand} & {\bf (b) Mid-air haptics as sensed by the tactile robot}\\
	\end{tabular}
	\caption{Mid-air haptics are generated by the ultrasonic array and can be felt with a human hand (a). Our robot uses a tactile sensor to test the sensations in place of a hand (b).}
	\label{fig_intro}

\end{figure}

When developing haptic feedback using ultrasonic devices, one needs to evaluate whether the desired sensations are produced as intended. Therefore, there is a need to understand how focal points of pressure interact with compliant skin to cause it to deform.
The focal points generated by the ultrasonic array can be modulated in intensity and position to create different sensations. For example, by moving a focal point along the path of a shape, such as a circle, the illusion of a static shape is produced that can be felt by a human hand~(Fig. \ref{fig_intro}). Alternatively, by modulating the intensity of the focal point, a pulsing sensation can be created. These methods can be combined to generate a variety of sensations with both changing intensities and positions.These modulated focal points deform the viscoelastic skin in a non-linear process to which a variety of modelling and experimental methods can be applied to test the sensation being produced. 

In the present study, we focus on using tactile sensing to measure the low-frequency deformation caused by the sensations. Specifically, we present a low-cost tactile robotic platform to test mid-air haptics. We combine a lightweight, desktop robot arm with a 3D-printed soft biomimetic tactile sensor~\cite{leporaSoftBiomimeticOptical2021,ward-cherrierTacTipFamilySoft2018}, to develop a system that can map mid-air haptic sensations. This tactile robot is applied to mid-air haptics created by an array of ultrasonic transducers. Our contributions are as follows:\\
\noindent 1) We introduce a low-cost desktop robotic system ($\sim$£2,000) utilizing biomimetic tactile sensing to map and test the performance of ultrasonic mid-air haptics.\\
\noindent 2) We demonstrate that we can accurately map the sensations produced by the ultrasonic array with comparable results to a higher-cost method.\\
\noindent 3) We show that we can add real-time control for more efficient exploration of mid-air haptic shapes.

This paper is structured as follows. First, we present the components of the testing platform, explaining the choice of the robot arm and the biomimetic tactile sensor that are used to test mid-air haptics. We then provide details of methodology we have developed for \textit{systematic mapping} of mid-air haptics. This involves using the robot arm to move the sensor in a grid pattern over the ultrasonic array, combining the data gathered into a map of the sensation. We map several distinct mid-air haptic stimuli and also provide a comparison with other mapping methods. Our methodology is then expanded by adding intelligence into the robot movements to instantiate \textit{autonomous haptic exploration} that uses the robot arm to move the tactile sensor to explore the space and decide where to move next based on what it has felt. This can provide a more efficient method for data collection and make the process more similar to how people use ``exploratory procedures" to interact dynamically with tactile stimuli when finding an object’s shape~\cite{ledermanHandMovementsWindow1987}.

\section{Background}

\begin{figure*}[ht]
	\centering
	\begin{tabular}[b] {ccc}
		\includegraphics[width=0.4\linewidth,trim={0 0 0 0},clip]{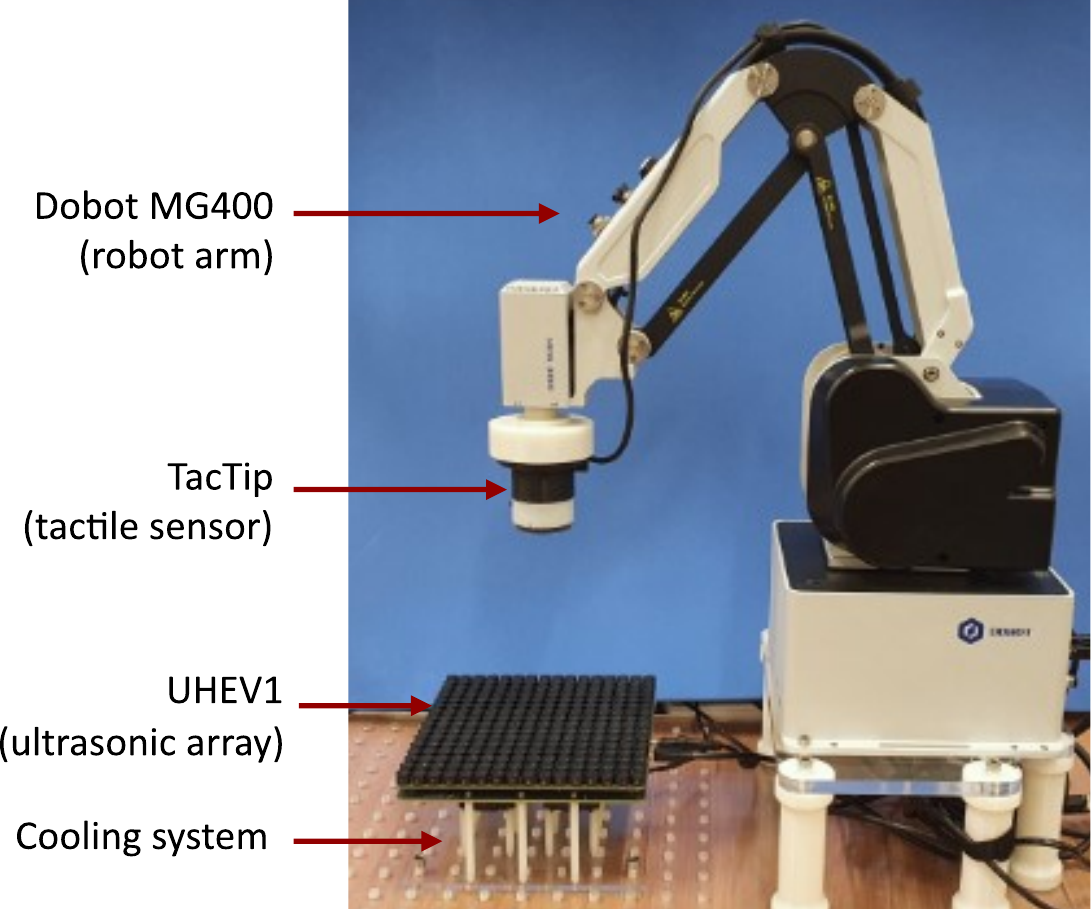} &
    	\includegraphics[width=0.22\linewidth,trim={0 0 10 0},clip]{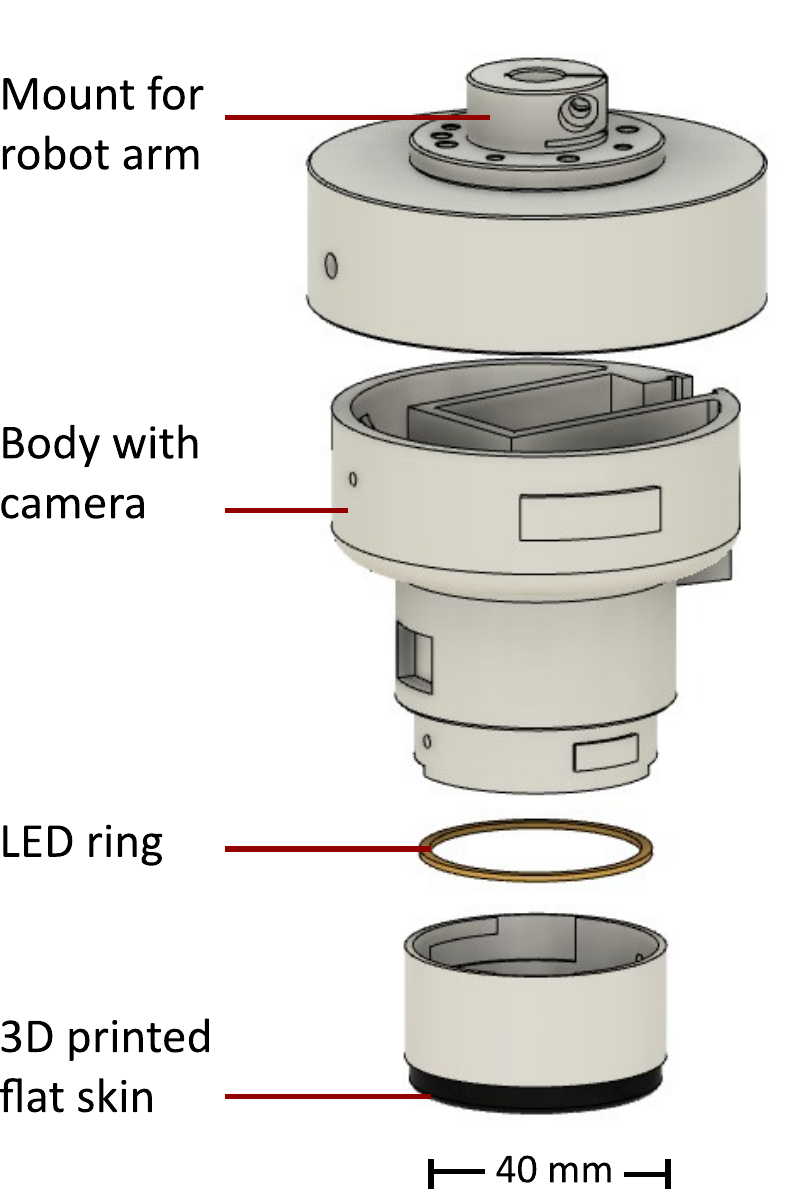} &
    	\includegraphics[width=0.27\linewidth,trim={0 0 0 0},clip]{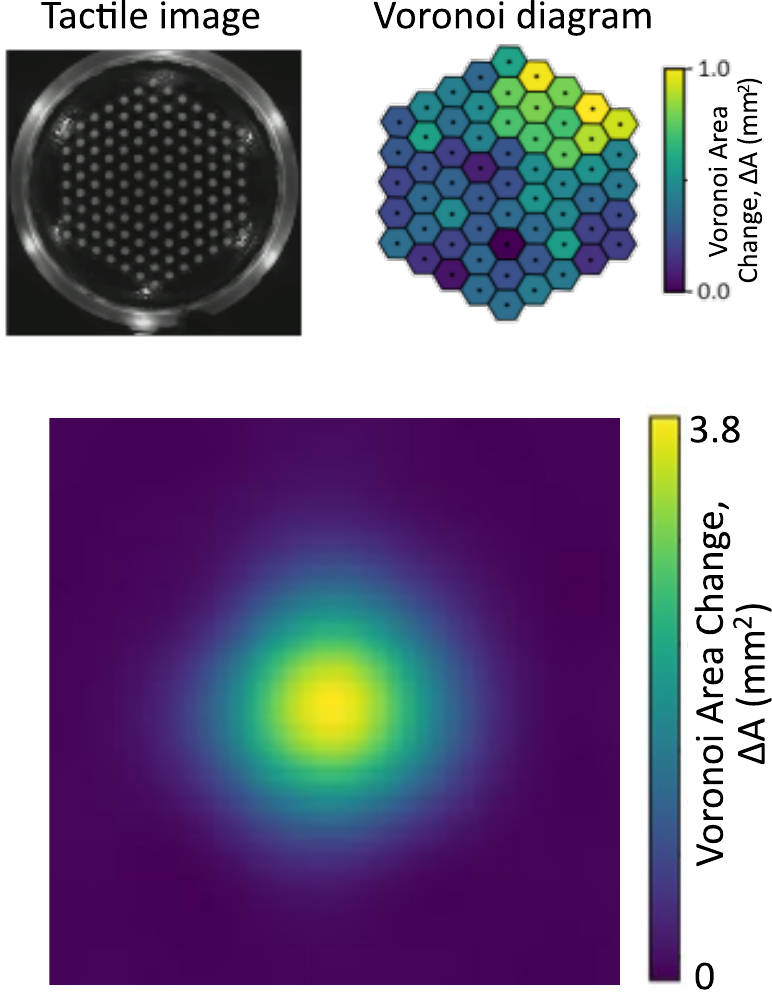}
    	\\
		{\bf (a) Tactile robotic setup} & {\bf (b) Tactile sensor components} & {\bf (c) Mapping of mid-air haptics}\\
	\end{tabular}
	\caption{Tactile robotic system to test mid-air haptics. The system consists of a Dobot MG400 robot arm, TacTip tactile sensor, UHEV1 ultrasonic array (a). The TacTip is an optical tactile sensor, with an internal camera which images the internal pins of the flat skin (b). The tactile image from the sensor is used to generate a representation of the deformation on the surface of the skin with a Voronoi diagram (c, upper). The system is used for mapping mid-air haptics (c, lower).}
	\label{fig_methods_overview}

\end{figure*}

\subsection{Testing mid-air haptics}
Various methods have been used to measure the haptic output from phased ultrasonic arrays. These methods can be quantitative, in which specific measurements can be taken such as the pressure of the ultrasound or the deflection imparted to skin, or qualitative, in which the overall shape of the stimulus is evaluated from creating a visual map. 

One method to measure interaction of the ultrasound with skin-like materials is a Laser Doppler Vibrometer~(LDV), which is an established tool for non-contact vibration measurement. By generating focal points onto a skin-like silicone surface, LDV can measure the resulting surface deformation to test how ultrasonic waves propagate in human skin \cite{frierUsingSpatiotemporalModulation2018} and give insight into the shapes generated by modulating the ultrasound \cite{chillesLaserDopplerVibrometry2019}.
An advantage of LDV is that it is sensitive to small displacements ($\lesssim$1 microns) of the material surface;
however, it relies on costly, specialized equipment ($\sim$£250,000).
An alternative is to use a pressure-field microphone to scan the ultrasound pressure of the generated focal point \cite{kappusSpatiotemporalModulationMidair2018}.
This method directly measures the ultrasonic output, which can be useful if a systematic evaluation of the sound pressure is needed; however, it is not enough to test the interaction with other materials such as human skin.

An oil bath is another useful qualitative testing tool, because acoustic pressure from an ultrasound haptic array deflects the oil surface, which can be imaged for visual inspection \cite{longRenderingVolumetricHaptic2014}.
An advantage of this method is that the experimenter can directly see the haptic output without needing additional data processing; 
however, a disadvantage is that the method does not permit an immediate quantification of the haptic sensation. The technique also suffers from artifacts due to ultrasound resonance within the oil \cite{longRenderingVolumetricHaptic2014}.

In practice, the most reliable method to test haptic displays is to gather feedback from human users. Participants have been asked to identify how many focal points they feel~\cite{carterUltraHapticsMultipointMidair2013}, what shapes they identify \cite{longRenderingVolumetricHaptic2014, ruttenInvisibleTouchHow2019}, or whether they can feel anything at all to determine the detection threshold across various ultrasound frequencies~\cite{takahashiTactileStimulationRepetitive2020}.
The main advantage of user studies is that they take input directly from humans, who are the intended users of the device;
however, relying on users is time-consuming, subjective, and costly, making these studies inappropriate for testing involving quantitative measurements or early hardware development.

All the above methods for sensing and evaluating mid-air haptics have their pros and cons. A systematic method of evaluation, such as the LDV, requires external setups with costly, specialized equipment. Similarly, microphones provide a means of systematic evaluation, but are not a biomimetic tool (measuring sound not touch) and do not consider the interaction with materials such as skin. While user studies are essential for evaluating the sensations, as the haptic displays are intended for human use, they are time-consuming, subjective and appropriate for later stages of development. Together, these considerations highlight the need for an efficient, biomimetic testing method for use during the earlier stages of device development.

\subsection{Tactile sensing for mid-air haptics}

Developments in tactile sensing technology provide a promising tool 
for emulating human touch.  Using a tactile sensor could address the need for a more biomimetic method for evaluating mid-air haptics, bringing in some of the advantages of quantitative methods with user studies. 

A microphone-based tactile sensor array has been used to sense mid-air haptics generated with an ultrasonic array \cite{sakiyamaEvaluationMultiPointDynamic2019}, with contact on the sensor surface causing a pressure change in an underlying cavity containing the microphone. This has been tested with sensor skins that emulate skin, such as a grooved-pattern to imitate fingerprints \cite{sakiyamaMidairTactileReproduction2020}. Being an array-based tactile sensor, it can capture spatial data, but the spatial resolution of the sensor is limited by the size of each element in its array (currently 11\,mm), each of which needs to fit a separate microphone.

Tactile sensing has been identified as an important modality for enabling robots to interact with their surroundings, leading to a developing area of research in interpreting tactile data. One task in common use is contour following, a haptic exploratory procedure used by humans to determine an object's shape \cite{ledermanHandMovementsWindow1987}. Similarly, a robotic system with a tactile sensor can be used to explore an object by maintaining contact with its contour  \cite{martinez-hernandezActiveSensorimotorControl2017,leporaExploratoryTactileServoing2017, leporaPixelsPerceptsHighly2019, liControlFrameworkTactile2013,kappassovTouchDrivenController2020}. 
The availability of advanced tactile robotic systems, and their use to emulate human tactile exploratory procedures, indicates their potential for testing mid-air haptics in a way similar to how humans would interact with mid-air haptic sensations.


\section{Methods}

\subsection{Tactile sensor}
In this work, we sense mid-air haptics by using a biomimetic tactile sensor mounted on a robot arm that moves it over the ultrasonic array (Fig. \ref{fig_methods_overview}a), similar to a setup used for probing physical stimuli~\cite{ward-cherrierTacTipFamilySoft2018, leporaSoftBiomimeticOptical2021}.


The TacTip is a soft biomimetic optical tactile sensor~\cite{leporaSoftBiomimeticOptical2021,ward-cherrierTacTipFamilySoft2018}; here we will explore its suitability for systematic sensing and evaluation of mid-air haptics. The tactile sensor is biologically inspired by \textit{glabrous} (hairless) human skin, which has internal dermal papillae that focus strain from the skin surface down to mechanoreceptors. The TacTip mimics this structure with an outer rubber-like skin connecting to inner nodular pins that amplify surface strain into inner mechanical movements. An internal camera tracks the movement of these artificial papillae optically, enabling ready transduction of the deformation of the skin via internal shear. Recently, these signals have been found to resemble recordings from biological tactile neurons~\cite{pestell2022a,pestell2022b}.

The design of the TacTip is modular, allowing for tips with different skin and pin properties to be used. The mid-air haptic display studied in this work emits tiny forces on the order of a few millinewtons \cite{frierSimulatingAirborneUltrasound2022}, which presents a major challenge to designing a sensor that can detect and map these small forces.
We met this challenge by introducing a highly compliant artificial tactile skin to maximize its deformation under small forces. The pins in the TacTip design also act as levers to amplify the small sensations produced by surface indentation. To find the most appropriate skin, we tested sensing surfaces of various shapes, both flat and hemispherical, as well as changing the internal support for the skin in terms of including or not including an internal supporting gel. Our conclusion is that the most suitable TacTip skin has a flat 40\,mm dia. profile without internal gel (Fig. \ref{fig_methods_overview}) \cite{alakhawandSensingUltrasonicMidAir2020}: then the skin deforms sufficiently to render detectable the forces from mid-air haptic focal points. 

Note also that human skin and rubber acoustic impedance are similar at $1.6\times10^9$ and $1.9\times10^9$ kg/m$^2$s compared with air at $4\times10^2$kg/m$^2$s. Hence, both skin and the TacTip will undergo similar acoustic radiation force from the ultrasound haptic array. The skin indentation may differ depending on the mechanical properties, but previous work with the TacTip has attempted to approximately match human skin~\cite{leporaSoftBiomimeticOptical2021}.

\subsection{Robotic arm}
We use a low-cost, desktop robot arm: the Dobot~MG400 4-axis arm designed for affordable automation. The base and control unit has footprint 190\,mm$\times$190\,mm, payload 750\,g, maximum reach 440\,mm and repeatability $\pm0.05\,$mm. The main constraint is that only the $(x,y,z)$-position and rotation around the $z$-axis of the end effector are actuated, but that is not an issue for this work. Another benefit of the Dobot MG400 desktop robot is the accessibility of the API: it is open-source and written in Python. To accompany this paper, we have released a version of our CRI robot interface libraries that integrates this API and that can be updated to include other functionality as the robot firmware is improved. 

\subsection{Ultrasound phased array}
To generate the mid-air haptic sensations for our experiments, we used the Ultrahaptics Evaluation Kit (UHEV1) developed by Ultraleap, which includes an array of 256 ultrasonic transducers operating at 40\,kHz to generate focal points in mid-air with an update rate of 16\,kHz. Using the device's accompanying software, the focal points generated by the array can be modulated to generate various sensations. With Spatiotemporal Modulation (STM), one focal point can be moved rapidly along the path of the desired shape, producing the illusion of a static shape that can be felt by a human hand in mid-air (Fig. \ref{fig_intro}a). 

In this work, to verify that our system works we test two main cases: (1) An unmodulated (UM) focal point and (2) Spatiotemporal modulation (STM) of a focal point. In the first case (UM), the acoustic pressure and the position of the focal point are constant with the focal point generated 20\,cm above the center of the array. In the second case, the acoustic pressure is constant while the position of the focal point is moved along the path of a circle of 20\,mm diameter, at spatiotemporal modulation frequency of 70\,Hz. We chose these two cases as they have been tested using a Laser Doppler Vibrometer (LDV) \cite{chillesLaserDopplerVibrometry2019, frierSimulatingAirborneUltrasound2022}, and so we can directly compare our results.

To extend our results, we test our methods on six more STM shapes: (1) line, (2) triangle, (3) square, (4) small cross, (5) large cross and (6) rose. This presents a variety of different shapes, which highlights that the method can sense corners, open curves and closed curves. The line, square, triangle, and small cross all have sides of 4\,cm. The large cross and rose are 6\,cm at the longest dimensions. All the shapes are generated at a spatiotemporal modulation frequency of 100\,Hz at height 15\,cm above the ultrasonic array with constant intensity.

\subsection{Systematic mapping of mid-air haptics}
Here, we present the methodology we developed to evaluate mid-air haptics with the tactile robotic system.

Our method involves \textit{systematic mapping}, using a predefined sequence of motions by the sensor: here a 9$\times$9 grid equally spaced 10\,mm apart to cover an 80\,mm-square region.
The sensor captures data which covers $\sim$20\,mm, with an overlap of  $\sim$10\,mm in the sensor data between each position. The overlap in the data reduces the noise from the sensor and gives higher confidence in the measurements. The robot arm moves the sensor across the grid points in a pre-defined sequence starting at the top-left and finishing at the bottom-right of the scans in Figs~4,5. The time taken to gather the data is $\sim$4\,min for each run across a haptic shape.

At each grid point, the sensor waits for 2 seconds for the skin of the sensor to reach close to steady state deformation, then captures tactile images for 1 second at a rate of 30 frames per second. These images are then processed to find the positions of each marker on the pins. These marker positions are used to generate a bounded Voronoi diagram, from which the change in areas of the cells give a third dimension of the sensor data associated with indentation (Fig. \ref{fig_methods_overview}c, upper panel) accompanying the $x$- and $y$-shear dimensions~\cite{cramphornVoronoiFeaturesTactile2018} (using the spatial module from SciPy library).
The areas of the cells in the Voronoi diagram, $\Delta A$, increase as the skin is compressed due to a pressure on its surface (Fig.~2c, upper panel). The average of $\Delta A$ over the 30 frames collected is calculated as the deformation of the tactile skin at each pin.

We combine the Voronoi area changes from individual sensor readings together using Gaussian Process Regression (GPR) with a squared-exponential covariance function (using the GP module in scikit-learn library).
The output of this process allows us to plot a representation of the mid-air haptic sensations (Fig. \ref{fig_methods_overview}c). For further details, we refer to our paper in which we first introduced this process \cite{alakhawandSensingUltrasonicMidAir2020}.

\subsection{Expanding the system: autonomous haptic exploration}

\begin{figure}[t]
    \centerline{\includegraphics[width=\linewidth,trim={0 0 0cm 0},clip]{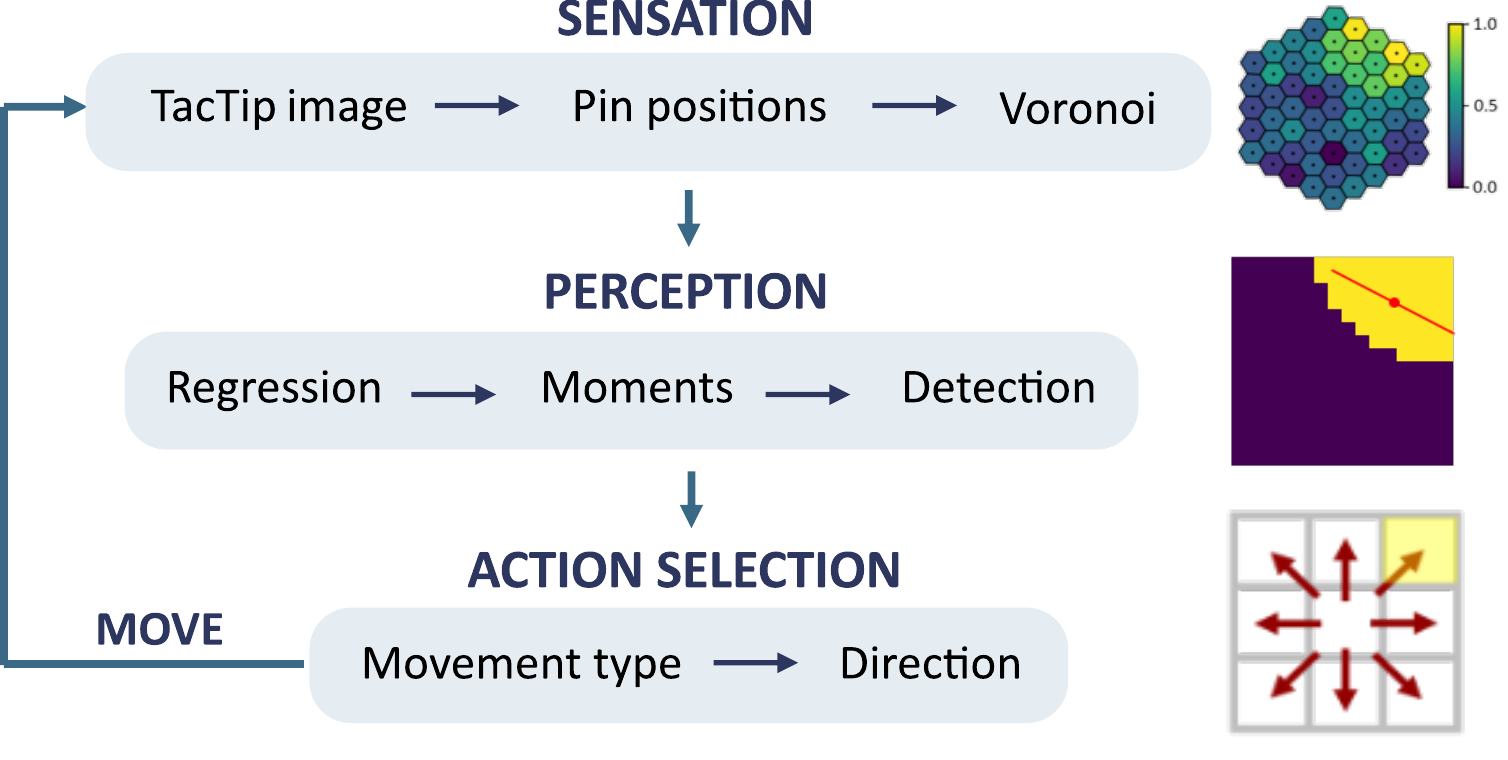}}
    \caption{Autonomous haptic exploration algorithm with three levels: sensation, perception, and action selection. First, the system senses the intensity of the stimulus felt over an array of pins from the sensor. Second, it detects the position and orientation of the stimulus. Third, it selects an action to take that will allow it to continue exploring the shape.}
    \label{fig_methods_exploration}
\end{figure}


\begin{algorithm}[b]
\caption{Action selection}\label{alg:action}
\begin{algorithmic}
\If{centroid $(\bar{x},\bar{y}) > 5\,$mm from sensor position}
    \State Move towards centroid
    \State Find angle $\theta$ of vector from sensor position to centroid
\Else
    \State Move along orientation angle, $\theta$, in degrees
    \State Choice of $\theta_1 = \theta,\ \theta_2 = \theta + 180^\circ$
    \State Find angle $\theta_i$ with minimum change from past action
\EndIf
\State Choose discrete action $(x, y)$ closest to the angle from set:
\State $\big\{(a,\!0),\!(a,\!a),\!(0,\!a),\!(\shortminus a,\!a),\!(\shortminus a,\!0),\!(\shortminus a,\!\shortminus a),\!(0,\!\shortminus a),\!(a,\!\shortminus a)\big\}$ with grid-size $a=10$\,mm
\end{algorithmic}
\end{algorithm}

As an extension to our methodology, we show how additional testing modes can be added to this system, such as using real-time control to more intelligently map a stimulus. We present a method for autonomous exploration, in which the system utilizes a sensation-perception-action loop to decide where to move by aiming to stay along the path of the shape (Fig. \ref{fig_methods_exploration}): the robot first senses the stimulus using touch, then perceives the local nature of the stimulus, from which it selects an appropriate action (here a movement). This process is repeated to explore the haptic shape.

\begin{figure*}[t]
	\centering
	\begin{tabular}[b]{@{}c@{\hspace{8pt}}c@{\hspace{2pt}}c@{\hspace{4pt}}c@{\hspace{2pt}}c@{\hspace{4pt}}c@{\hspace{2pt}}c@{}}
		\frame{\includegraphics[width=0.178\linewidth,trim={0 0 0 0},clip]{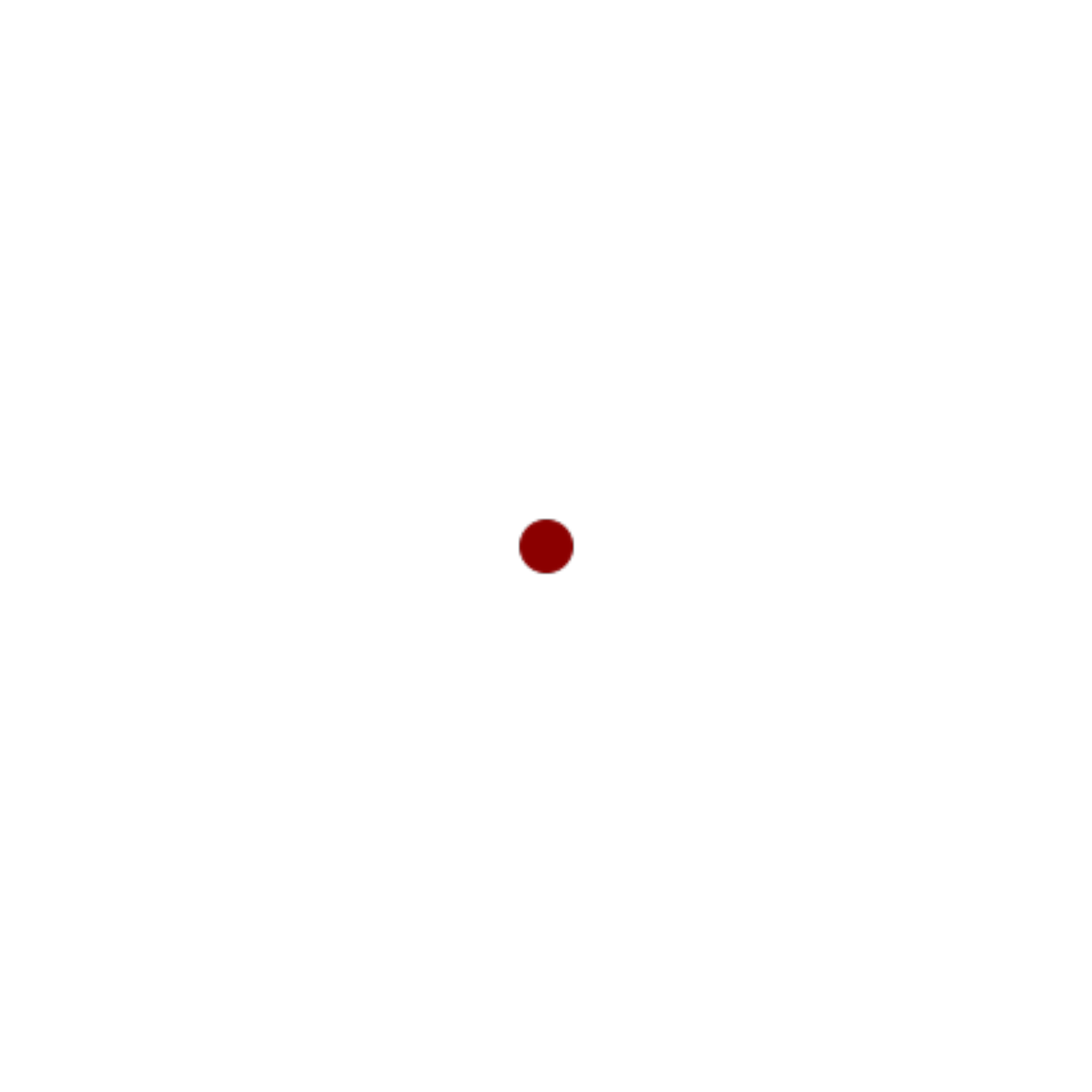}} &
		\includegraphics[width=0.178\linewidth,trim={0 0 0 0},clip]{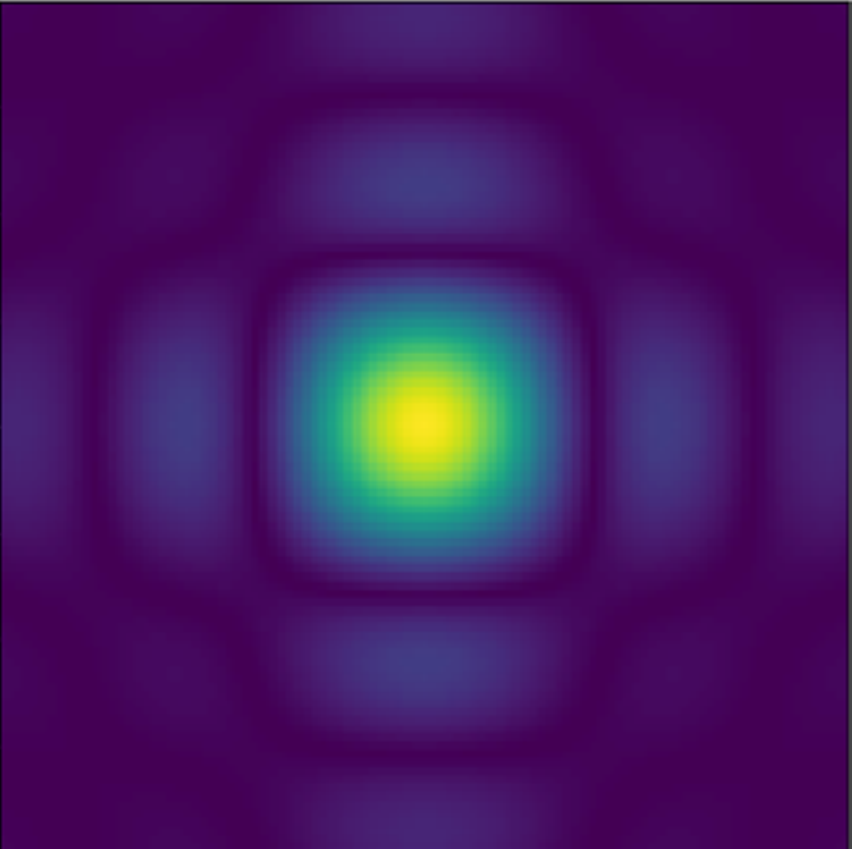} &
		\includegraphics[width=0.055\linewidth,trim={0 0 0 0},clip]{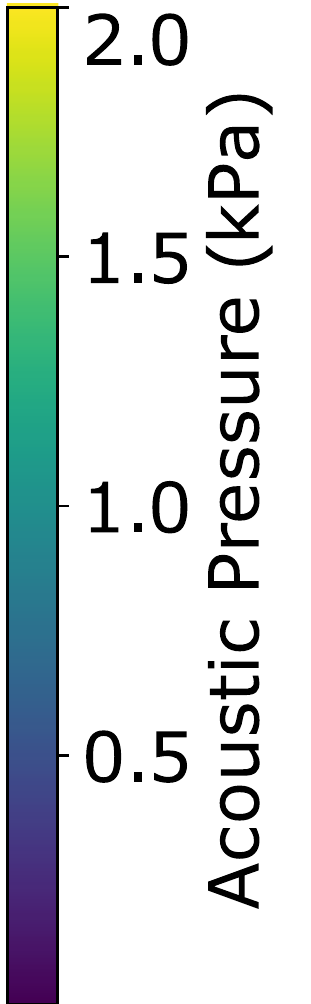} &
		\includegraphics[width=0.178\linewidth,trim={0 0 0 0},clip]{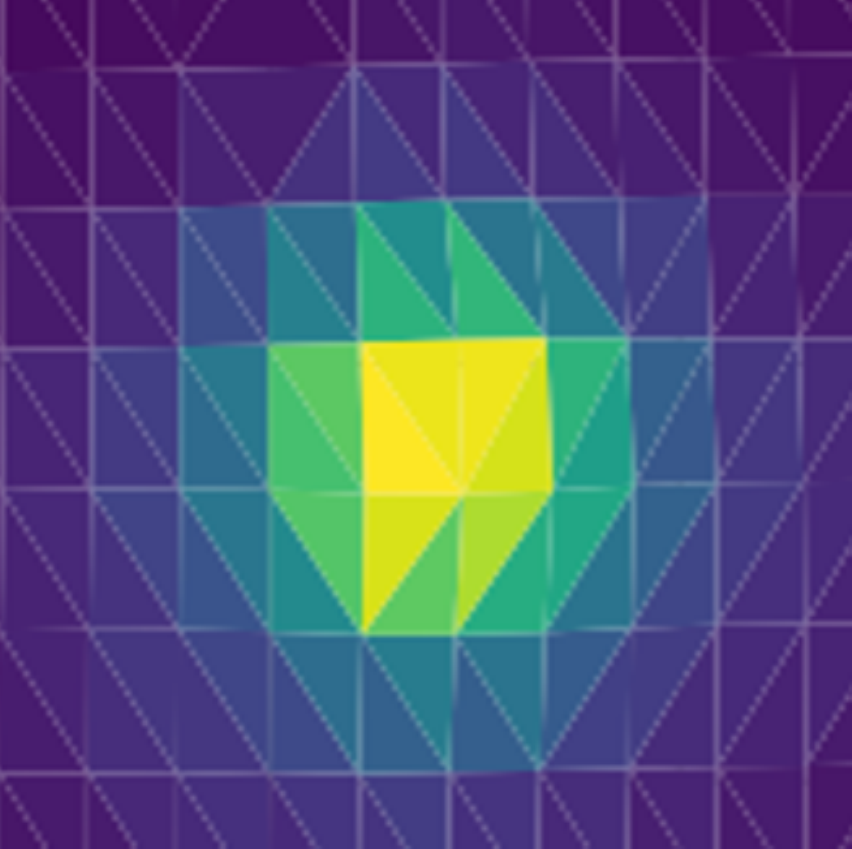} &
		\includegraphics[width=0.055\linewidth,trim={0 0 0 0},clip]{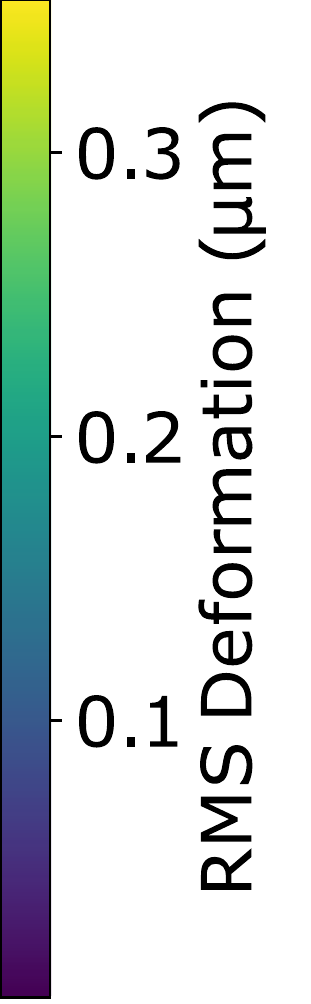} &
		\includegraphics[width=0.178\linewidth,trim={0 0 0 0},clip]{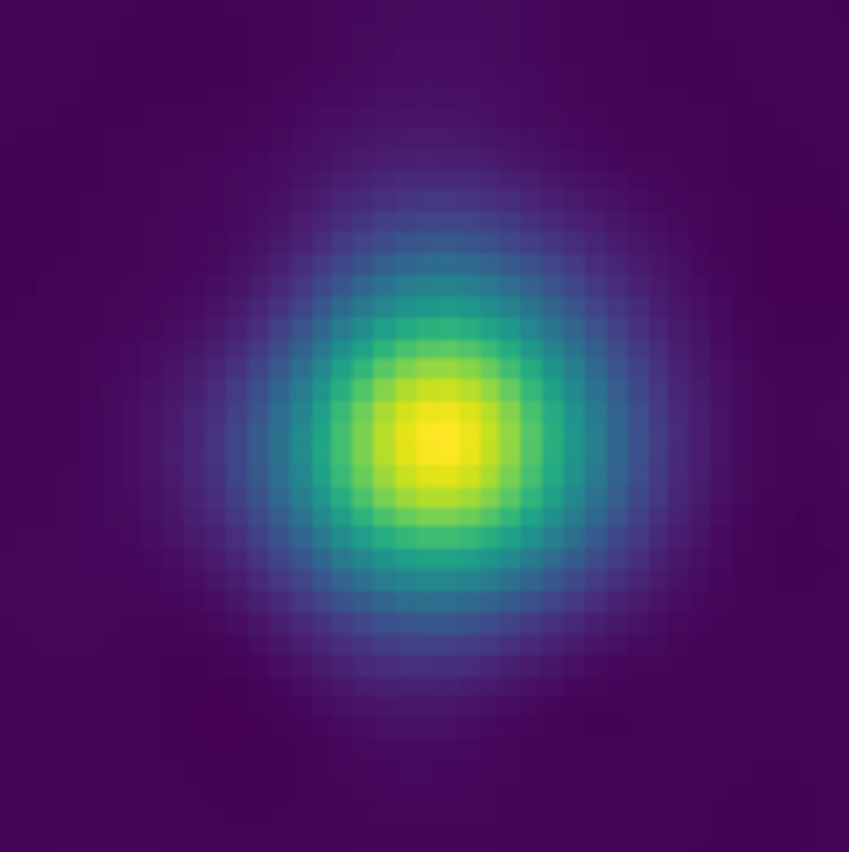} &
		\includegraphics[width=0.055\linewidth,trim={0 0 0 0},clip]{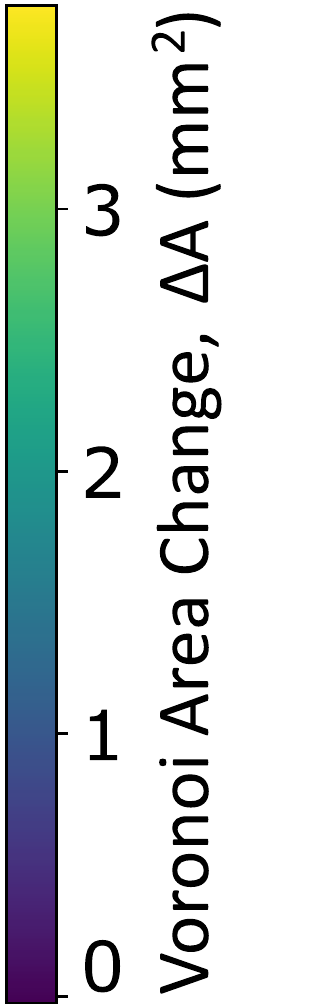}
		\\
		\frame{\includegraphics[width=0.178\linewidth,trim={0 0 0 0},clip]{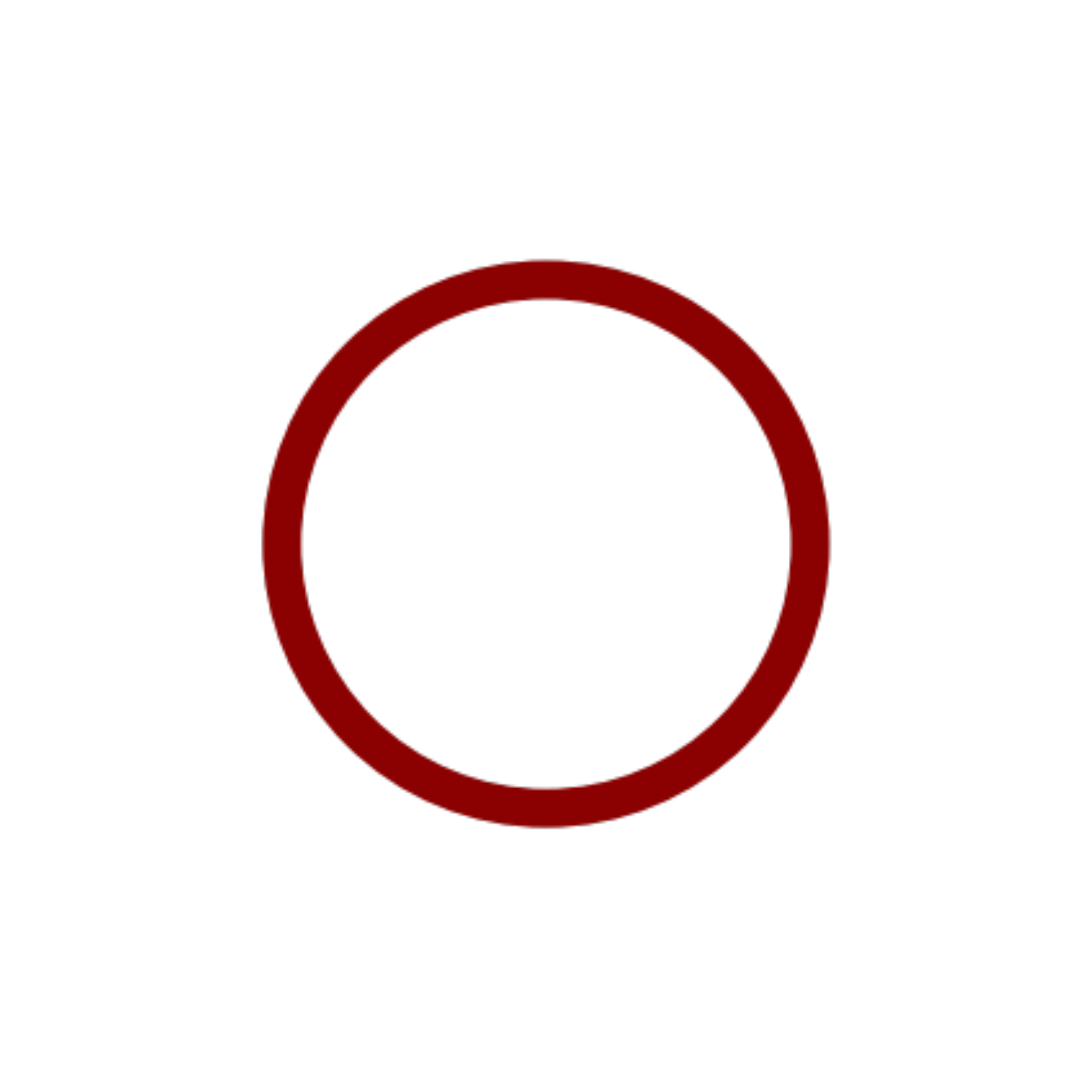}} &
		\includegraphics[width=0.178\linewidth,trim={0 0 0 0},clip]{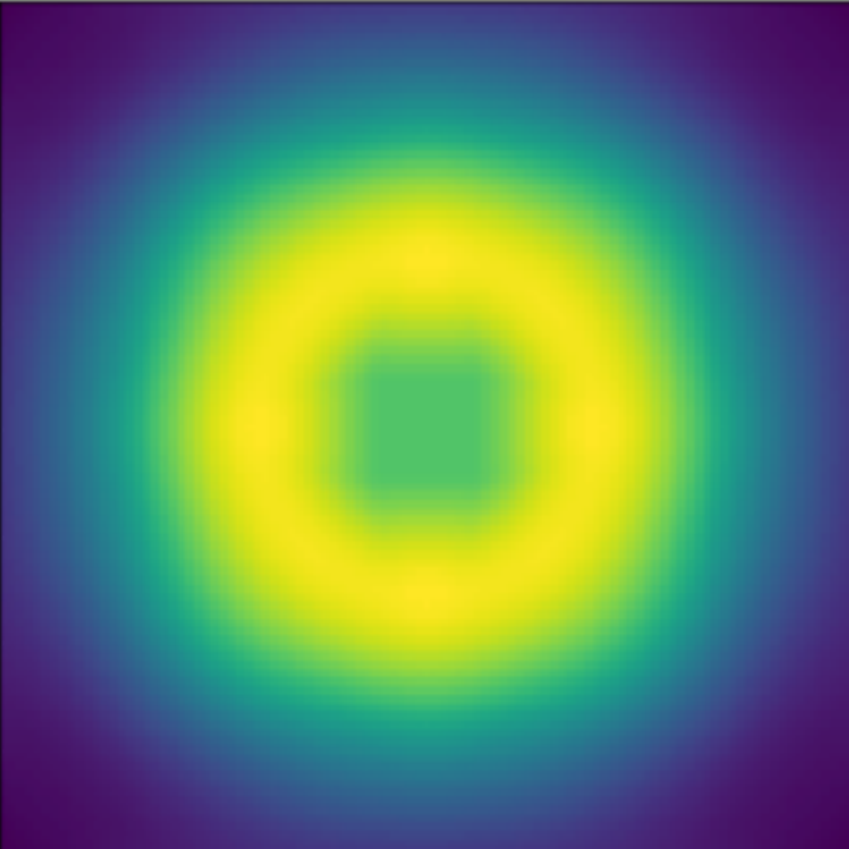} &
		\includegraphics[width=0.055\linewidth,trim={0 0 0 0},clip]{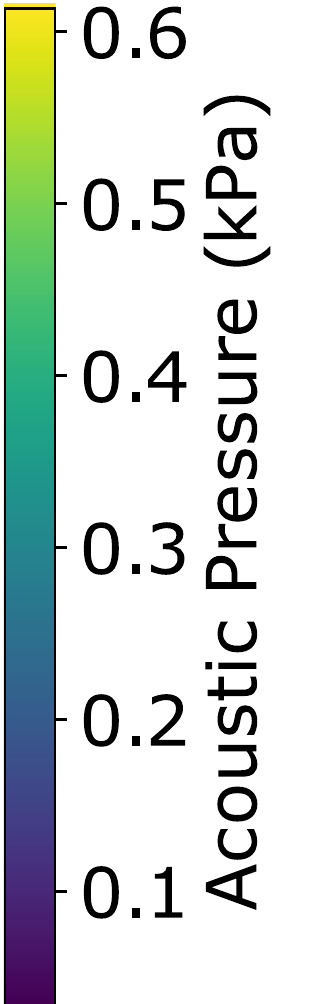} &
		\includegraphics[width=0.178\linewidth,trim={0 0 0 0},clip]{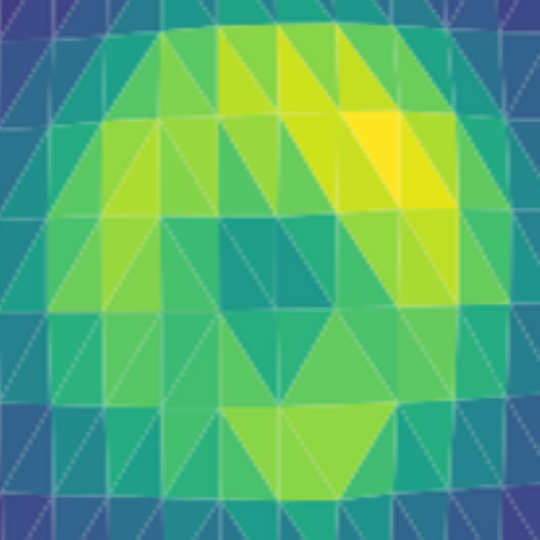} &
		\includegraphics[width=0.055\linewidth,trim={0 0 0 0},clip]{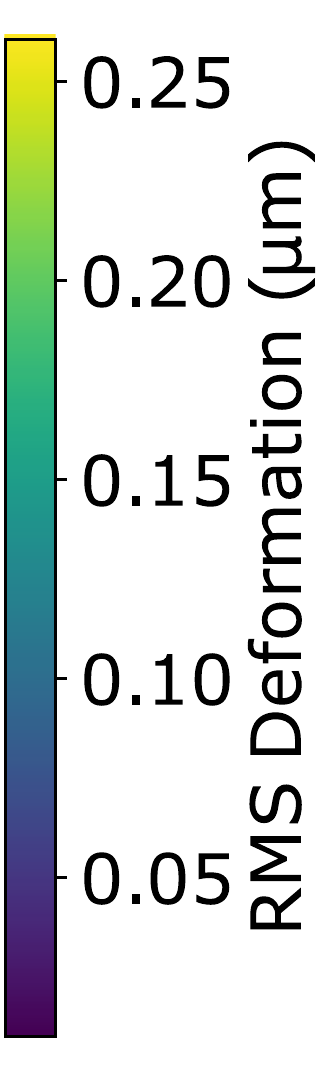} &
		\includegraphics[width=0.178\linewidth,trim={0 0 0 0},clip]{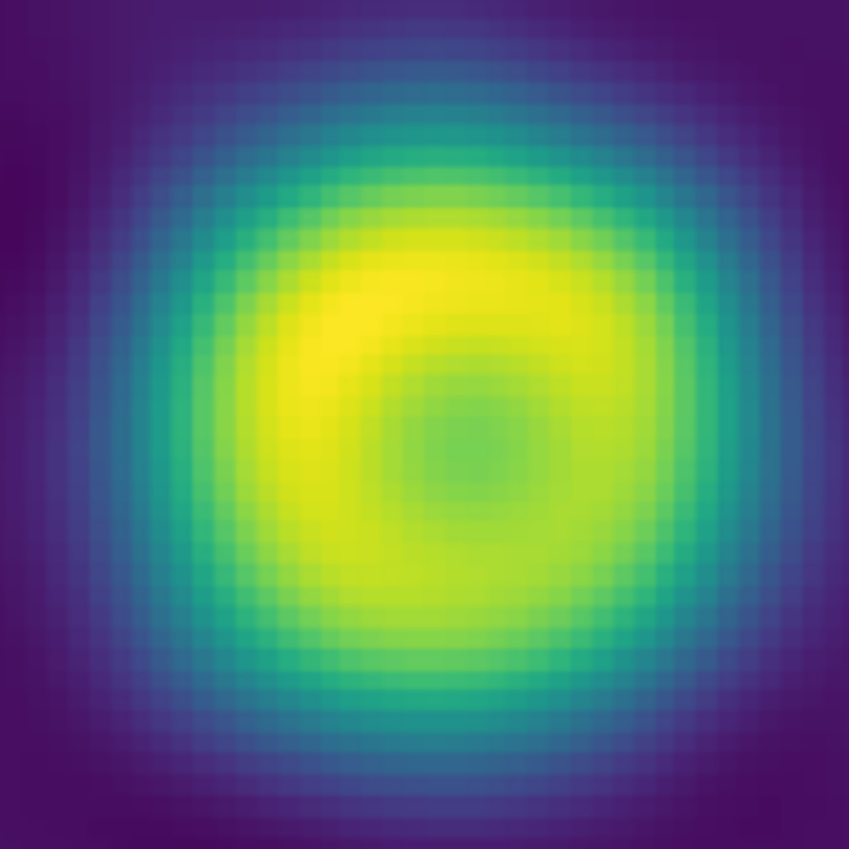} &
		\includegraphics[width=0.055\linewidth,trim={0 0 0 0},clip]{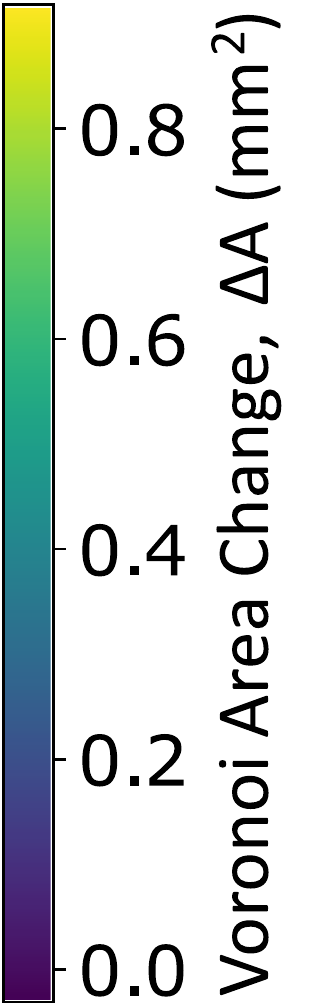}
		\\
		{\bf (a) Focal point path} & {\bf (b) Acoustic simulation} & {} &{\bf (c) LDV measurement} & {} & {\bf (d) Tactile measurement} & {}
	\end{tabular}
	\caption{Testing the mapping of mid-air haptics by the tactile robot. (a) The path of the focal point; (b) the resulting simulated acoustic field; (c) the measured displacement of skin-like material due to the haptic sensation with a Laser Doppler Vibrometer \cite{chillesLaserDopplerVibrometry2019}; and (d) our results from the tactile robot. All plots cover a 40\,mm$\times$40\,mm square area.}
	\label{fig_results_validation}
\end{figure*}

The sensation step involves capturing and processing a tactile image  to generate a bounded Voronoi diagram with areas representing stimulus intensity (see Section~IIIC). 
For the perception process, we use an analytic model to interpret the stimulus intensities felt by the TacTip to locate and characterize the haptic sensation. First, we perform Gaussian Process Regression (GPR) over the data, then we binarize the regressed map using a fixed threshold to reduce background noise; finally, we calculate the image moments using a standard technique from computer vision:
\begin{equation}
M_{ij}= \sum_{x} \sum_{y} x^i y^j I(x,y),\label{eq_moments}
\end{equation}
where $I(x,y)$ is the intensity of the stimulus as measured by the Voronoi area changes at each pin. Using the calculated moments, we can find the centroid and orientation of the local haptic stimulus from the first-order moments ($M_{ij}$) and second-order central moments ($\mu^\prime_{i,j}$), respectively:
\begin{equation}
(\bar{x},\bar{y})= \Big(\frac{M_{10}}{M_{00}}, \frac{M_{01}}{M_{00}}\Big),\hspace{0.4em}
\theta=\frac{1}{2}\tan^{-1}\Big(\frac{2\mu^\prime_{1,1}}{\mu ^\prime_{2,0} - \mu^\prime_{0,2}}\Big).
\label{eq_centroid}
\end{equation}

The action selection step decides where to move next, based on the calculated centroid and orientation of the local haptic stimulus. A set of discrete allowed movements are predefined in a grid space. Algorithm \ref{alg:action} is used to determine which of those movements to take, based on an intuitive, heuristic action-selection process which does not require any training. The goal of the algorithm is to choose a movement to continue exploring the haptic shape.

\section{Results}

\subsection{Testing the tactile robot on mid-air haptics}

To verify the capabilities of our tactile robot for mapping mid-air haptic stimuli, we initially consider two distinct stimuli: (1) a focal point, which is stationary and unmodulated (UM); and (2) a circle, which is spatiotemporally modulated (STM). (Details in Methods, Sec. IIIC.) The systematic mapping process (Sec. IIID) is used to form detailed maps of the haptic stimuli as sensed by the tactile robot.


Fig. \ref{fig_results_validation} shows our results for mapping haptic stimuli alongside some comparator methods (with a comparison in Table \ref{table_results1}). The left panels show the input paths of the focal point. The middle panels show the simulated acoustic output of the array and the deformation it causes on a skin-like surface measured by a Laser Doppler Vibrometer (taken from an experiment carried out in \cite{chillesLaserDopplerVibrometry2019}). Comparing these results show that our system appears to accurately sense the mid-air haptic stimuli, shown by the simulation of the acoustic pressure output. In particular, by comparing our sensed results to the acoustic simulation, we are able to sense the acoustic pressure above a threshold of $\sim$0.4\,kPa.
In comparison with the measurements taken by the LDV, our results appear very similar: they both show the variation in the sensations, with a higher intensity in the middle which gets lower as you move away from the center of the focal point and the path of the circle.

\begin{table}[h]

\caption{Comparison of results shown in Fig. \ref{fig_results_validation}.}
\centering
    \begin{tabular}{llccc}
    \toprule
    \textbf{}                            & \textbf{}          & \textbf{Acoustic}          & \textbf{LDV}              & \textbf{Tactile } \\
                                         &   \textbf{}  & \textbf{Simulation}           & \textbf{Measurement}      & \textbf{Measurement}     \\
                                         \hline
    \textbf{Peak}                       & Point               & 1.0\,kPa                         & 0.35\,\micro m         & 3.77\,mm\textsuperscript{2}                 \\
    \textbf{Value}                      & Circle              & 0.6\,kPa                         & 0.26\,\micro m         & 0.98mm\,\textsuperscript{2}                   \\
    \hline
    
    \multirow{2}{*}{\textbf{Size }}      & Point   & 13\,mm              & 25\,mm          & 19\,mm                  \\
                                         & Circle  & 40\,mm              & 48\,mm          & 34\,mm                  \\
    
    \hline
    
    \multirow{2}{*}{\textbf{RMSE}}   & Point  & -               & 19\%          & 13\%                   \\
                                        & Circle  & -               & 11\%          & 10\%                  \\

    \bottomrule
    \end{tabular}
\label{table_results1}
\end{table}

Our measurements with the tactile sensor indicate that the focal point causes a round indentation of 19\,mm diameter (Table~\ref{table_results1}), measured from the distance at which the signal becomes $>$20\% of that of the central peak. In comparison, the LDV measurements show that the focal point creates an indentation on the skin-like material 25\,mm in diameter, also measured at $>$20\% of the peak value~\cite{chillesLaserDopplerVibrometry2019}. Both these values are larger than the 13\,mm diameter from the acoustic simulation, consistent with acoustic pressure sourcing a broader indentation of the elastic skin surface.

Likewise, the circular stimulus (Fig~4, lower panels) causes a ring-like indentation. This is measured with an outer diameter of 34\,mm with the tactile sensor and 48\,mm with LDV, compared with a 40\,mm from the acoustic simulation, all to $>$20\% of the peak value on the ring (Table \ref{table_results1}). All these values are significantly larger than the 20\,mm diameter of the focal point path, as expected.

\begin{figure*}[t]
	\centering
	\begin{tabular}[b]{@{}c@{\hspace{6pt}}c@{\hspace{6pt}}c@{\hspace{6pt}}c@{\hspace{6pt}}c@{\hspace{6pt}}c@{}}
		\frame{\includegraphics[width=0.15\linewidth,trim={0 0 0 0},clip]{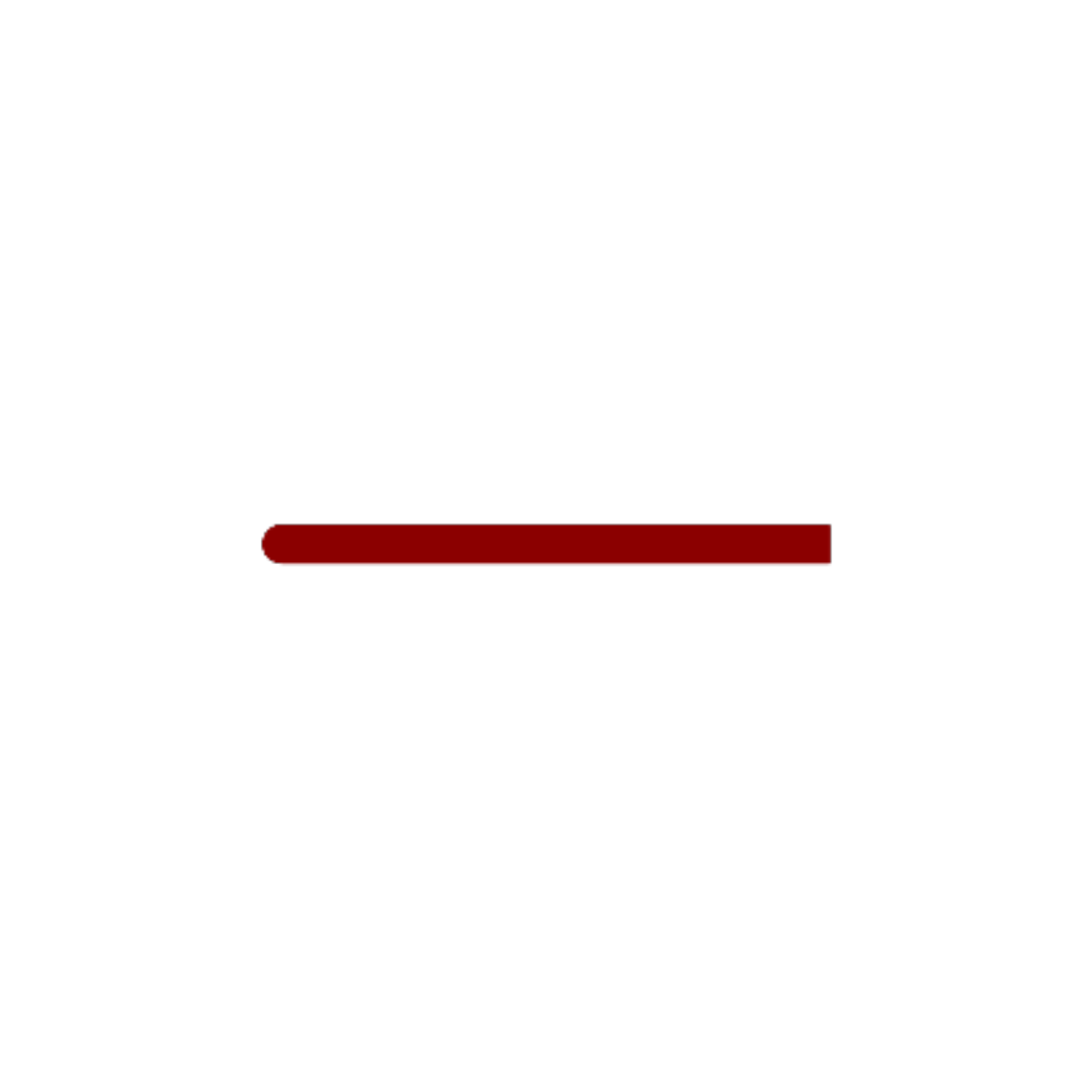}} &
		\frame{\includegraphics[width=0.15\linewidth,trim={0 0 0 0},clip]{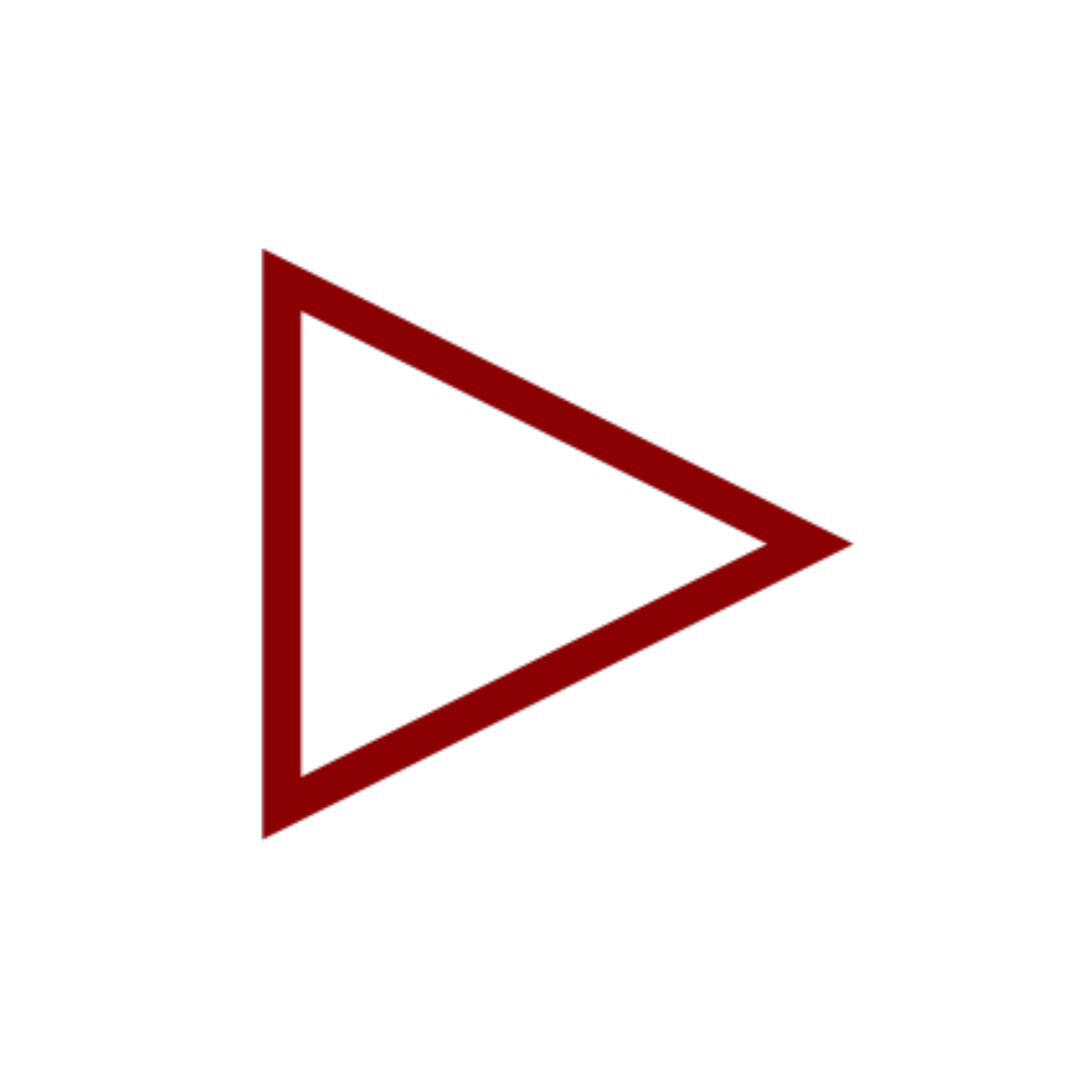}} &
		\frame{\includegraphics[width=0.15\linewidth,trim={0 0 0 0},clip]{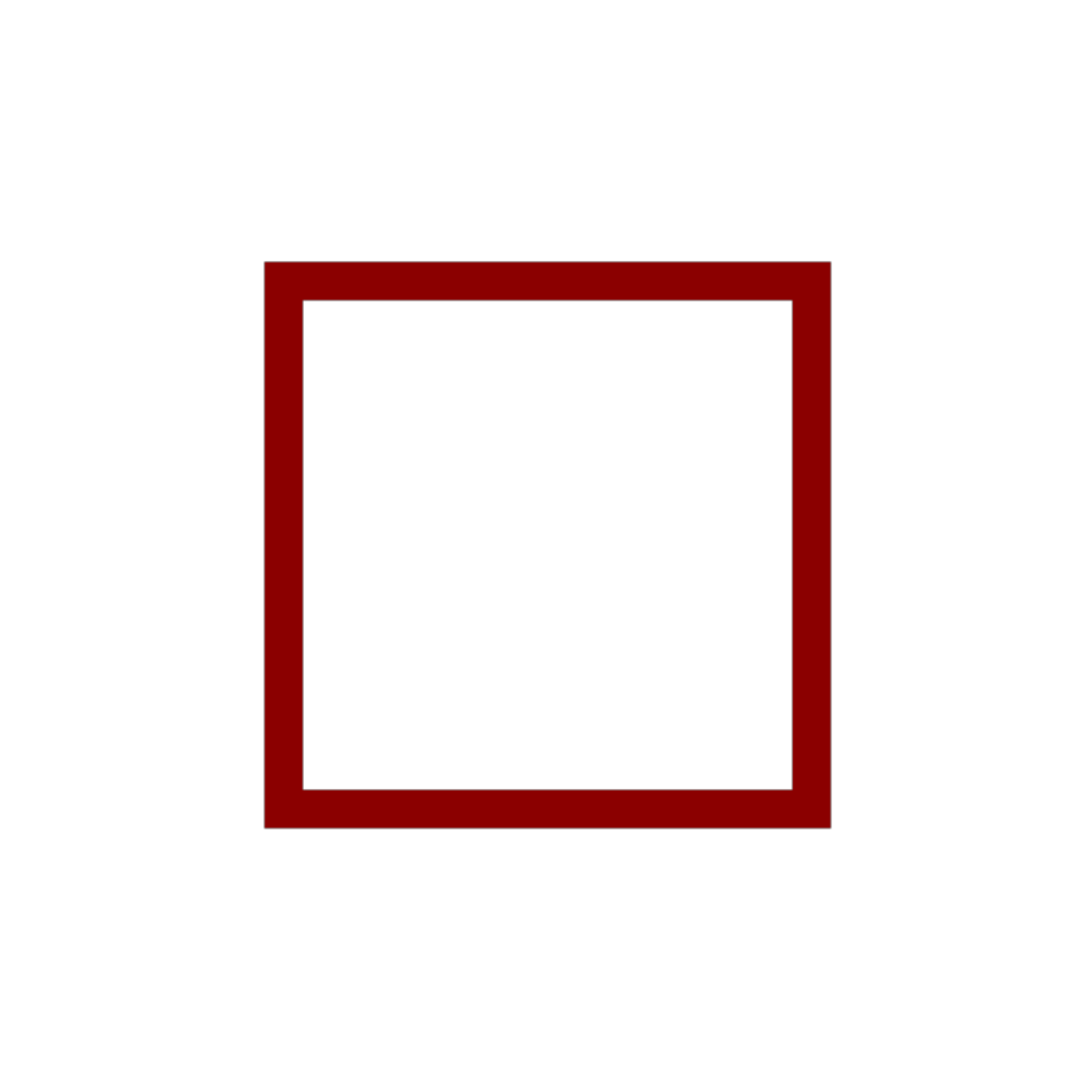}} &
		\frame{\includegraphics[width=0.15\linewidth,trim={0 0 0 0},clip]{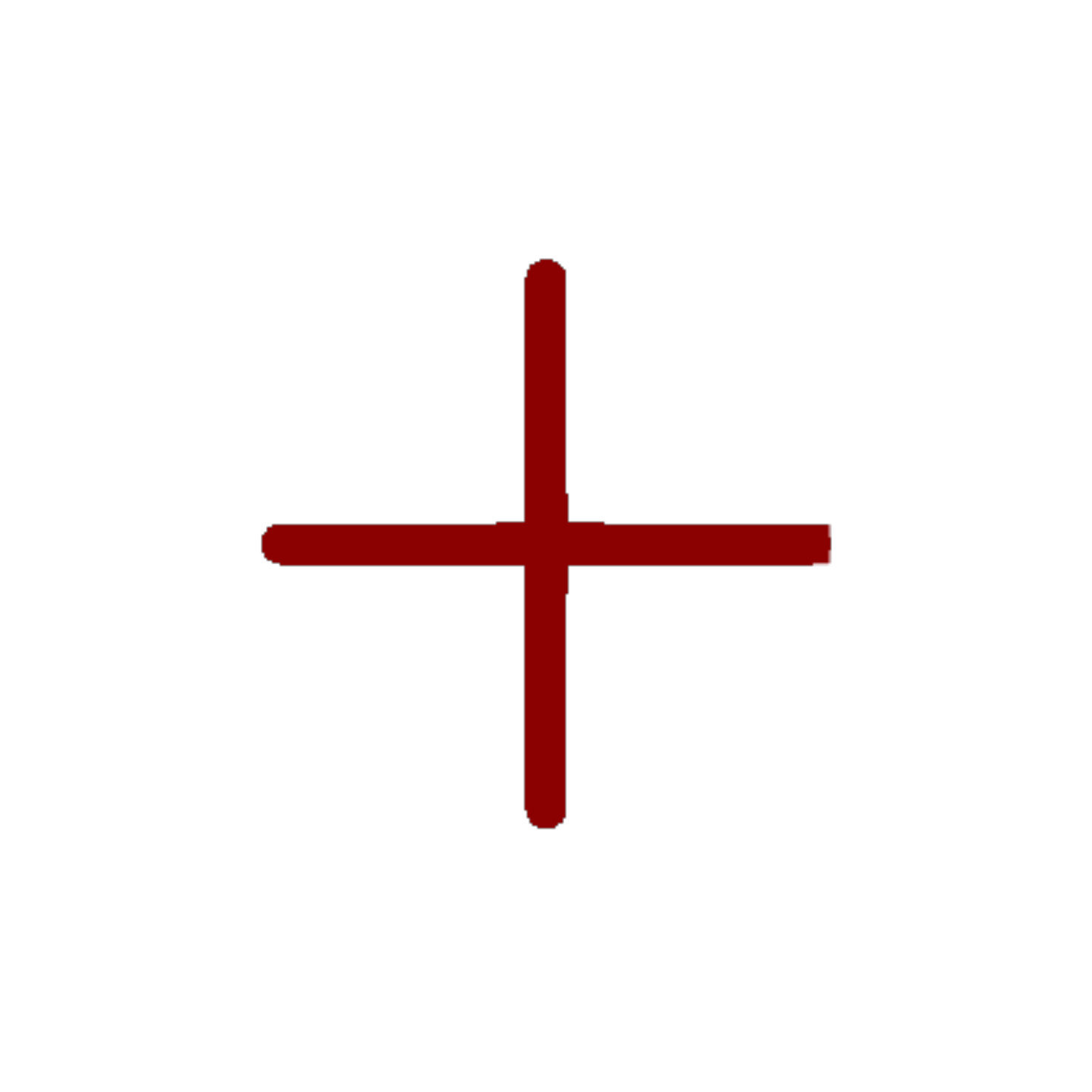}} &
		\frame{\includegraphics[width=0.15\linewidth,trim={0 0 0 0},clip]{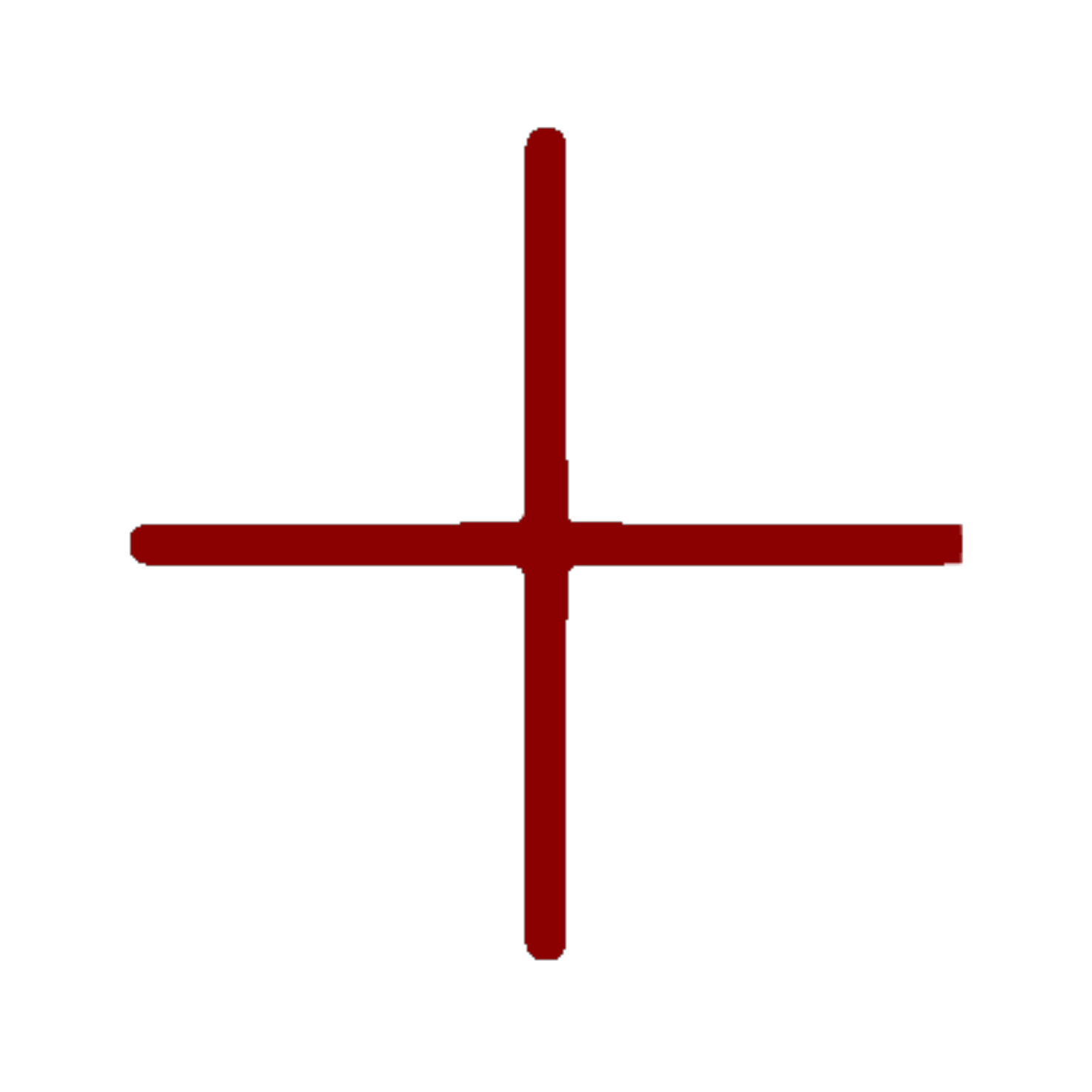}} &
		\frame{\includegraphics[width=0.15\linewidth,trim={0 0 0 0},clip]{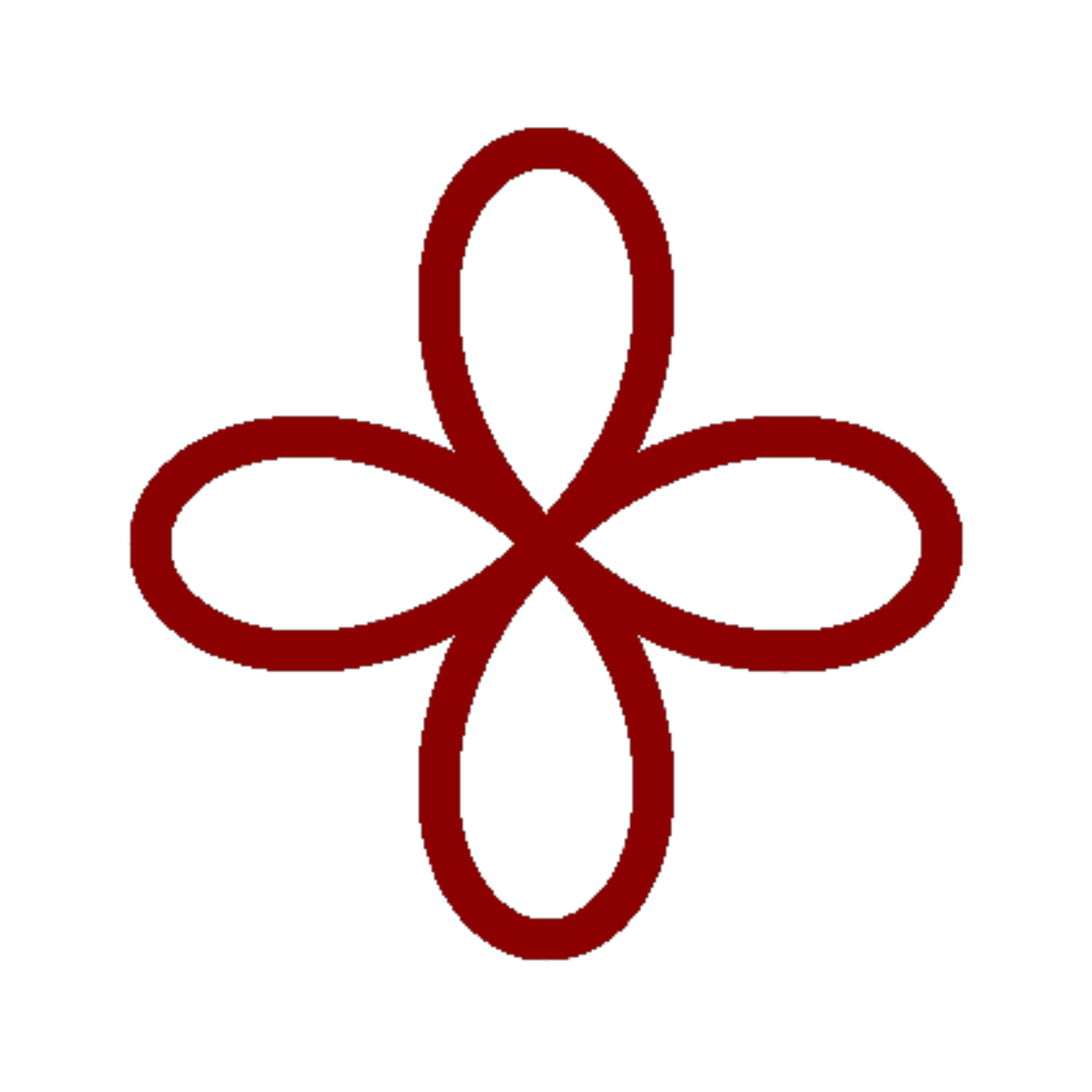}}
		\\
    	\frame{\includegraphics[width=0.15\linewidth,trim={0 0 0 0},clip]{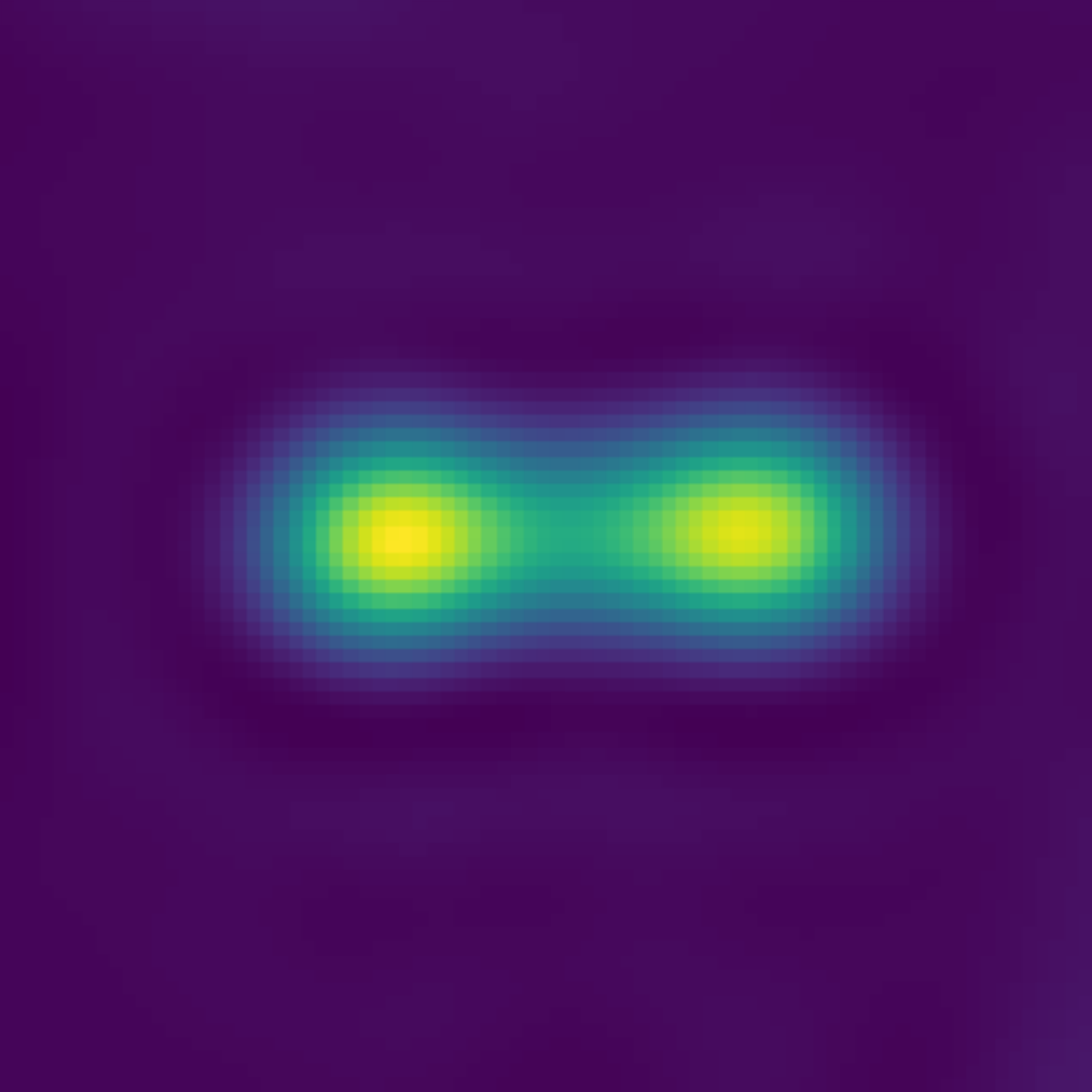}} &
		\frame{\includegraphics[width=0.15\linewidth,trim={0 0 0 0},clip]{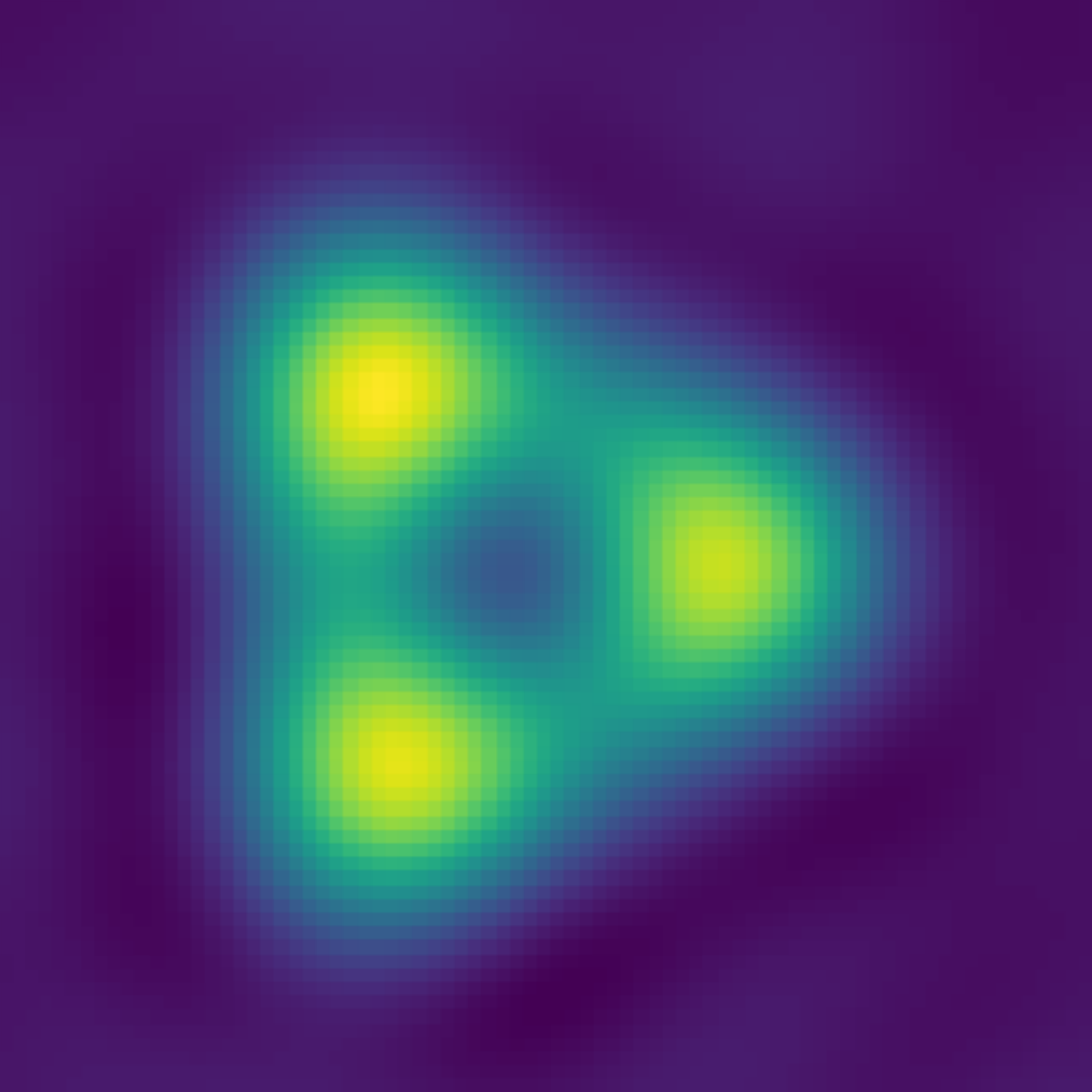}} &
		\frame{\includegraphics[width=0.15\linewidth,trim={0 0 0 0},clip]{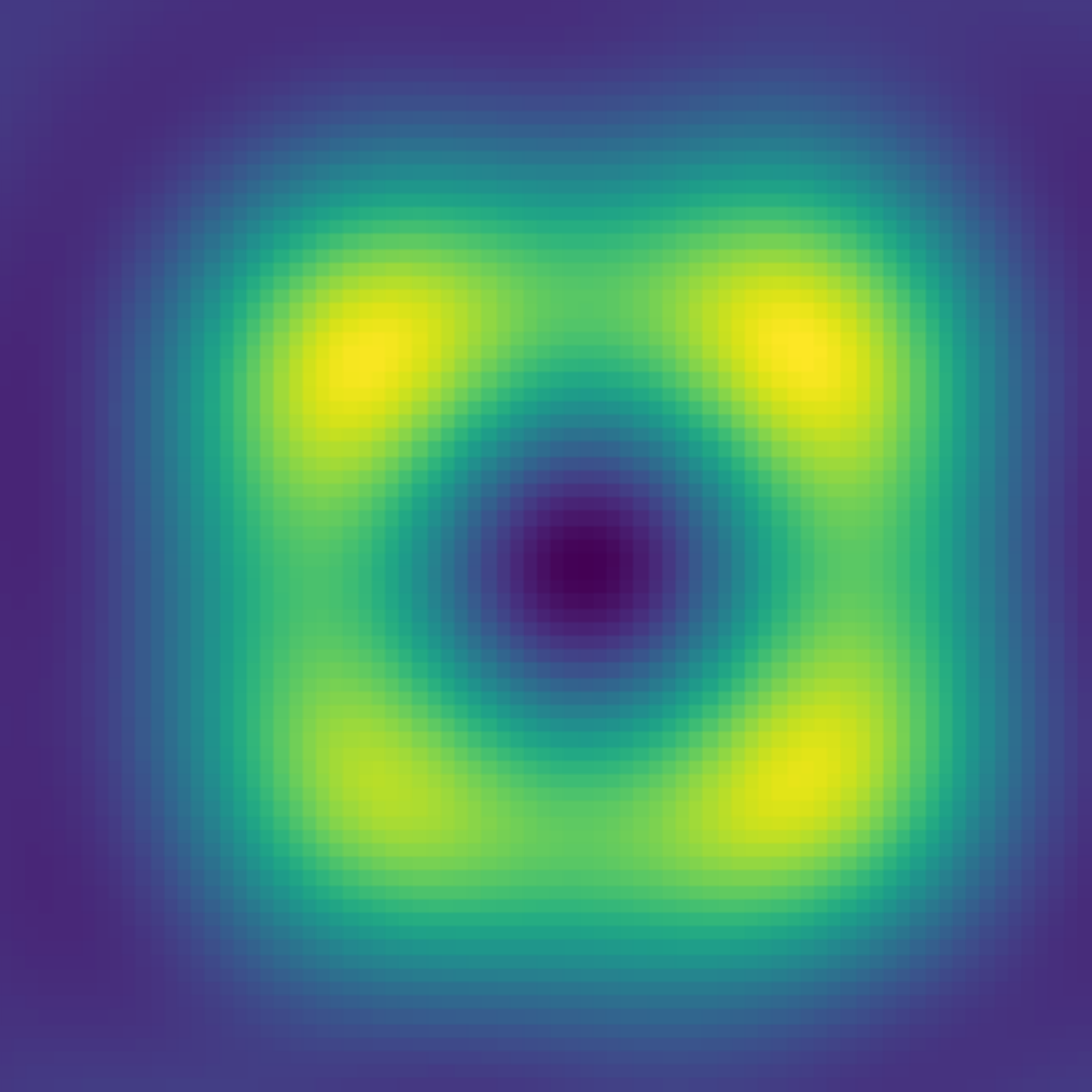}} &
		\frame{\includegraphics[width=0.15\linewidth,trim={0 0 0 0},clip]{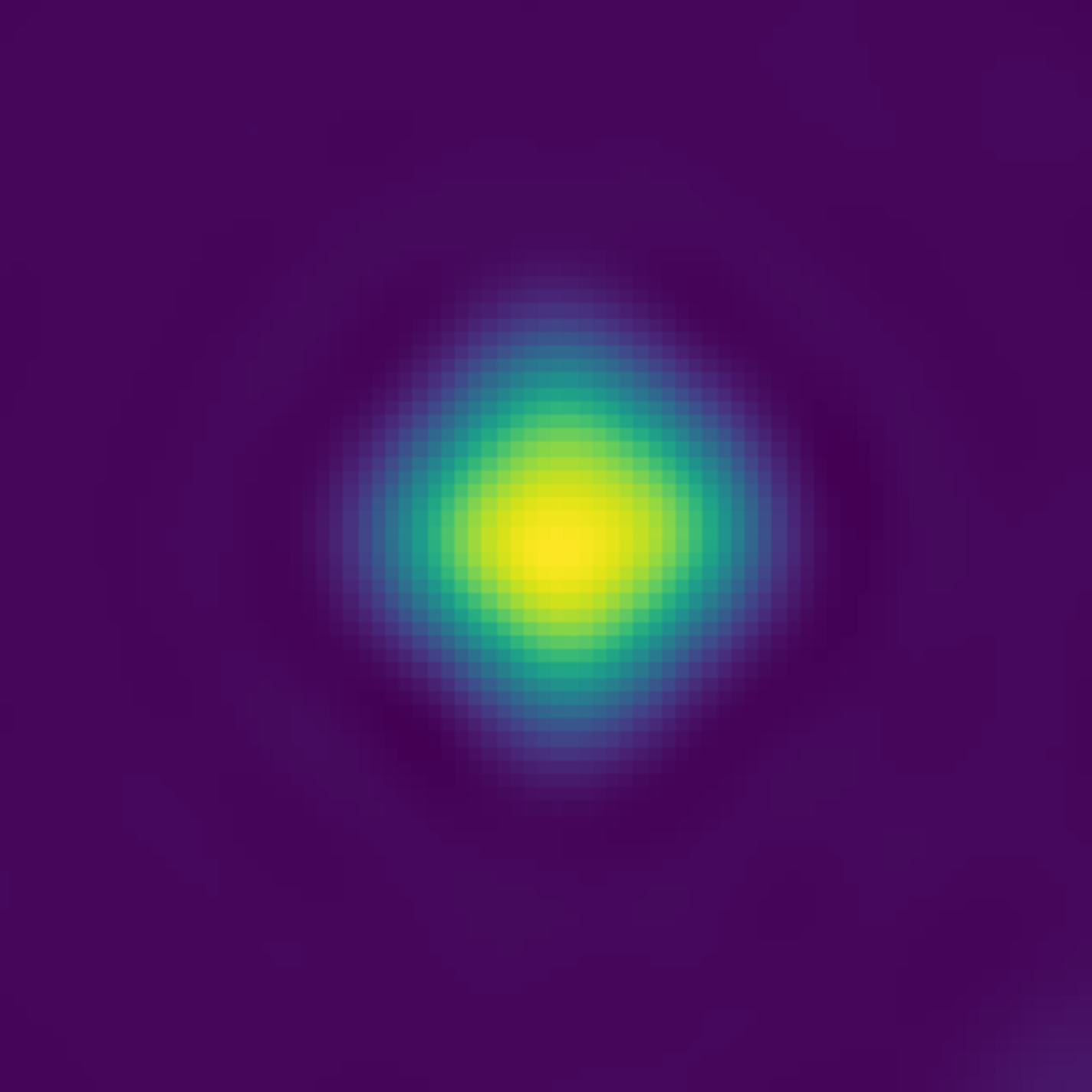}} &
		\frame{\includegraphics[width=0.15\linewidth,trim={0 0 0 0},clip]{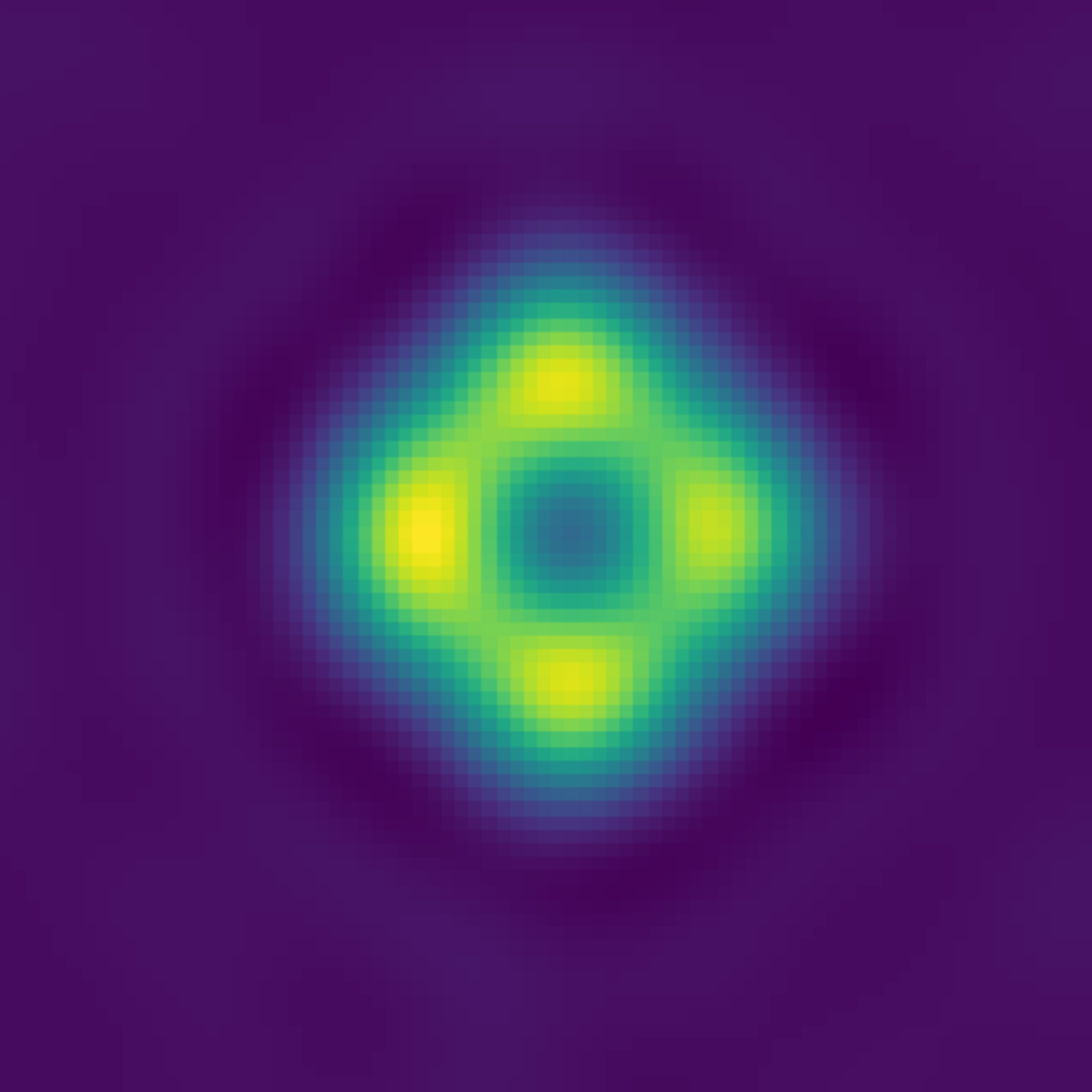}} &
		\frame{\includegraphics[width=0.15\linewidth,trim={0 0 0 0},clip]{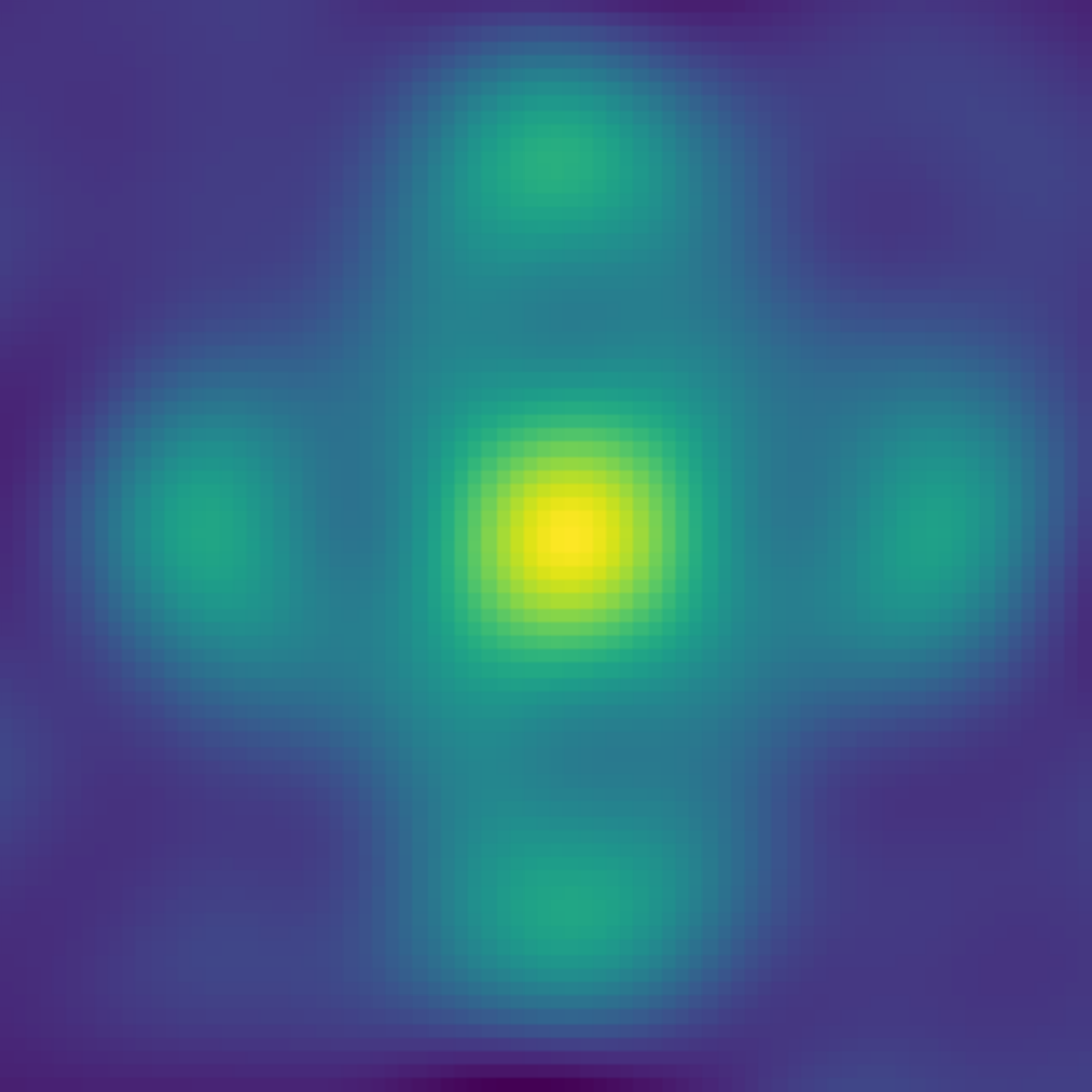}}
	\end{tabular}
	\caption{Systematic mapping of six mid-air haptic shapes: line, triangle, square, small cross, large cross, and rose. The plotted regions cover an 80\,mm$\times$80\,mm square area. The top panel shows the spatiotemporal modulation path of the focal points, the lower panel shows the output of the system, normalized between 0-1 (using the values for max. $\Delta A$ in Table \ref{table_results2}).}
	\label{fig_results_shapes}
\end{figure*}

In these tests, the LDV consistently gave larger estimates than the tactile measurements, which we attribute to the lower spatial resolution of the LDV spreading the signal. Collecting data with the LDV is a relatively slow process compared to tactile and results in single rather than multiple measurements. Additionally, another issue with the LDV is that data is collected at an angle that needs correcting, which makes it difficult to accurately scale the output.

To assess the comparative accuracy of the results, we calculated the “pixel-to-pixel” error of both the LDV and tactile data compared to the acoustic simulation, using a root mean square error (RMSE). After scaling the data between 0-1, we present the RMSE in Table \ref{table_results1} as a percentage. The RMSE for the tactile data is lower for both stimuli, which we attribute to the higher spatial resolution of the tactile map.

\subsection{Mapping mid-air haptic shapes}
To show our tactile robot can also evaluate a variety of mid-air haptic stimuli, we apply it on six distinct shapes generated by the ultrasonic array: (1)~line, (2)~triangle, (3)~square, (4)~small cross, (5)~large cross, and (6)~rose. (See~Methods Sec. IIIC for details of their generation.)

Our systematic mapping method produces detailed visualizations of the mid-air haptic shapes, visually appearing to closely resemble the focal point paths (cf. paths in top row of Fig. \ref{fig_results_shapes} with the maps in the bottom row). 
We observe that the sensor signal is strongest on the corners for all shapes.
%
Overall, the robotic system maps shapes similarly to how humans described them in user studies (tested on the square, triangle, and circle), who interpreted these as constituting the appropriate geometry, but commenting on how blurry they felt~\cite{hajasMidAirHapticRendering2020}.

\begin{table}[b]
\caption{Testing mid-air haptic shapes.}
\centering
    \begin{tabular}{lcc}
    \toprule
                            & \textbf{Approx. path length}      & \textbf{Max. $\Delta A$}      \\
    \textbf{Shape}          & \textbf{(mm)}           & \textbf{(mm\textsuperscript{2})}        \\ 
    \hline
    Line                    & 4                     & 1.54                                      \\
    Triangle                & 13                    & 0.49                                      \\
    Square                  & 16                    & 0.22                                      \\
    Small cross             & 8                     & 1.65                                      \\
    Large cross             & 12                    & 0.73                                      \\
    Rose                    & 28                    & 0.74                                      \\
    
    \bottomrule
    \end{tabular}
\label{table_results2}
\end{table}

Table \ref{table_results2} presents the maximum Voronoi area change sensed by the tactile robot, $\Delta A$, for each shape along with the approximate path length the focal point takes. A larger $\Delta A$ indicates a larger indentation on the skin of the sensor, which would be felt as a stronger sensation. The largest values for $\Delta A$, 1.54 for the line and 1.65 for the small cross, correspond to the shapes with the shortest path lengths. The relationship is not inversely proportional, however, as we can see that the rose has the highest path length, but not the smallest $\Delta A$. Looking at the map of the rose (Fig. \ref{fig_results_shapes}, right), we can identify this highest $\Delta A$ is in the center, while the petals of the rose have a lower intensity. In contrast, the large cross appears to produce no sensations in the center, even though the focal point passes through that position. This highlights the importance of testing the mid-air haptic sensations, as they might not produce the desired effect for all paths.

\begin{figure*}[ht]
	\centering
	\begin{tabular}[b]{@{}c@{\hspace{6pt}}c@{\hspace{6pt}}c@{\hspace{6pt}}c@{\hspace{6pt}}c@{\hspace{6pt}}c@{}}
		\frame{\includegraphics[width=0.13\linewidth,trim={45 40 35 40},clip]{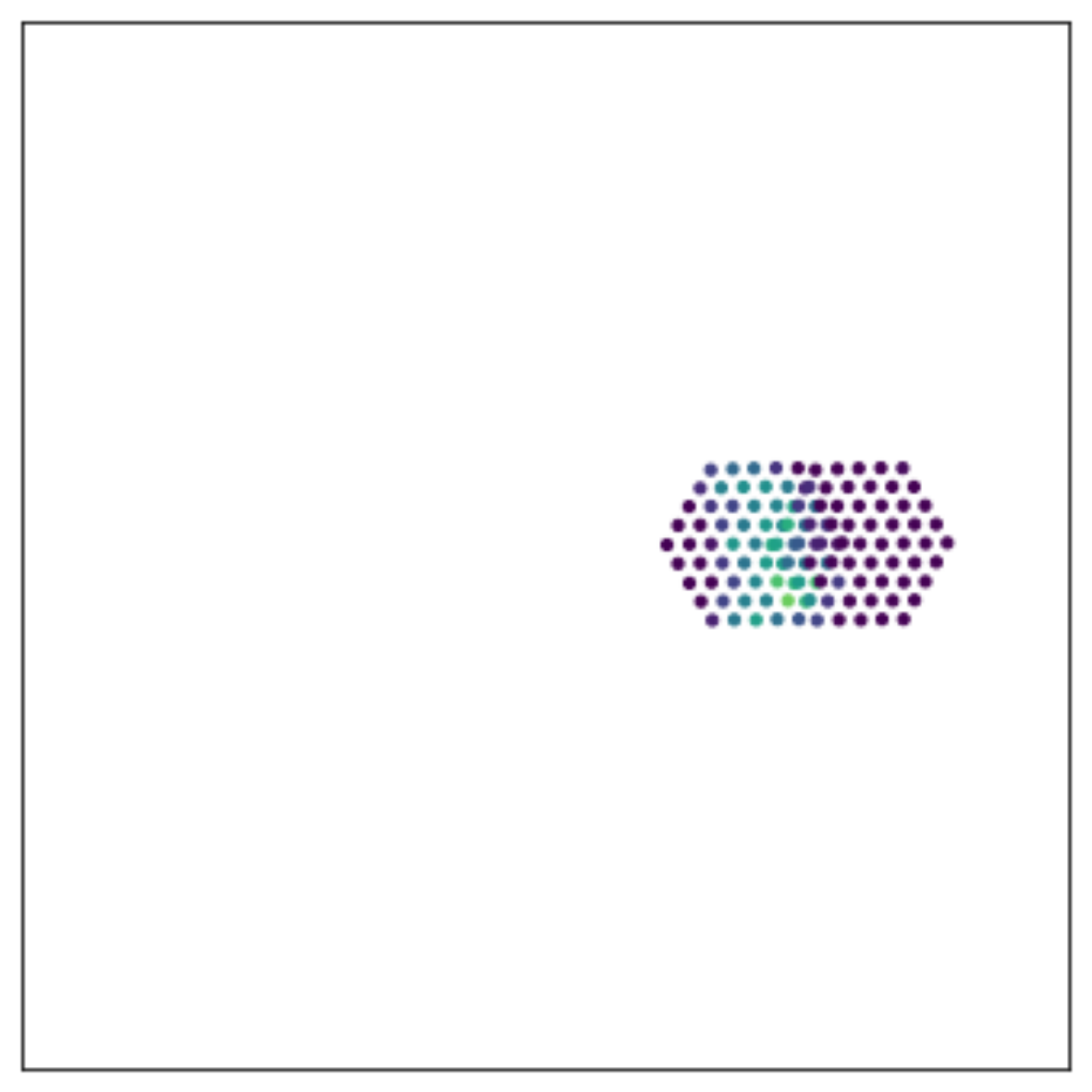}} &
		\frame{\includegraphics[width=0.13\linewidth,trim={45 40 35 40},clip]{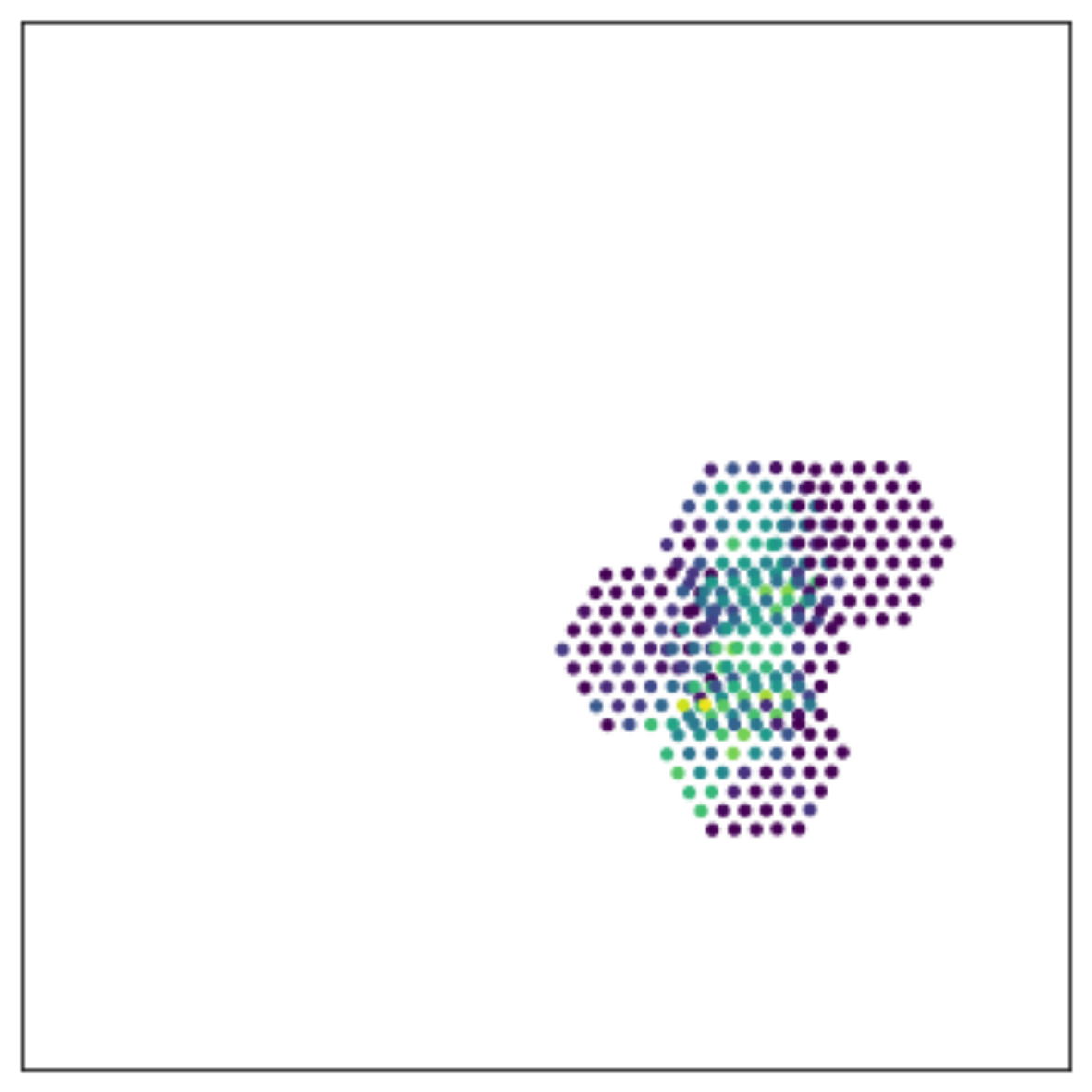}} &
		\frame{\includegraphics[width=0.13\linewidth,trim={45 40 35 40},clip]{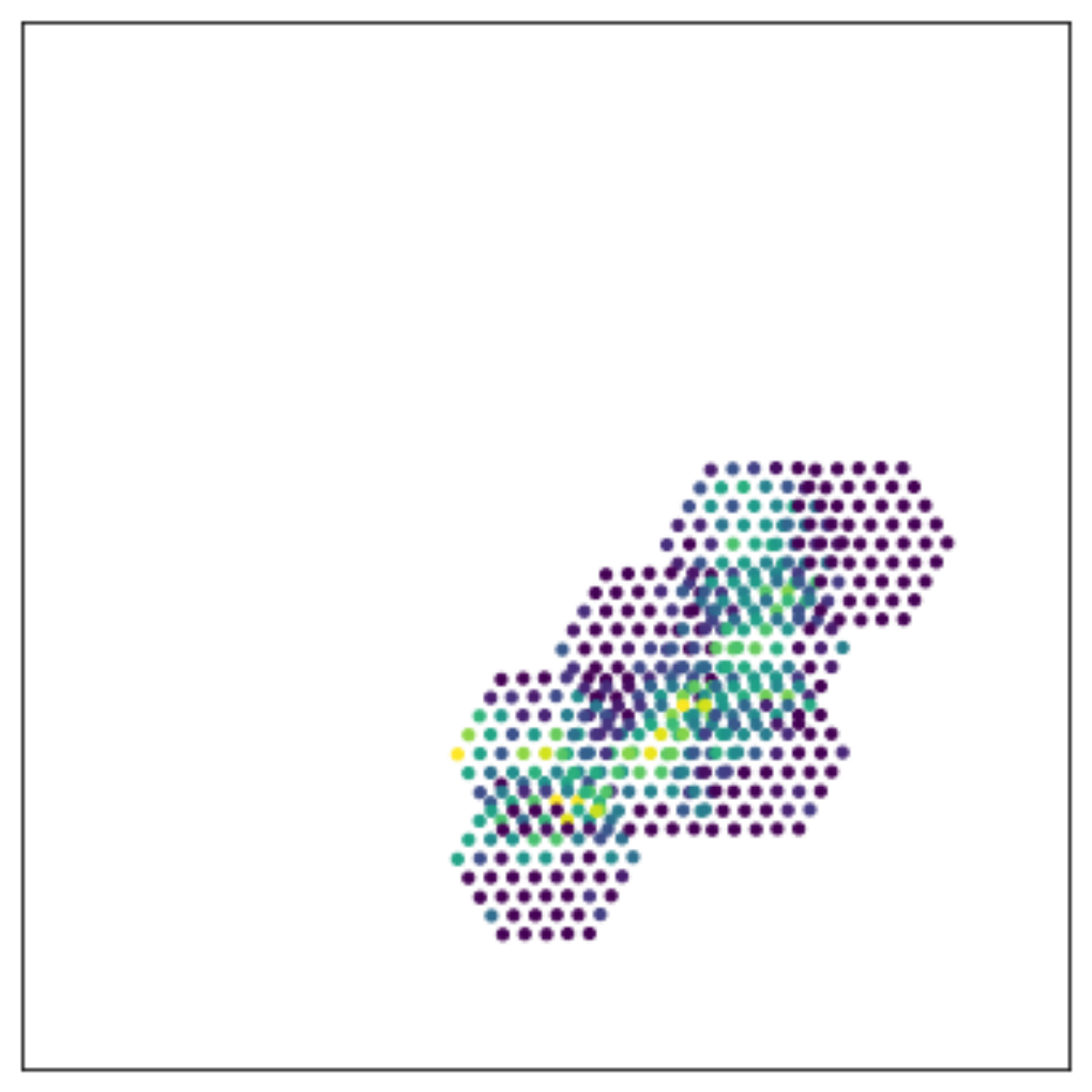}} &
		\frame{\includegraphics[width=0.13\linewidth,trim={45 40 35 40},clip]{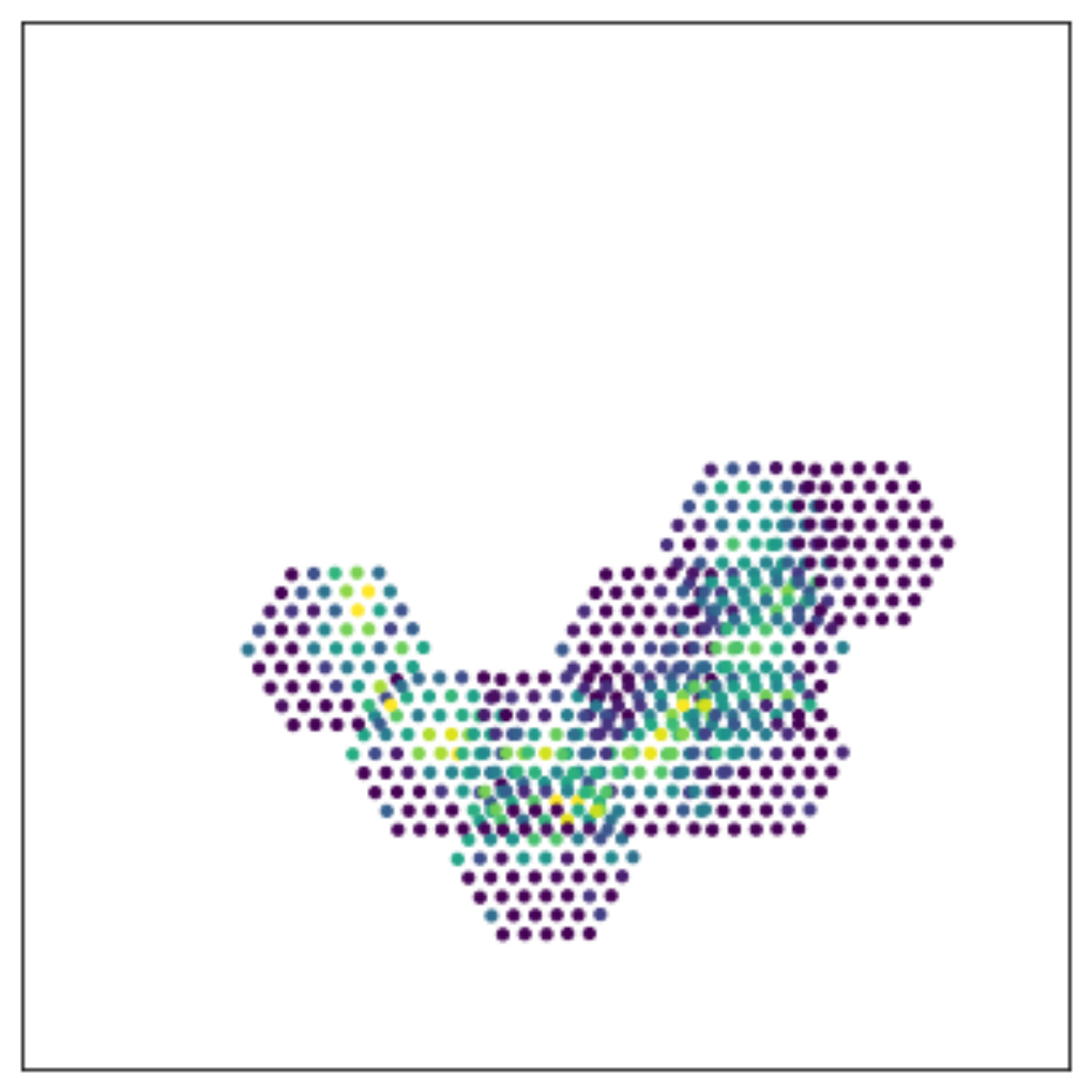}} &
		\frame{\includegraphics[width=0.13\linewidth,trim={45 40 35 40},clip]{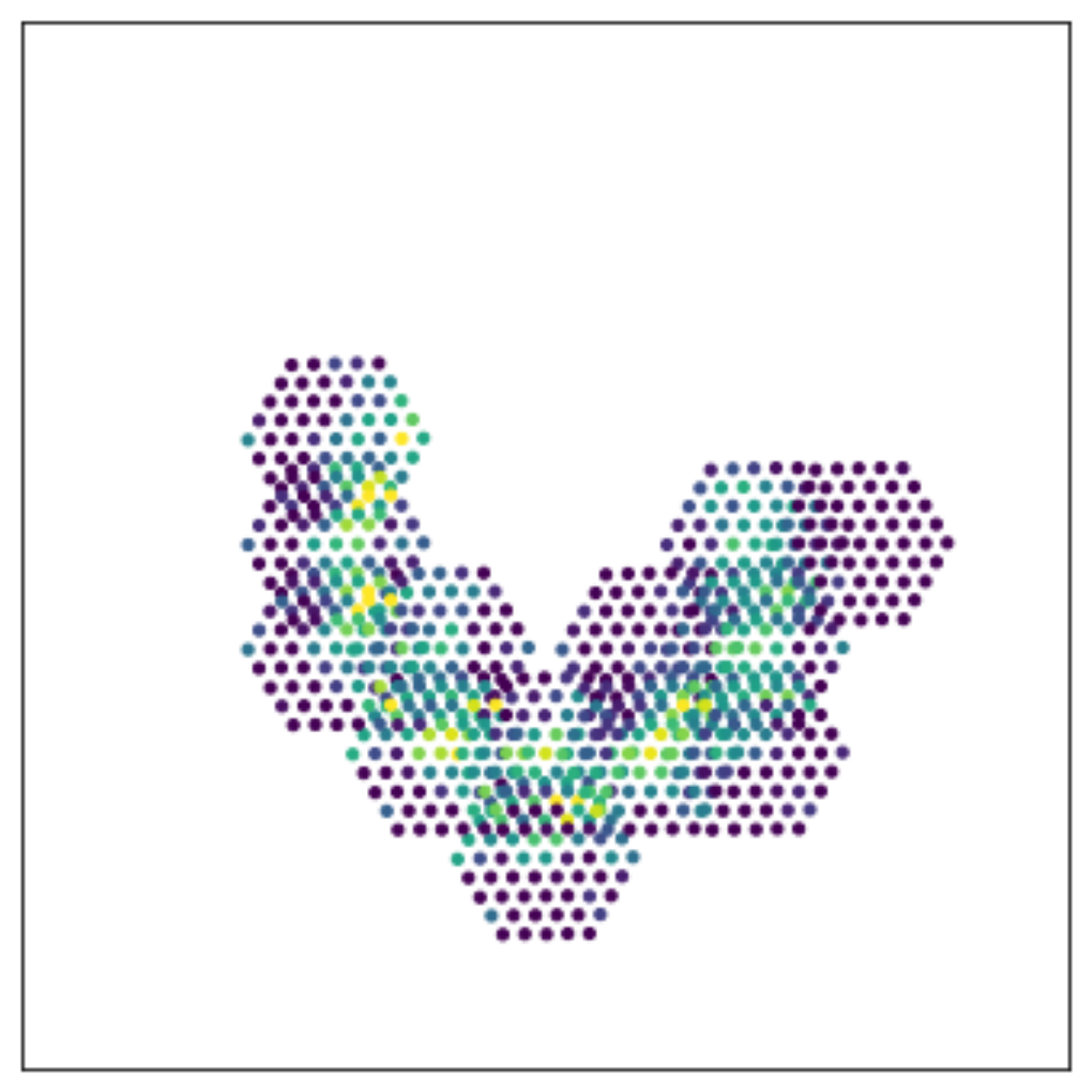}} &
		\frame{\includegraphics[width=0.13\linewidth,trim={45 40 35 40},clip]{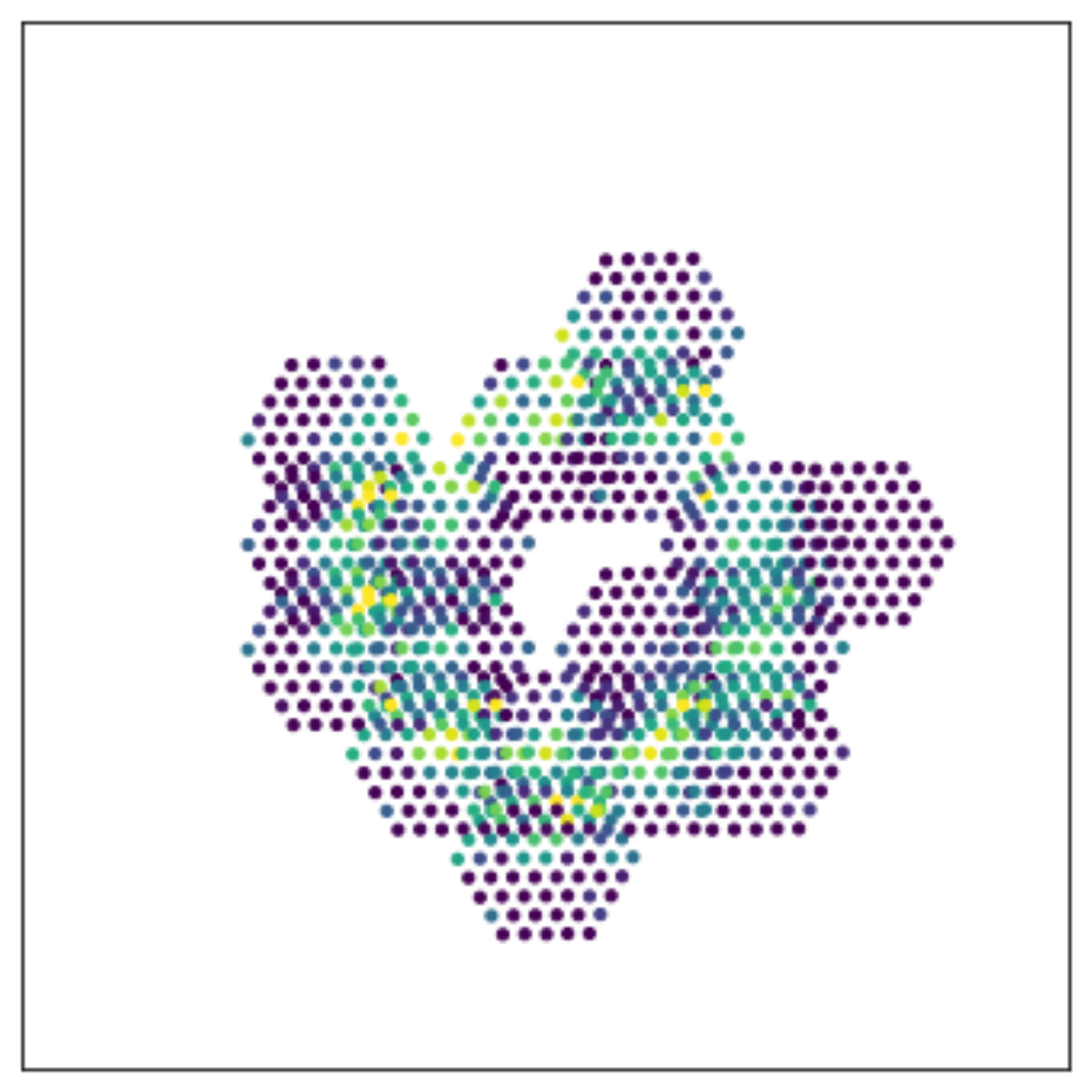}}
		\\
    	\frame{\includegraphics[width=0.13\linewidth,trim={45 40 35 40},clip]{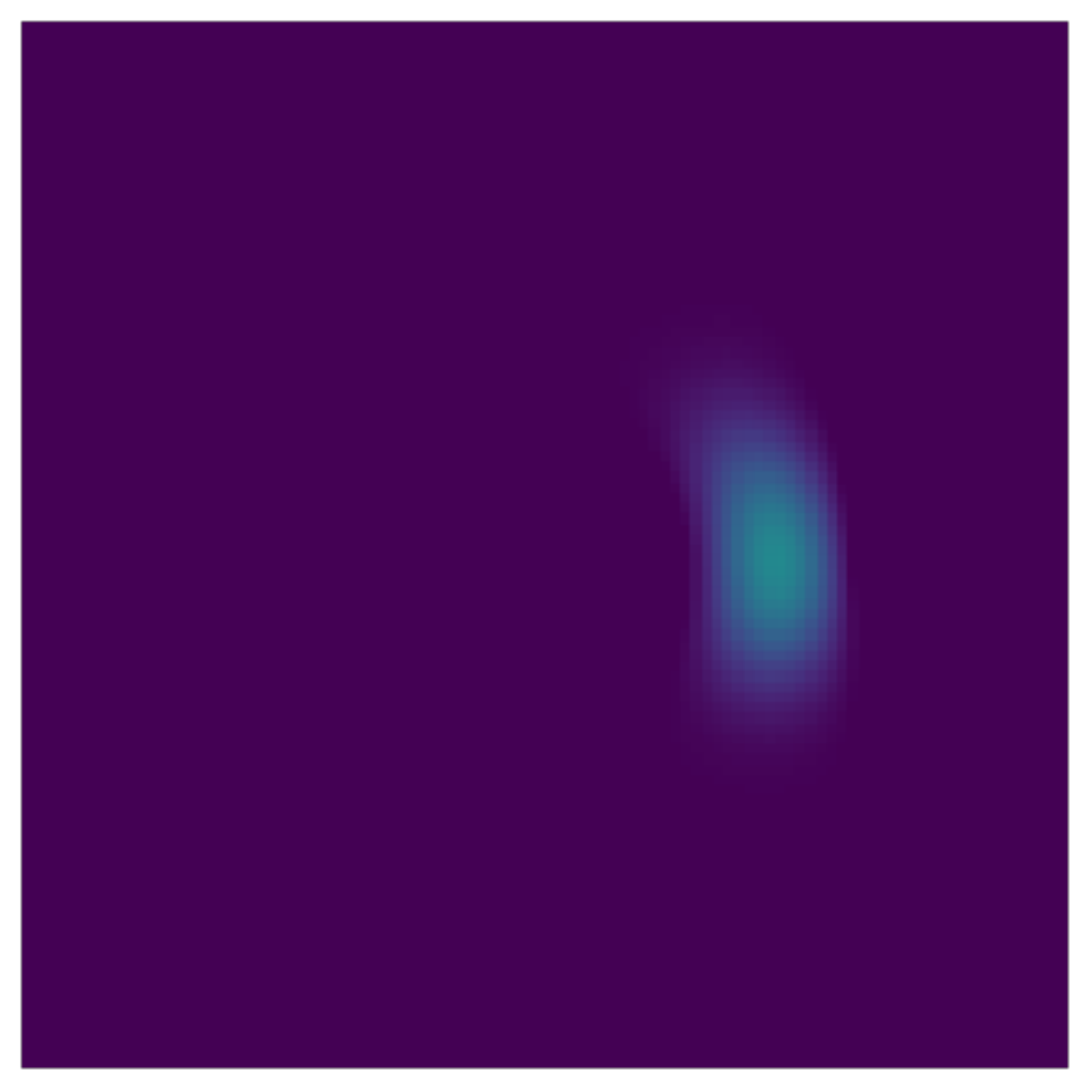}} &
		\frame{\includegraphics[width=0.13\linewidth,trim={45 40 35 40},clip]{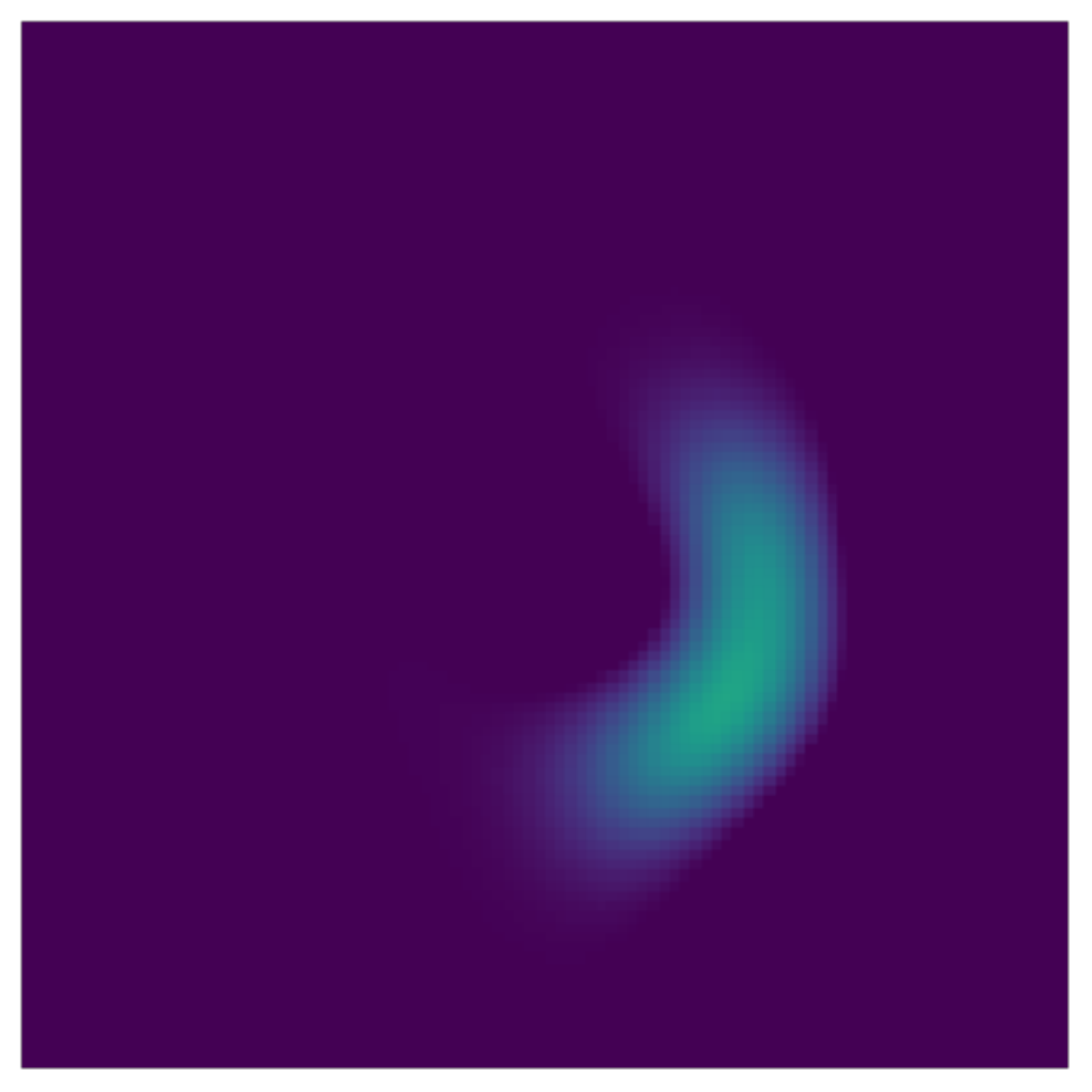}} &
		\frame{\includegraphics[width=0.13\linewidth,trim={45 40 35 40},clip]{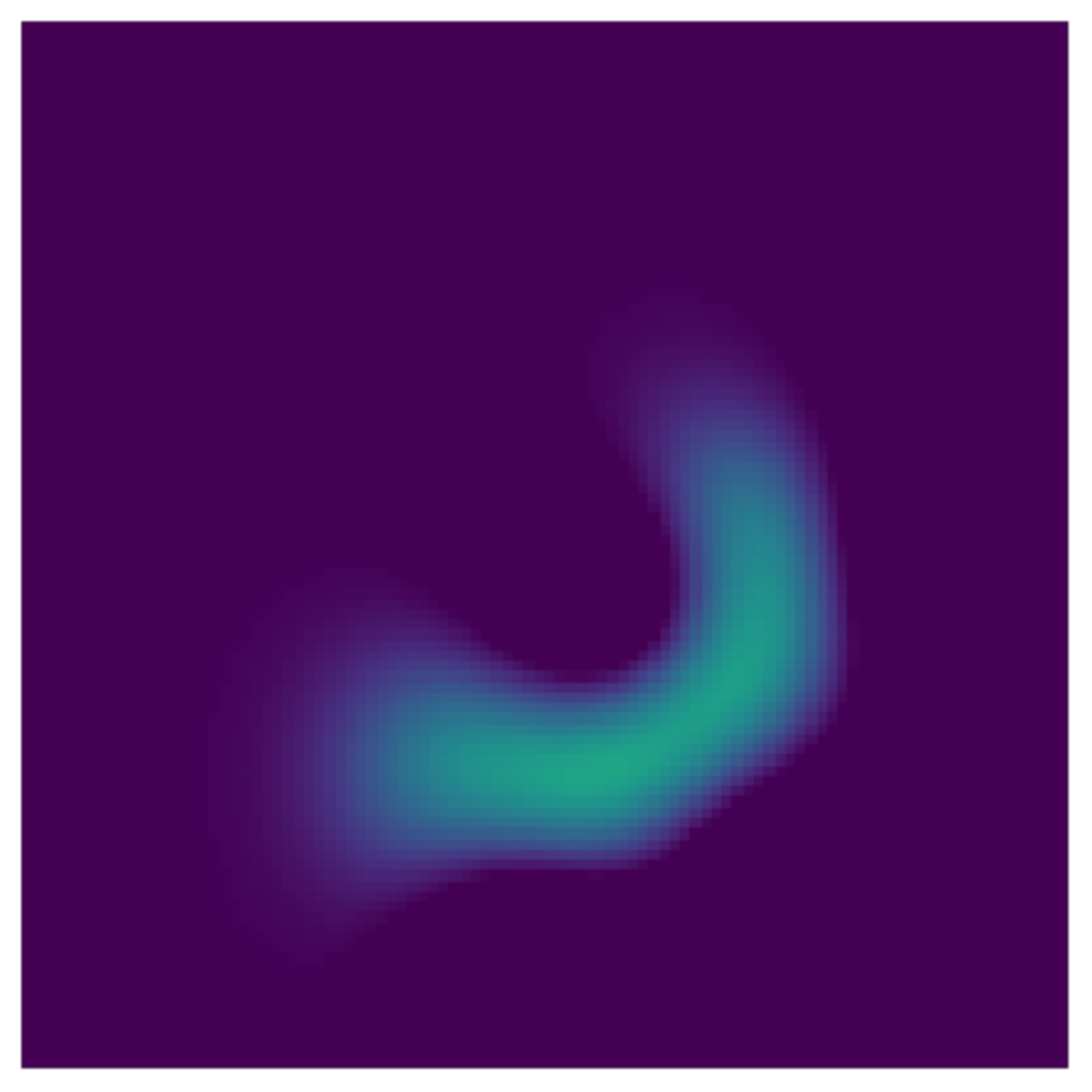}} &
		\frame{\includegraphics[width=0.13\linewidth,trim={45 40 35 40},clip]{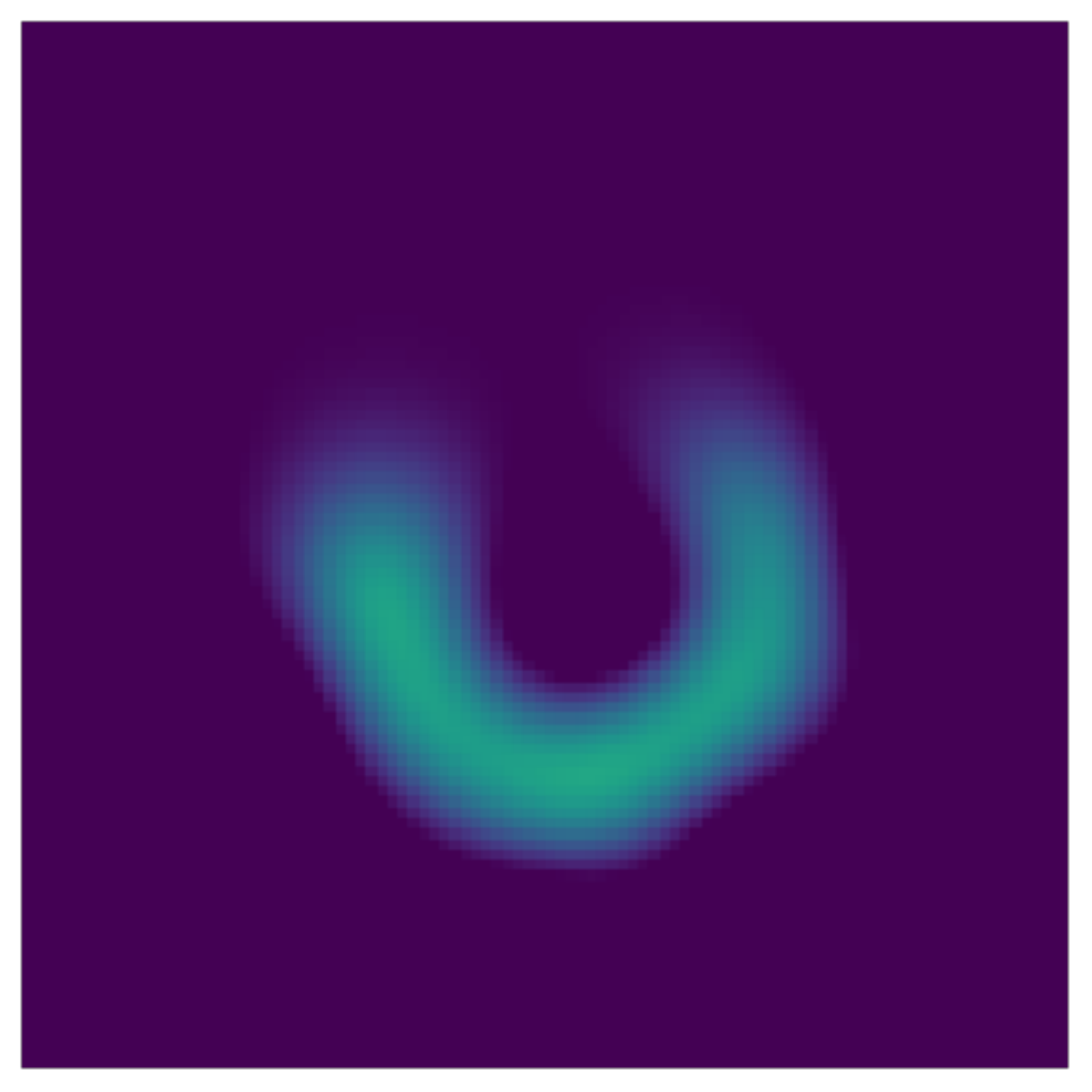}} &
		\frame{\includegraphics[width=0.13\linewidth,trim={45 40 35 40},clip]{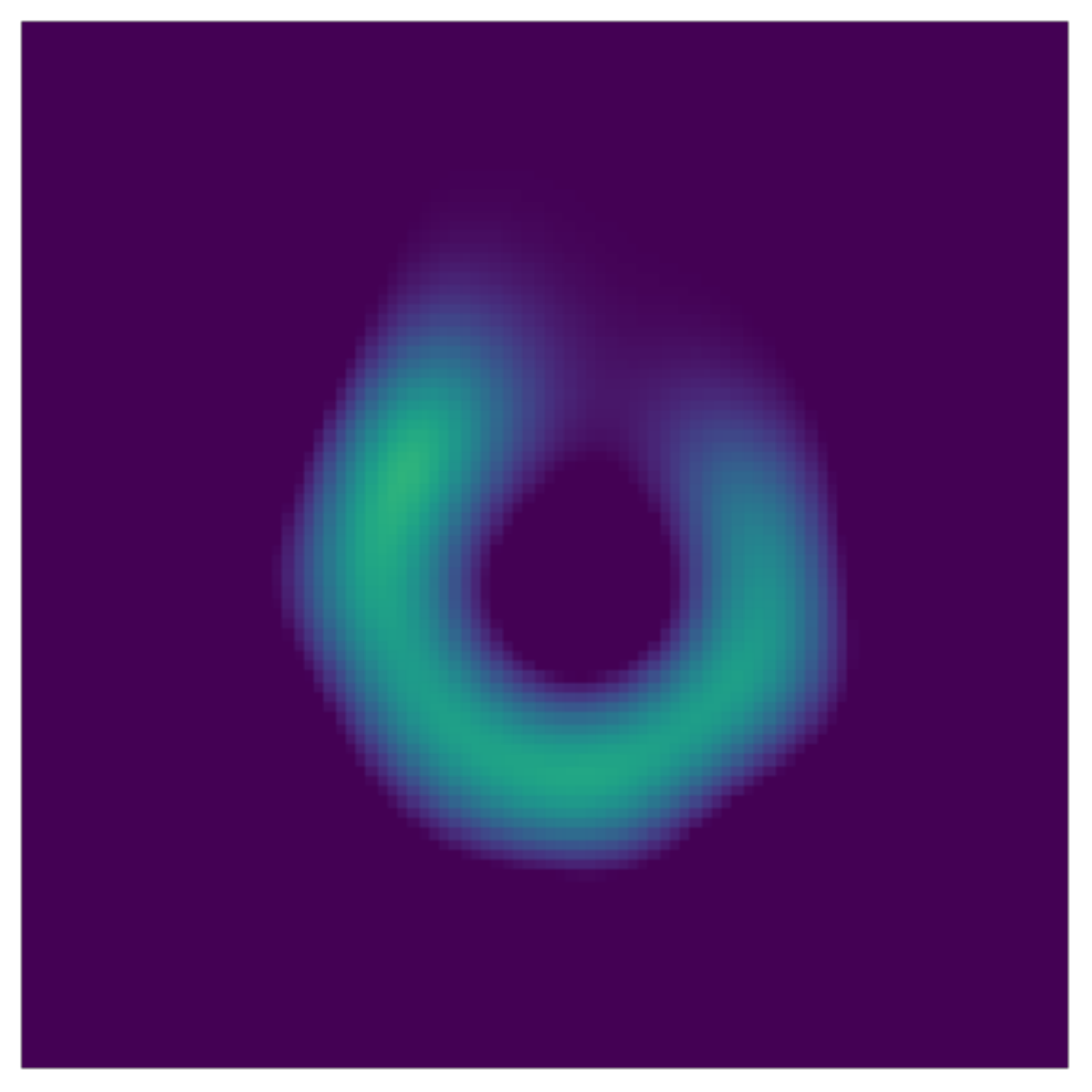}} &
		\frame{\includegraphics[width=0.13\linewidth,trim={45 40 35 40},clip]{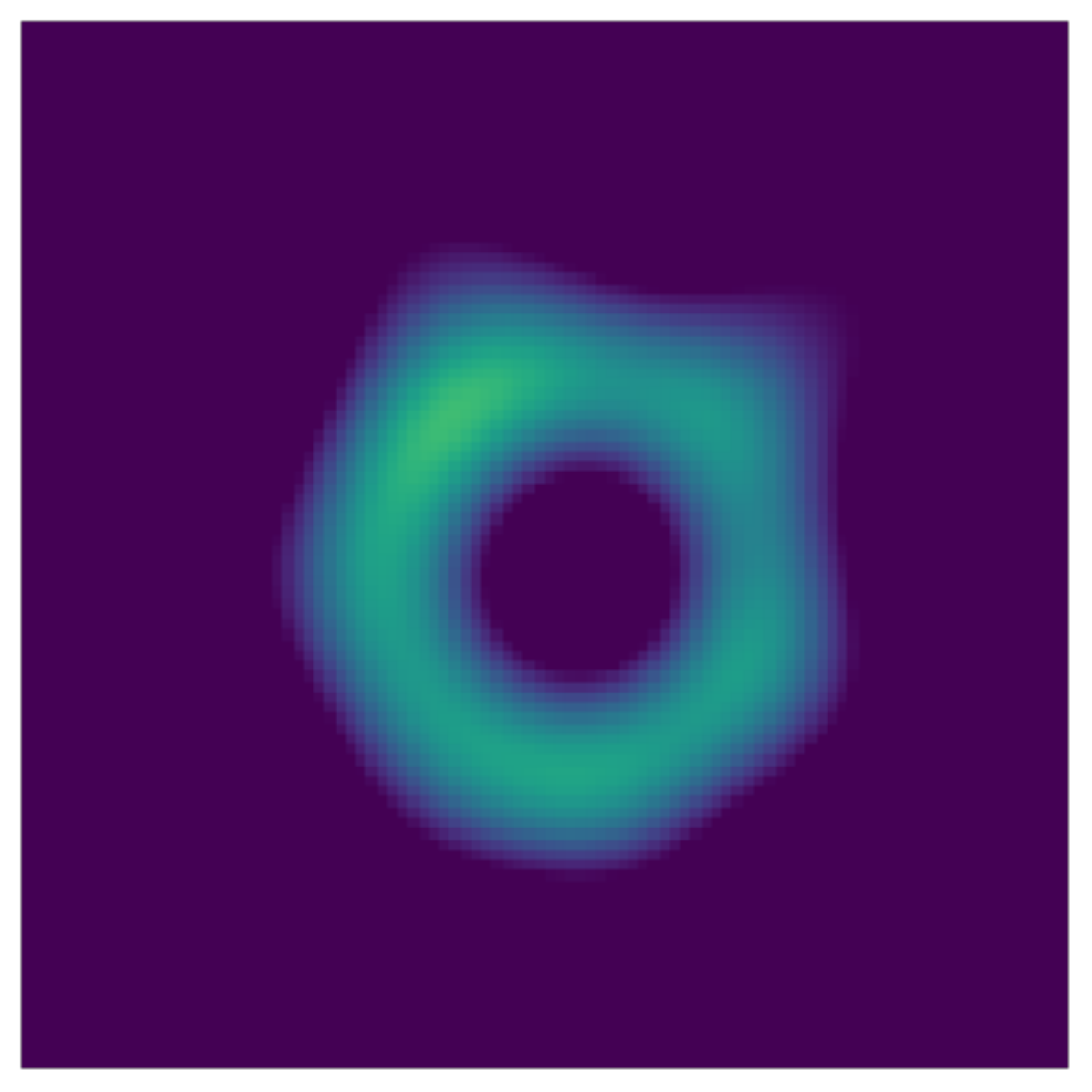}}
		\\ 
		\\ 
		\frame{\includegraphics[width=0.13\linewidth,trim={45 40 35 40},clip]{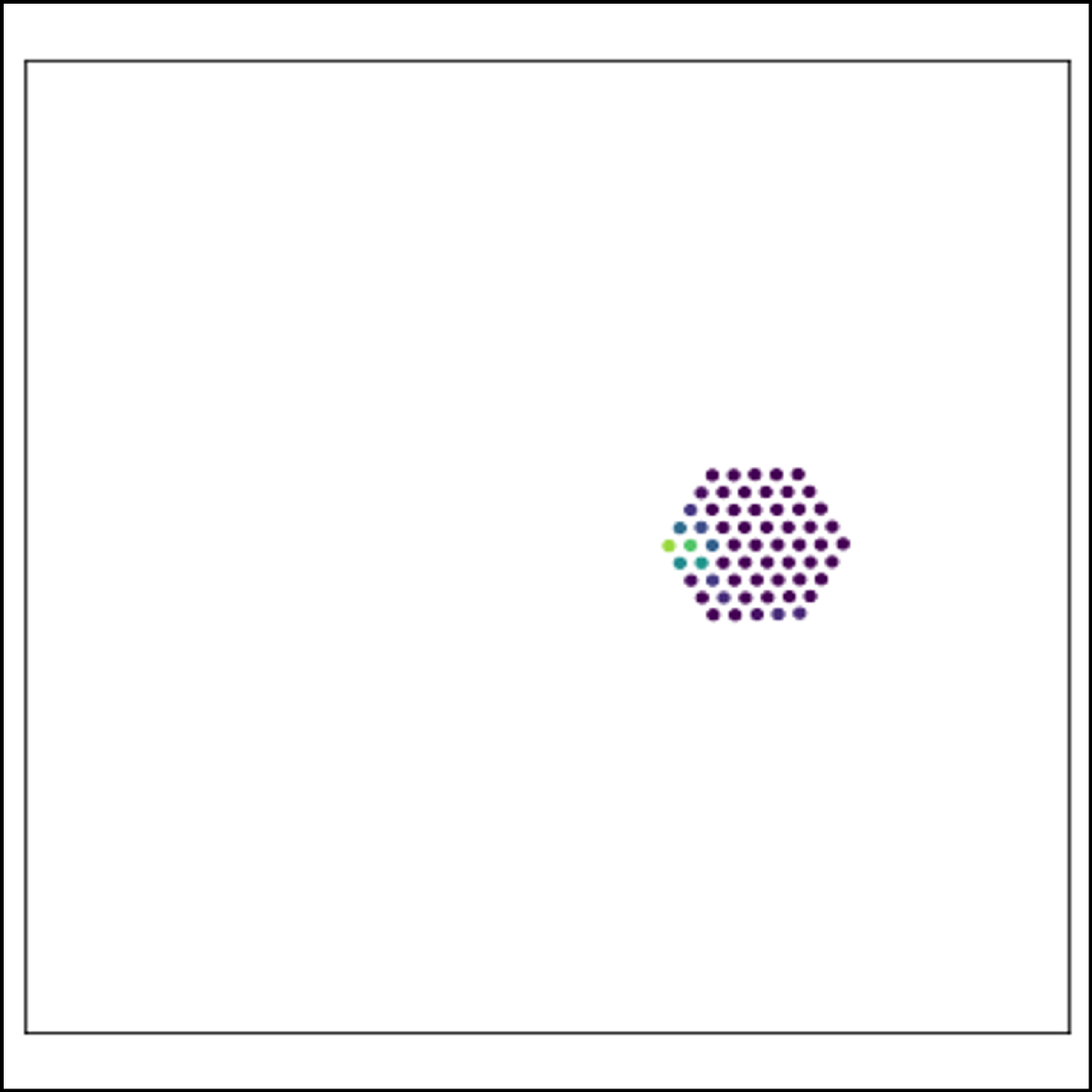}} &
		\frame{\includegraphics[width=0.13\linewidth,trim={45 40 35 40},clip]{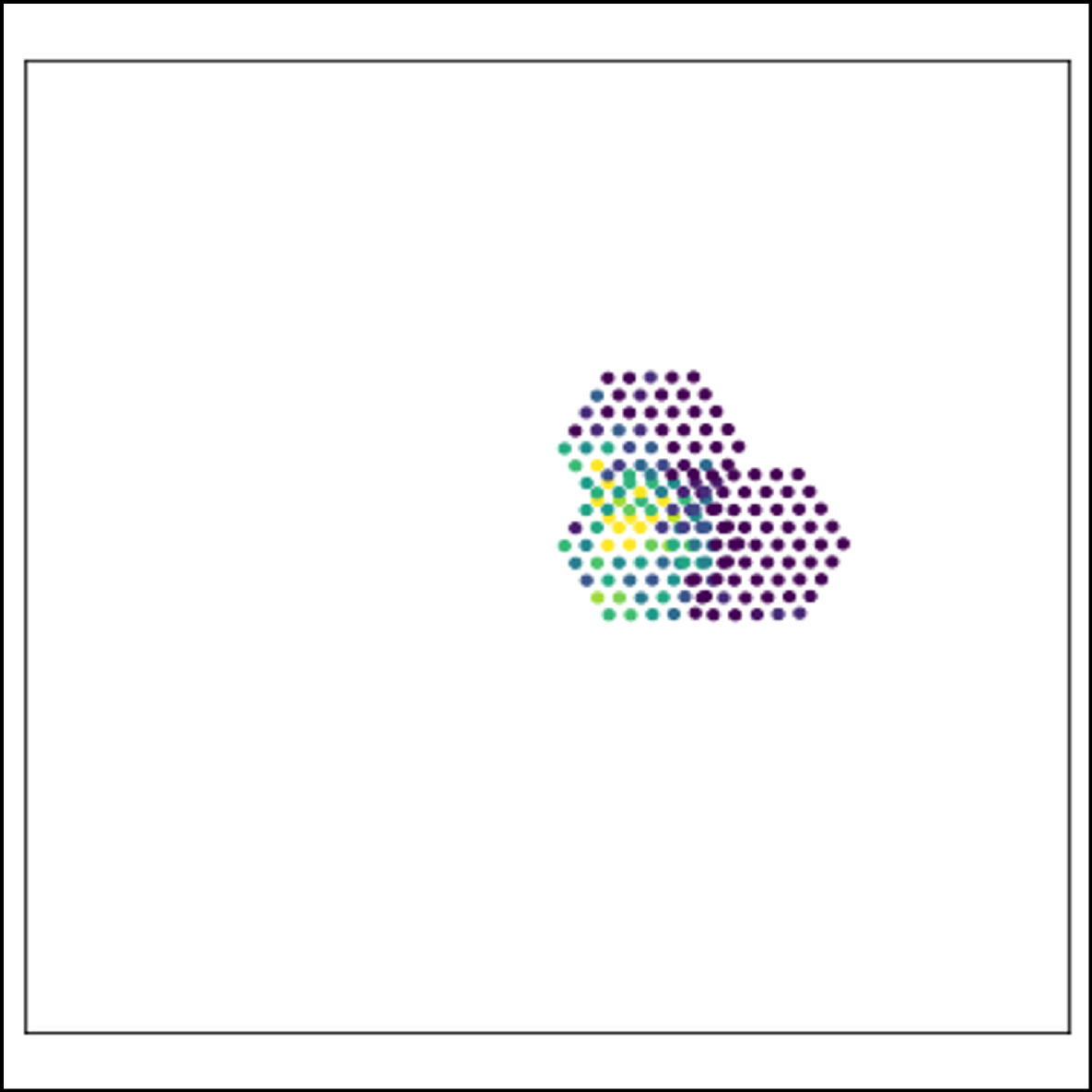}} &
		\frame{\includegraphics[width=0.13\linewidth,trim={45 40 35 40},clip]{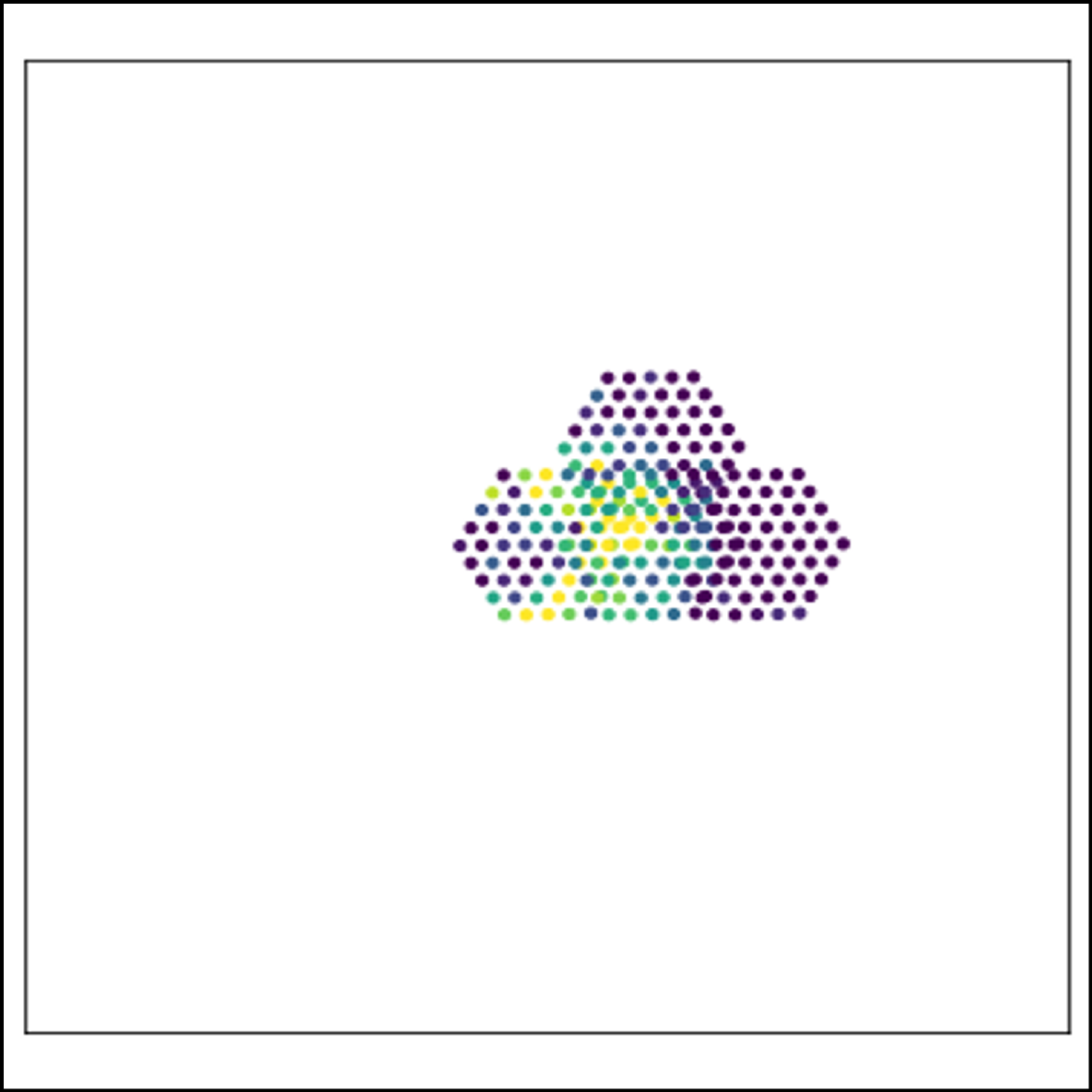}} &
		\frame{\includegraphics[width=0.13\linewidth,trim={45 40 35 40},clip]{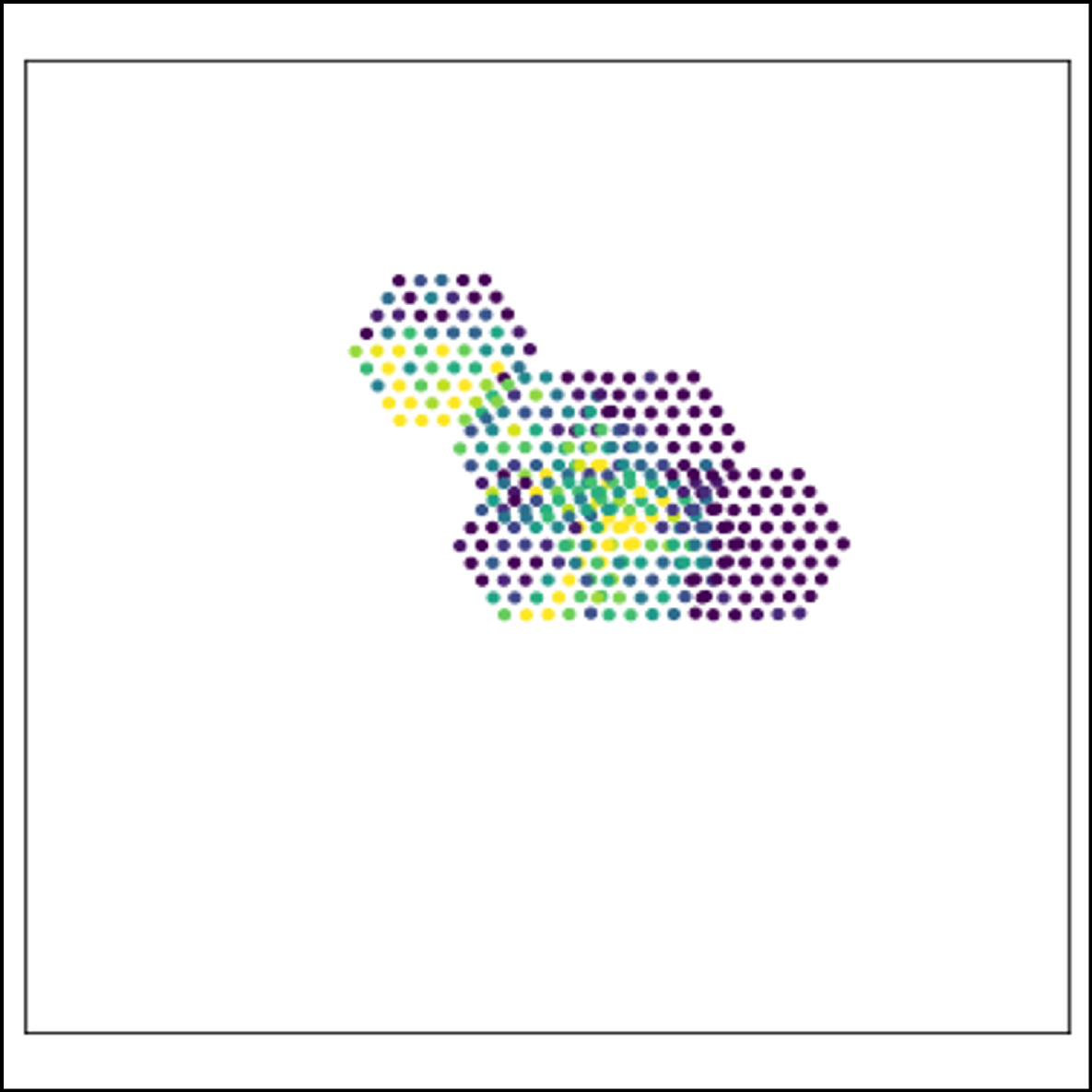}} &
		\frame{\includegraphics[width=0.13\linewidth,trim={45 40 35 40},clip]{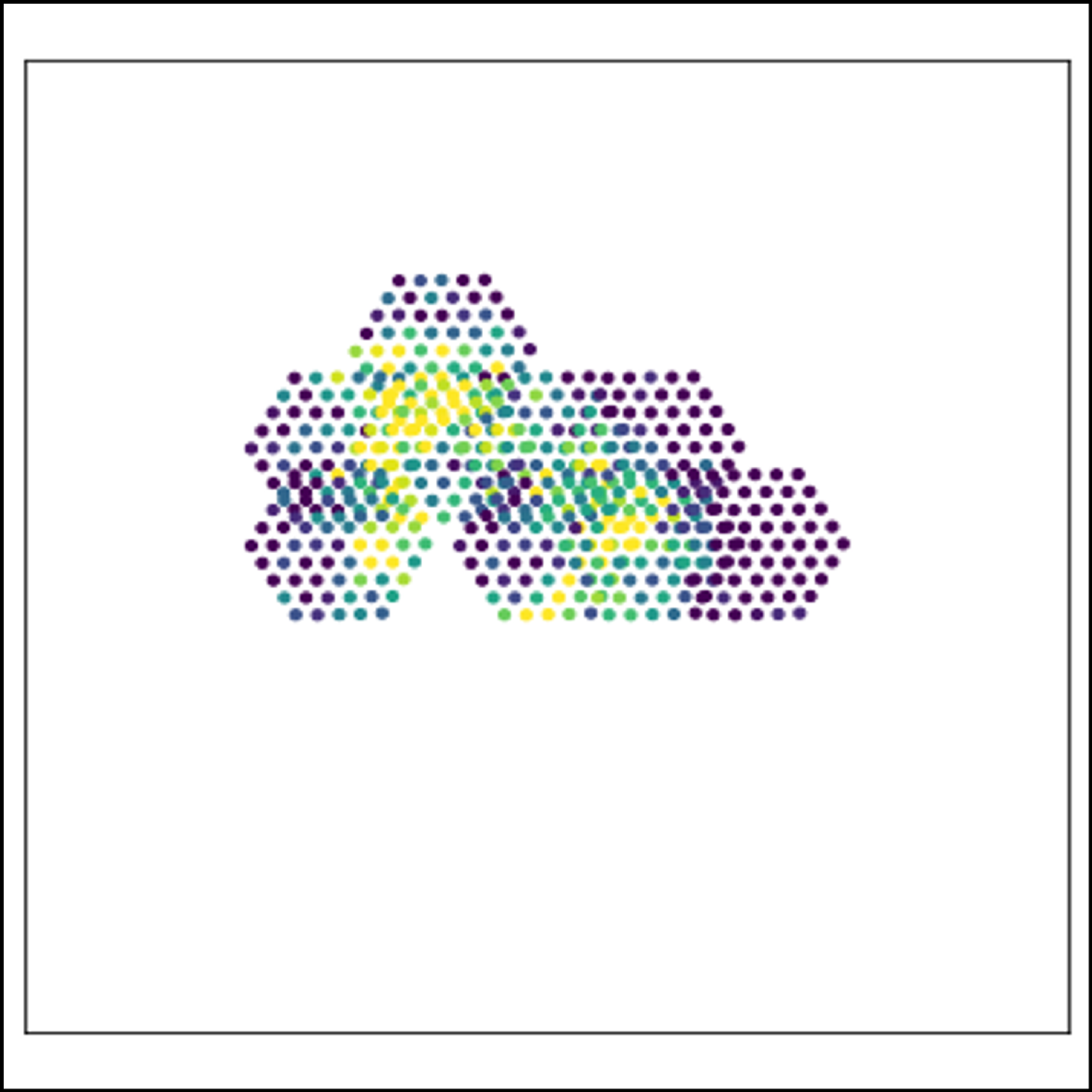}} &
		\frame{\includegraphics[width=0.13\linewidth,trim={45 40 35 40},clip]{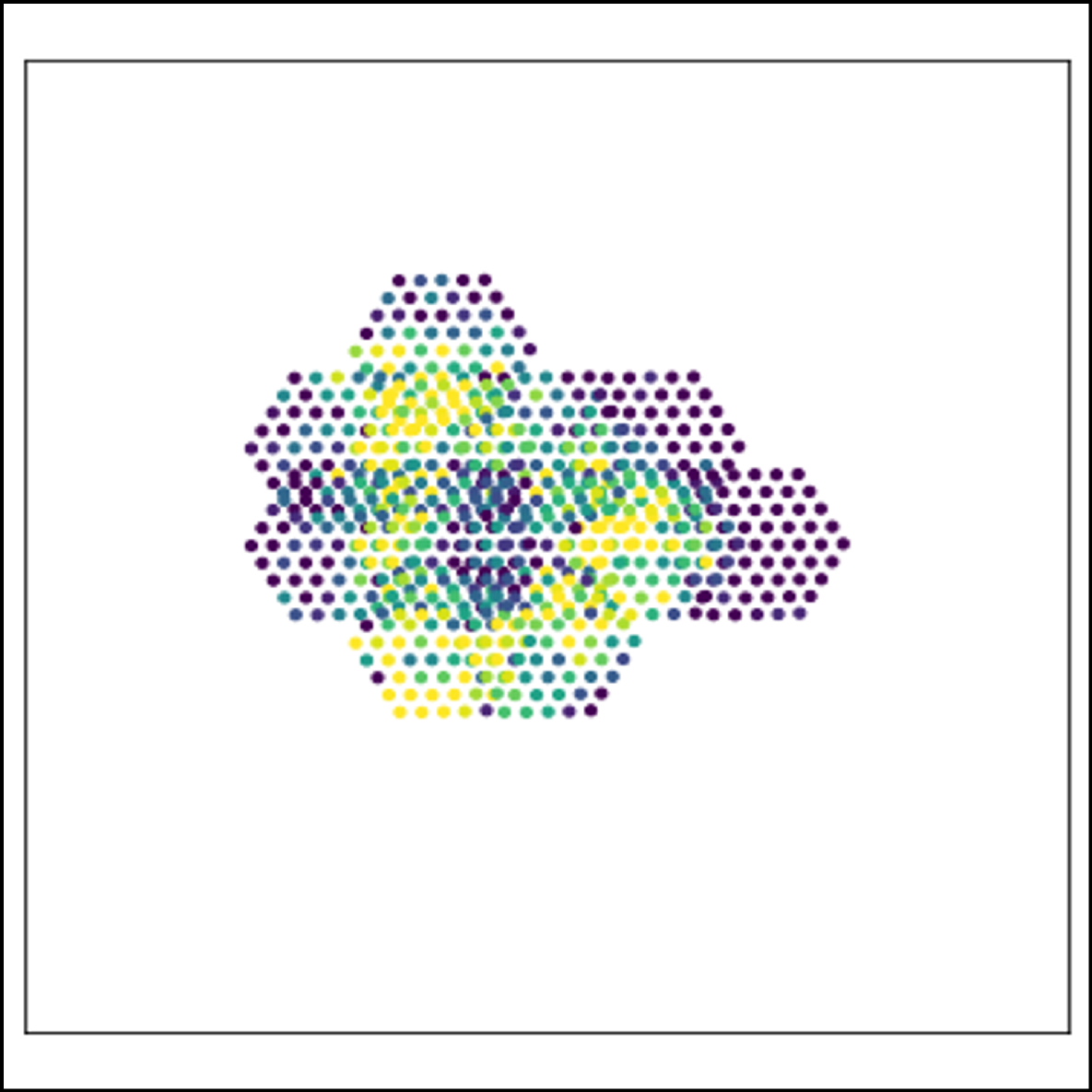}}
		\\
    	\frame{\includegraphics[width=0.13\linewidth,trim={45 40 35 40},clip]{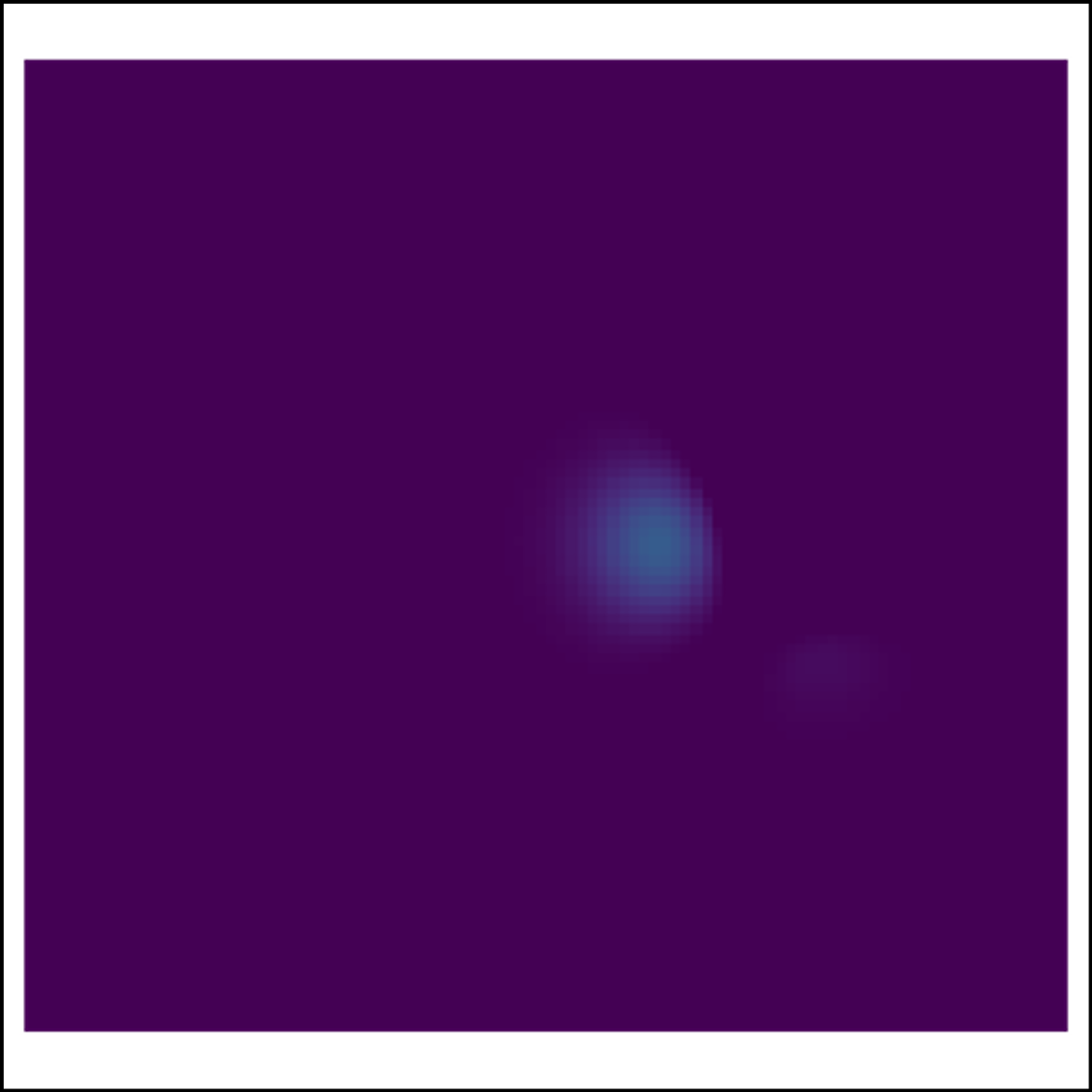}} &
		\frame{\includegraphics[width=0.13\linewidth,trim={45 40 35 40},clip]{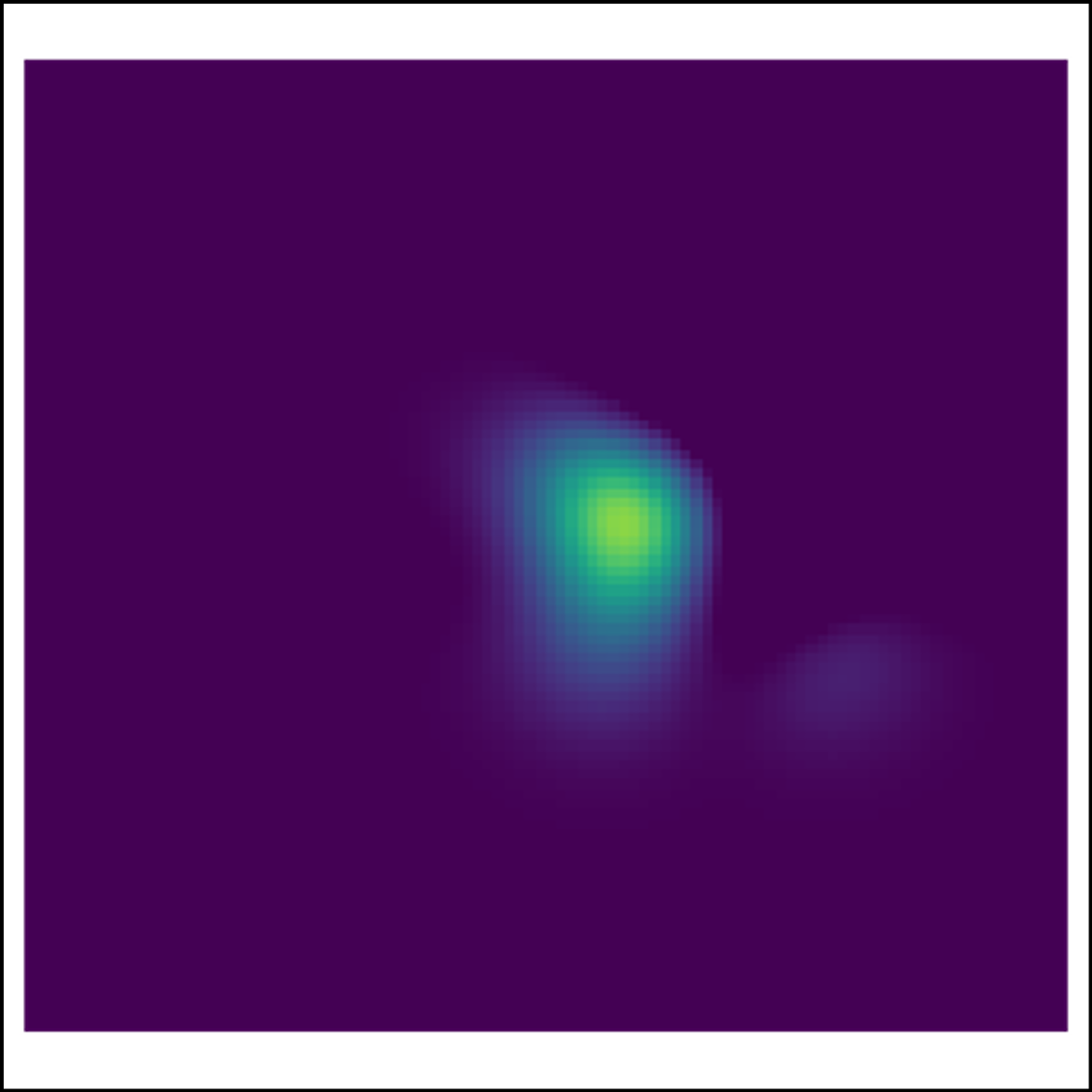}} &
		\frame{\includegraphics[width=0.13\linewidth,trim={45 40 35 40},clip]{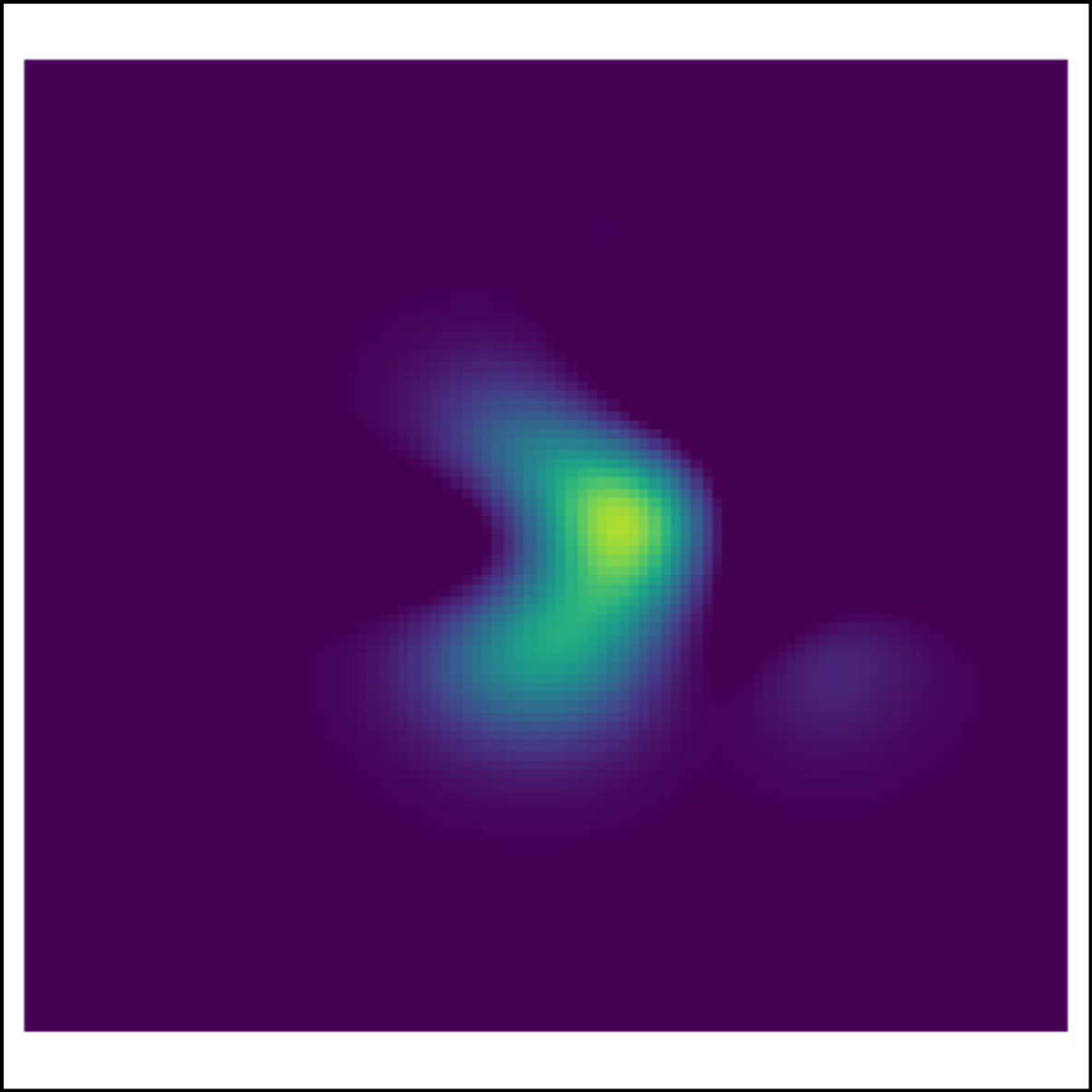}} &
		\frame{\includegraphics[width=0.13\linewidth,trim={45 40 35 40},clip]{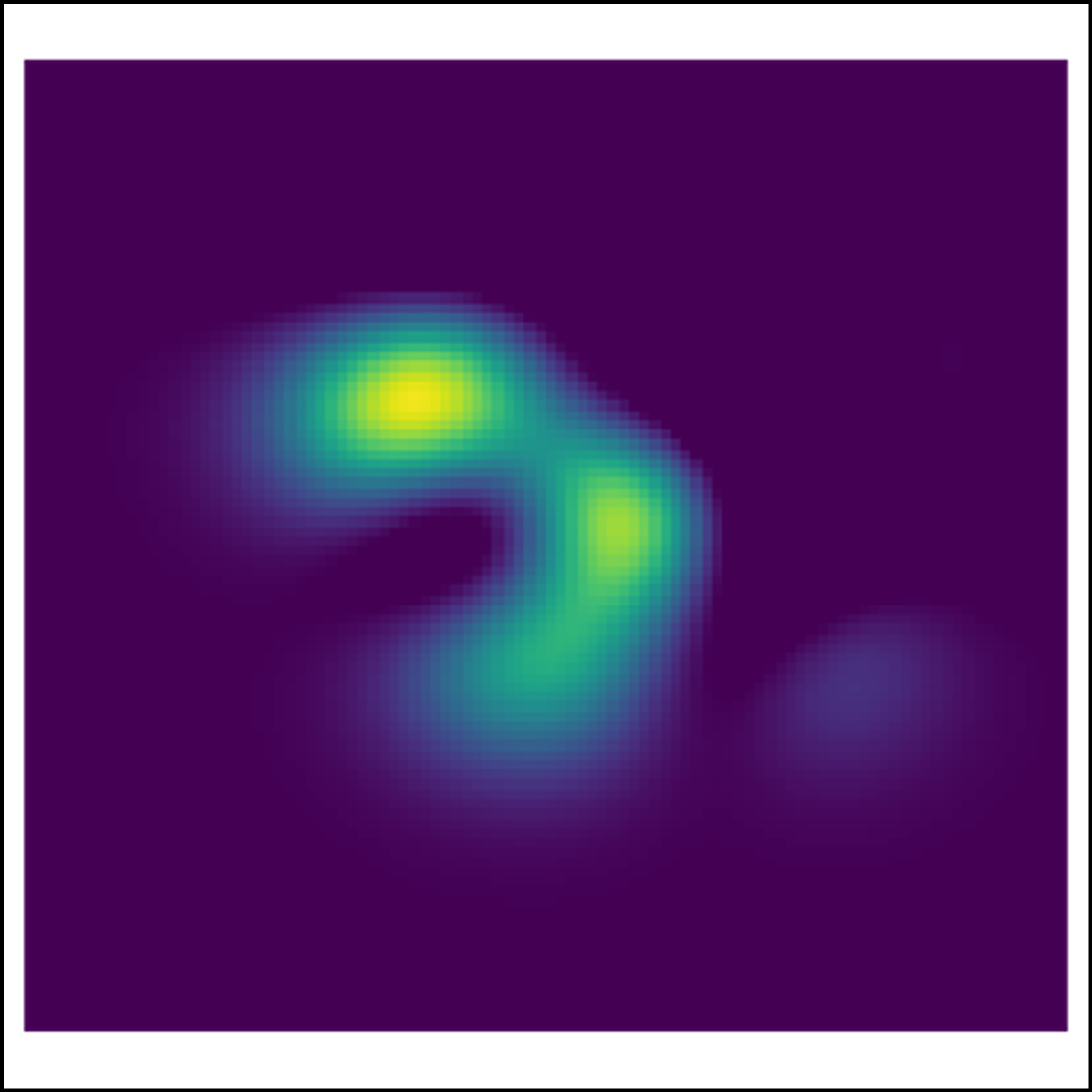}} &
		\frame{\includegraphics[width=0.13\linewidth,trim={45 40 35 40},clip]{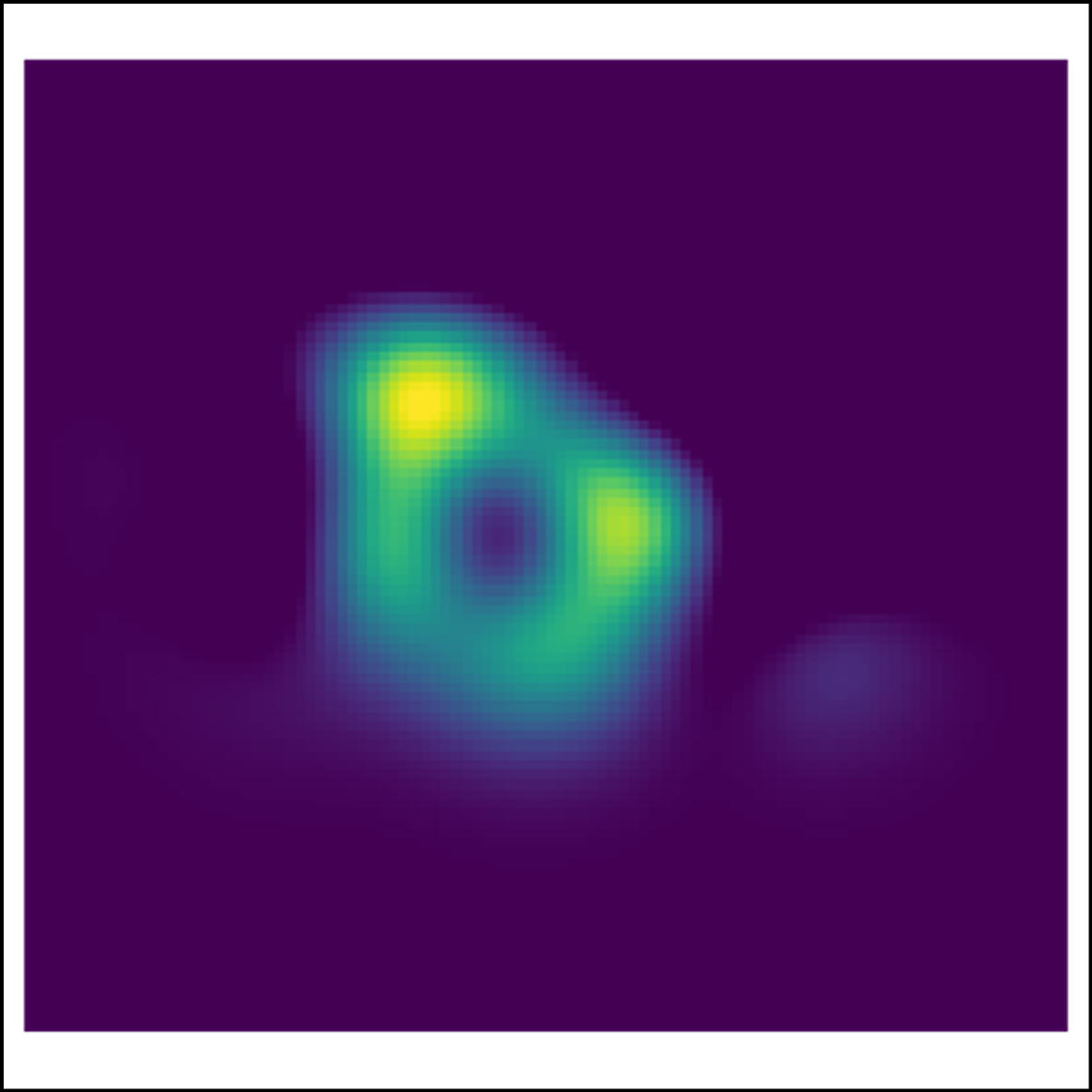}} &
		\frame{\includegraphics[width=0.13\linewidth,trim={45 40 35 40},clip]{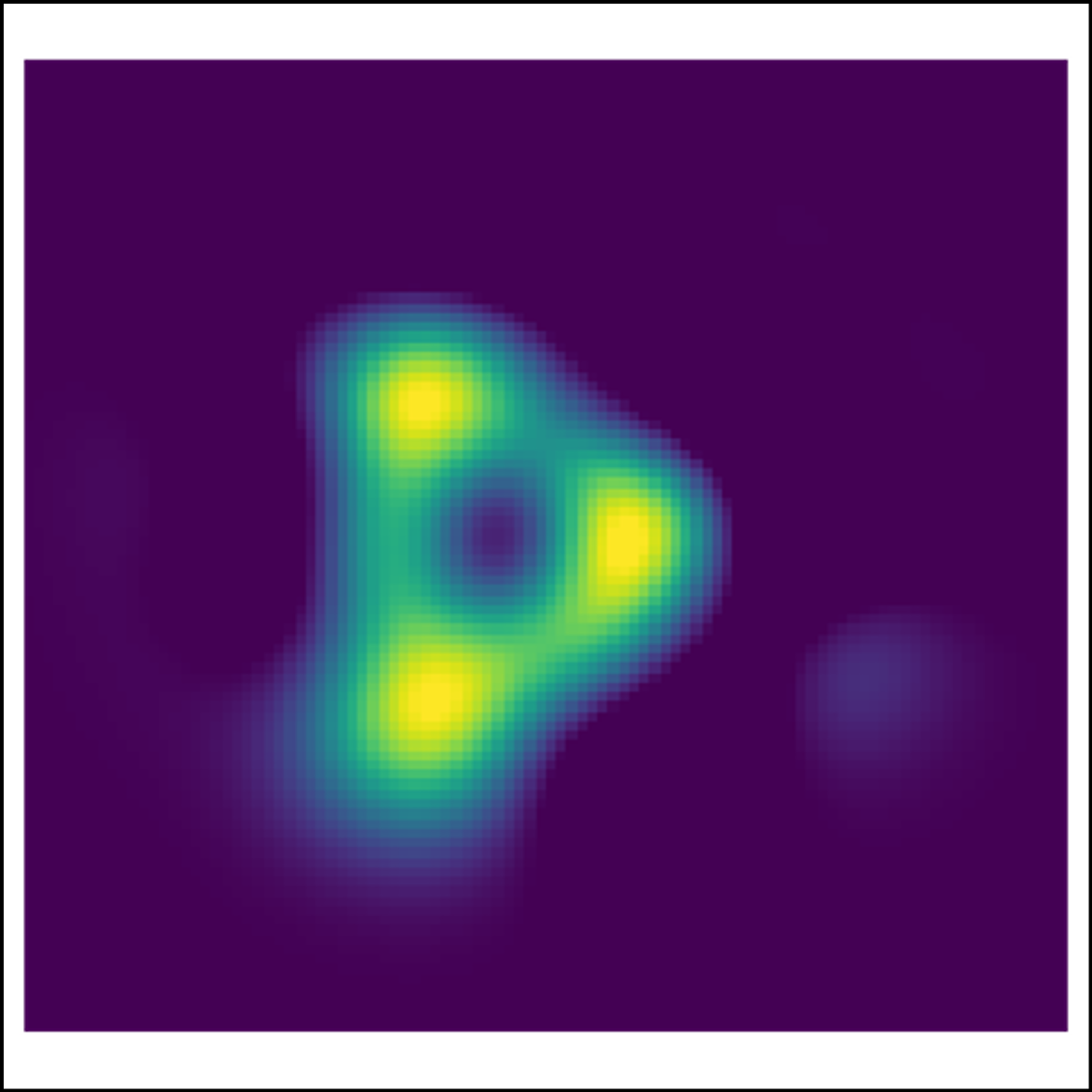}}
		\\
		\\ 
		\frame{\includegraphics[width=0.13\linewidth,trim={45 40 35 40},clip]{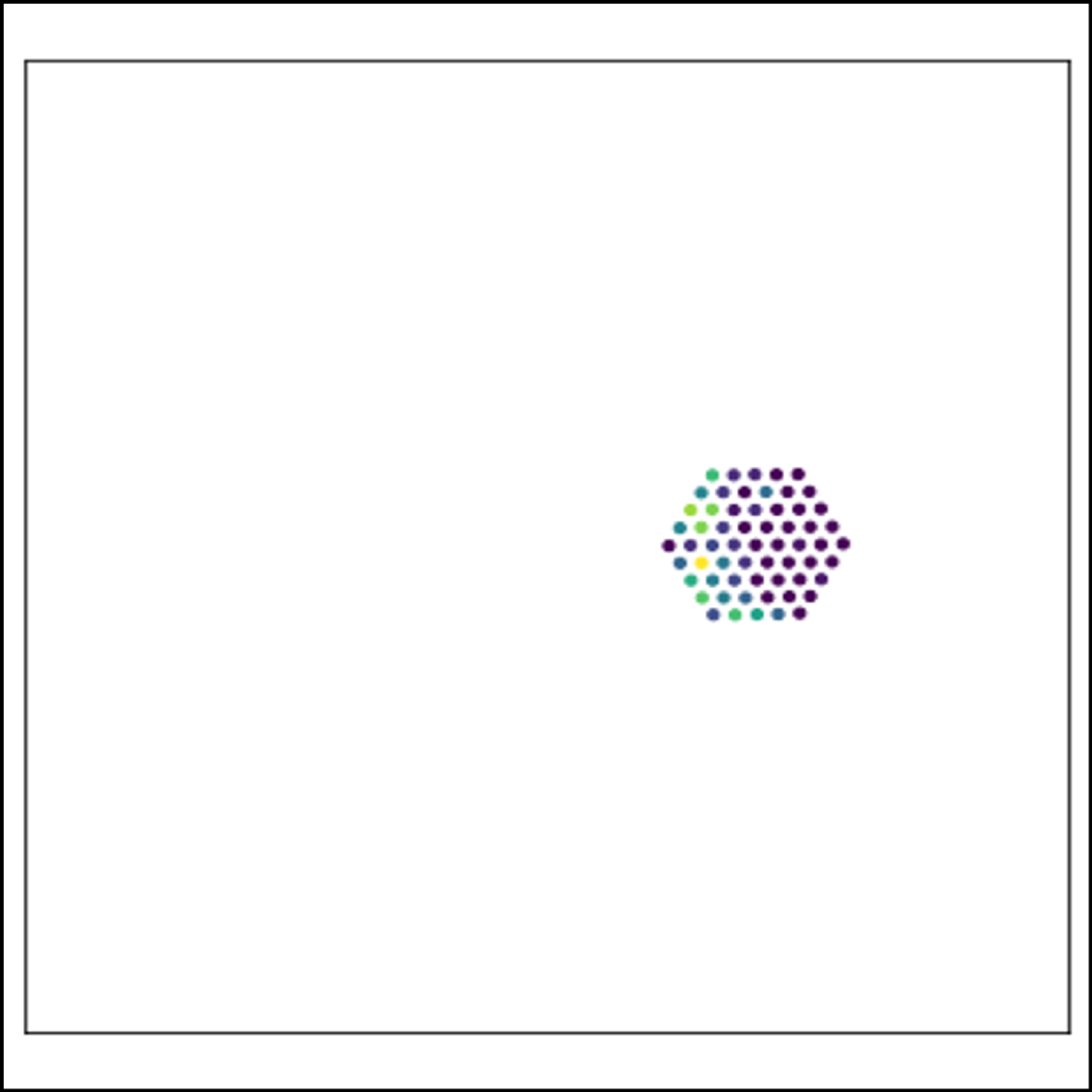}} &
		\frame{\includegraphics[width=0.13\linewidth,trim={45 40 35 40},clip]{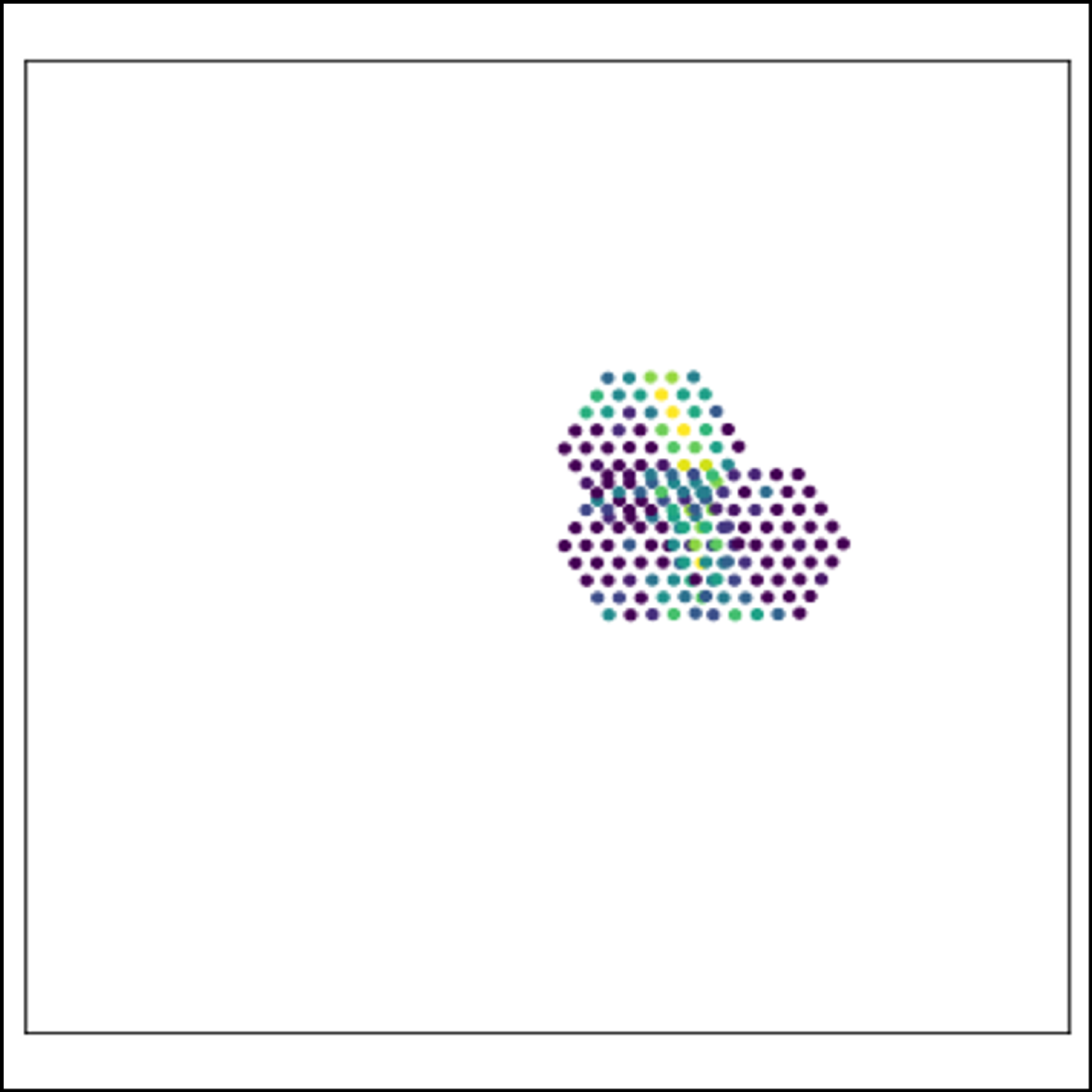}} &
		\frame{\includegraphics[width=0.13\linewidth,trim={45 40 35 40},clip]{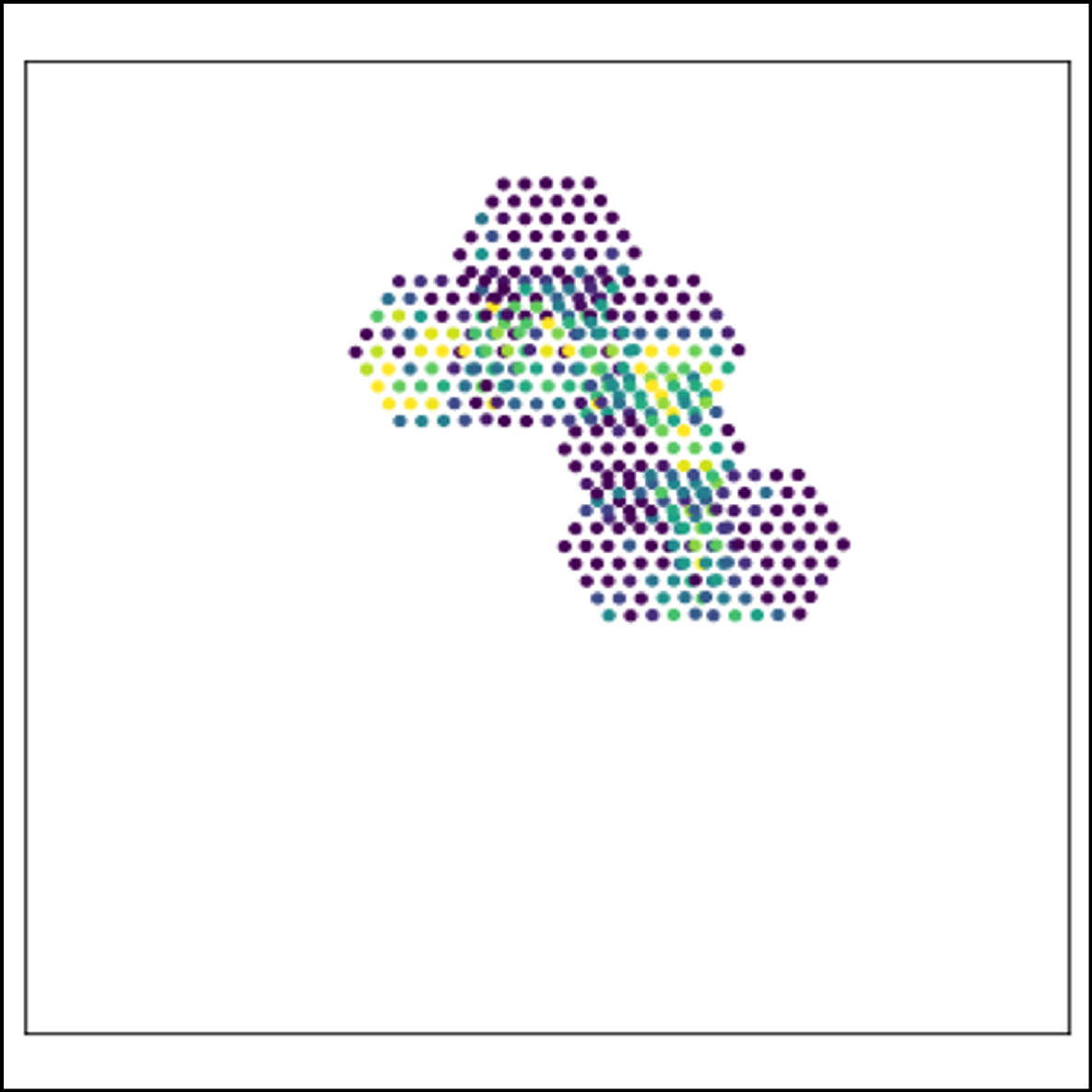}} &
		\frame{\includegraphics[width=0.13\linewidth,trim={45 40 35 40},clip]{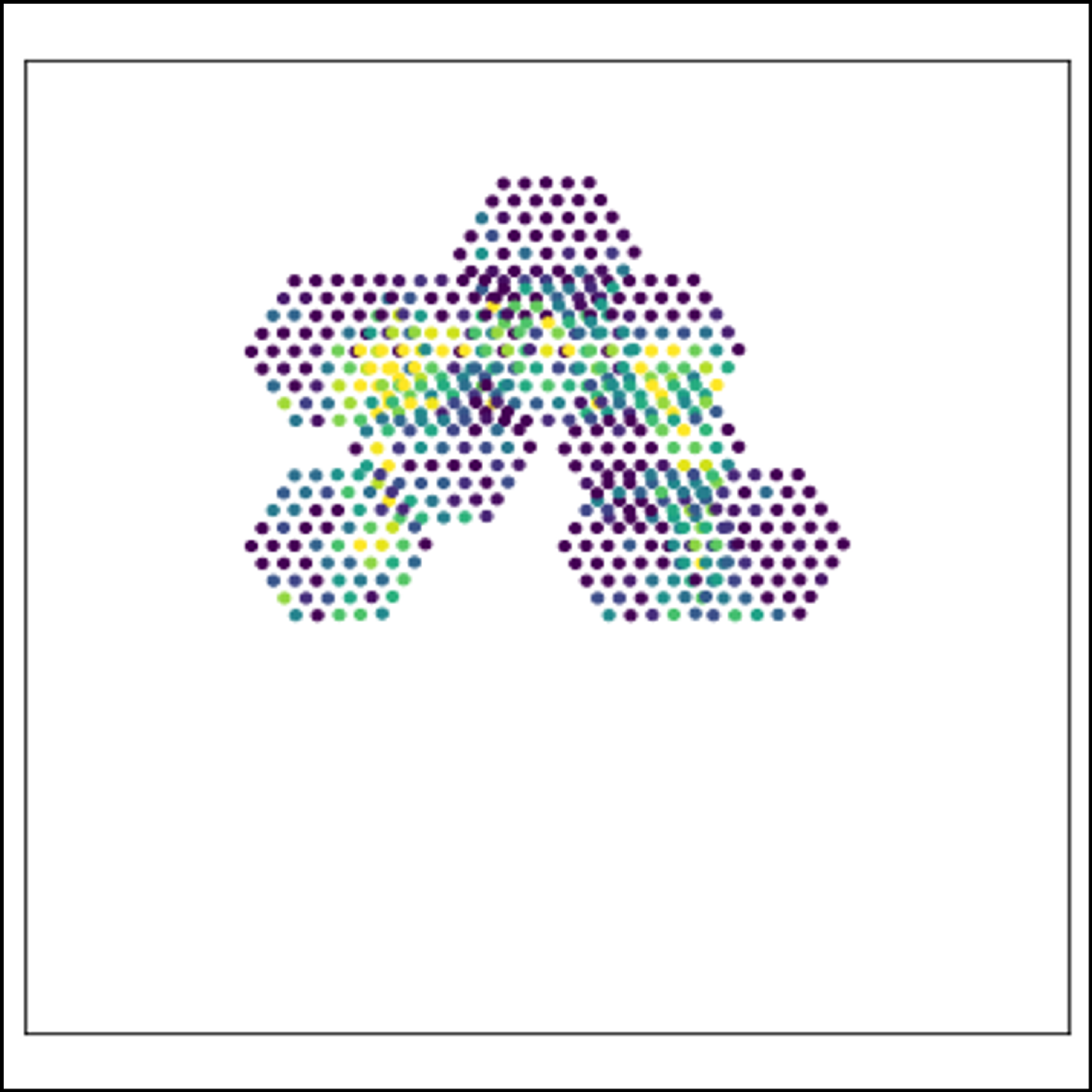}} &
		\frame{\includegraphics[width=0.13\linewidth,trim={45 40 35 40},clip]{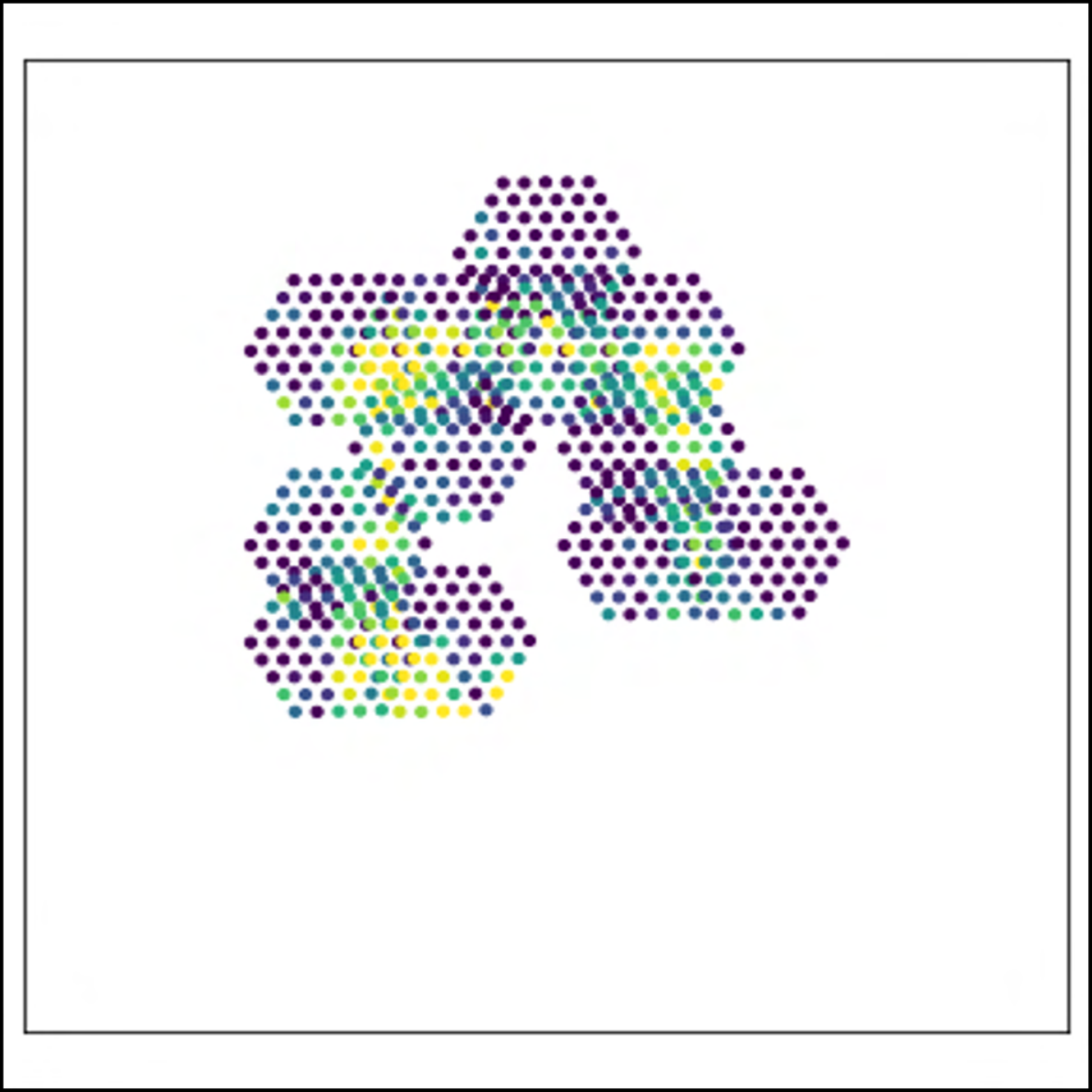}} &
		\frame{\includegraphics[width=0.13\linewidth,trim={45 40 35 40},clip]{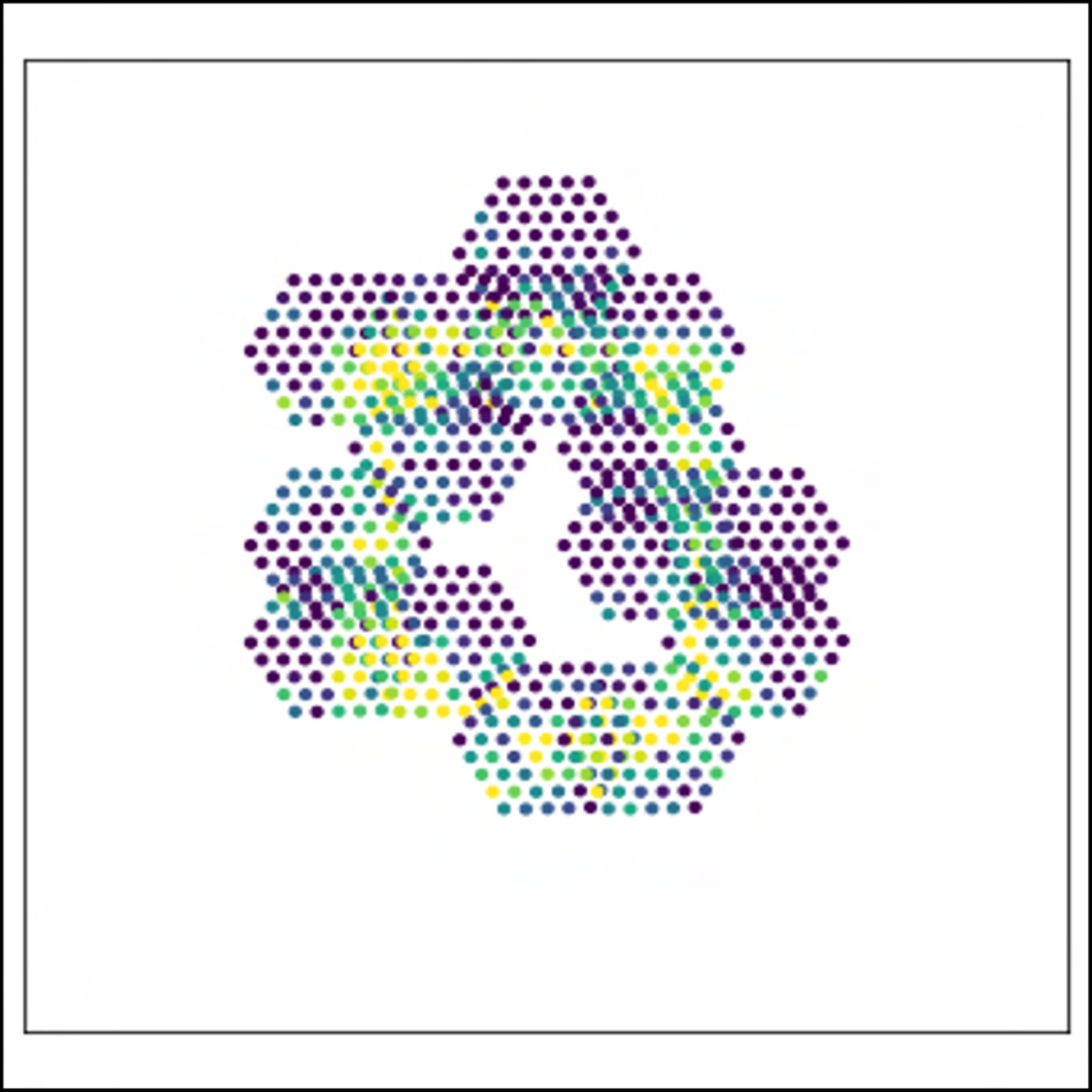}}
		\\
    	\frame{\includegraphics[width=0.13\linewidth,trim={45 40 35 40},clip]{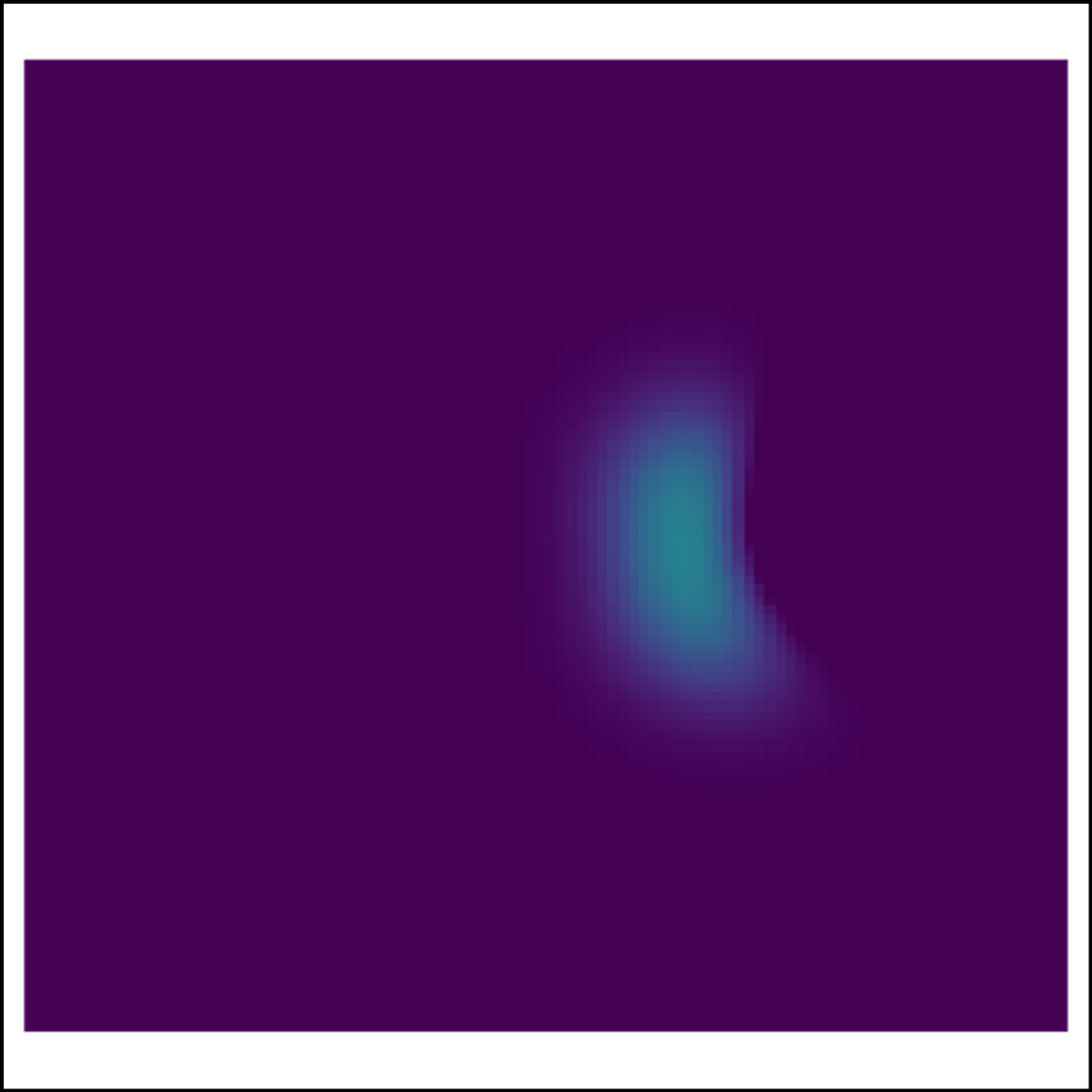}} &
		\frame{\includegraphics[width=0.13\linewidth,trim={45 40 35 40},clip]{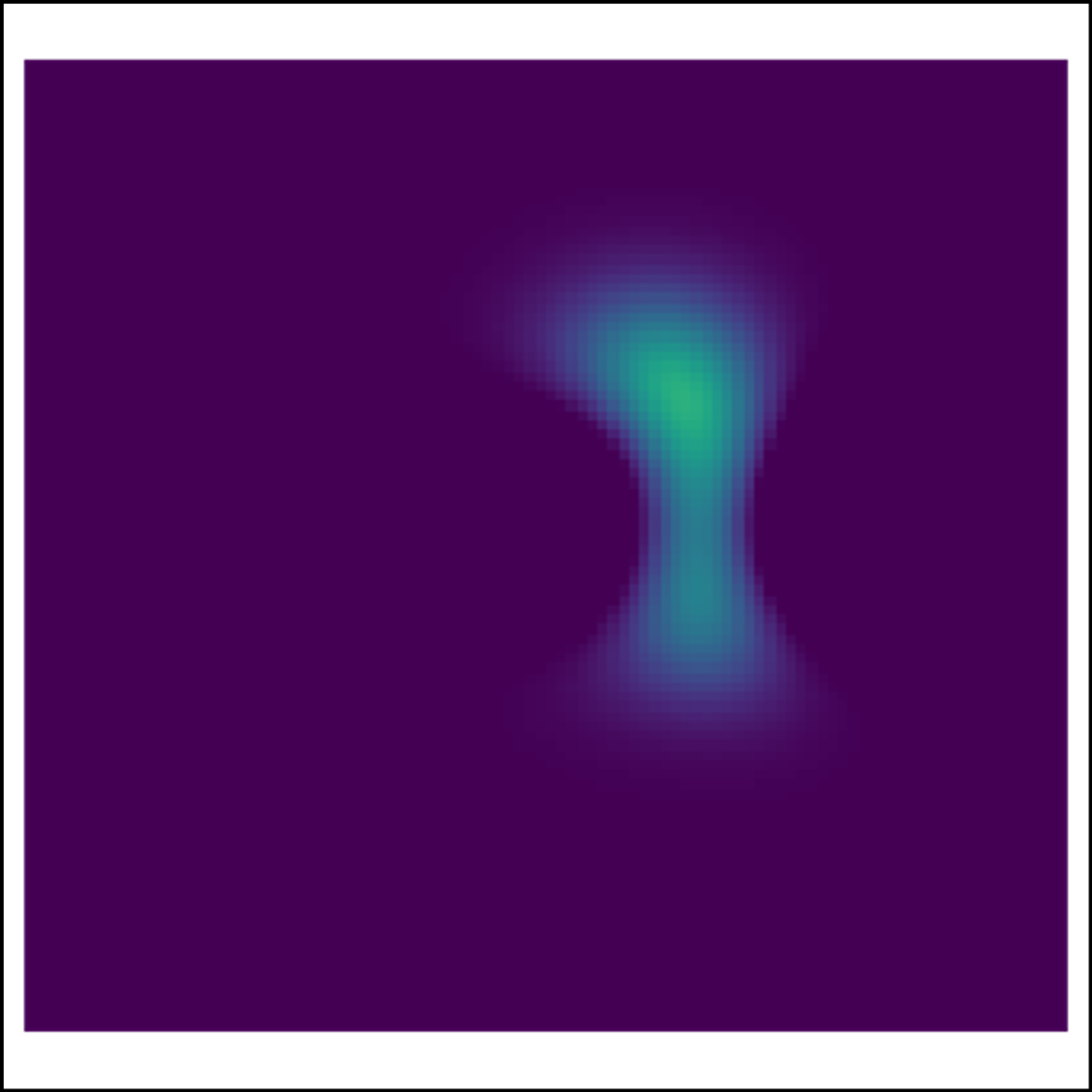}} &
		\frame{\includegraphics[width=0.13\linewidth,trim={45 40 35 40},clip]{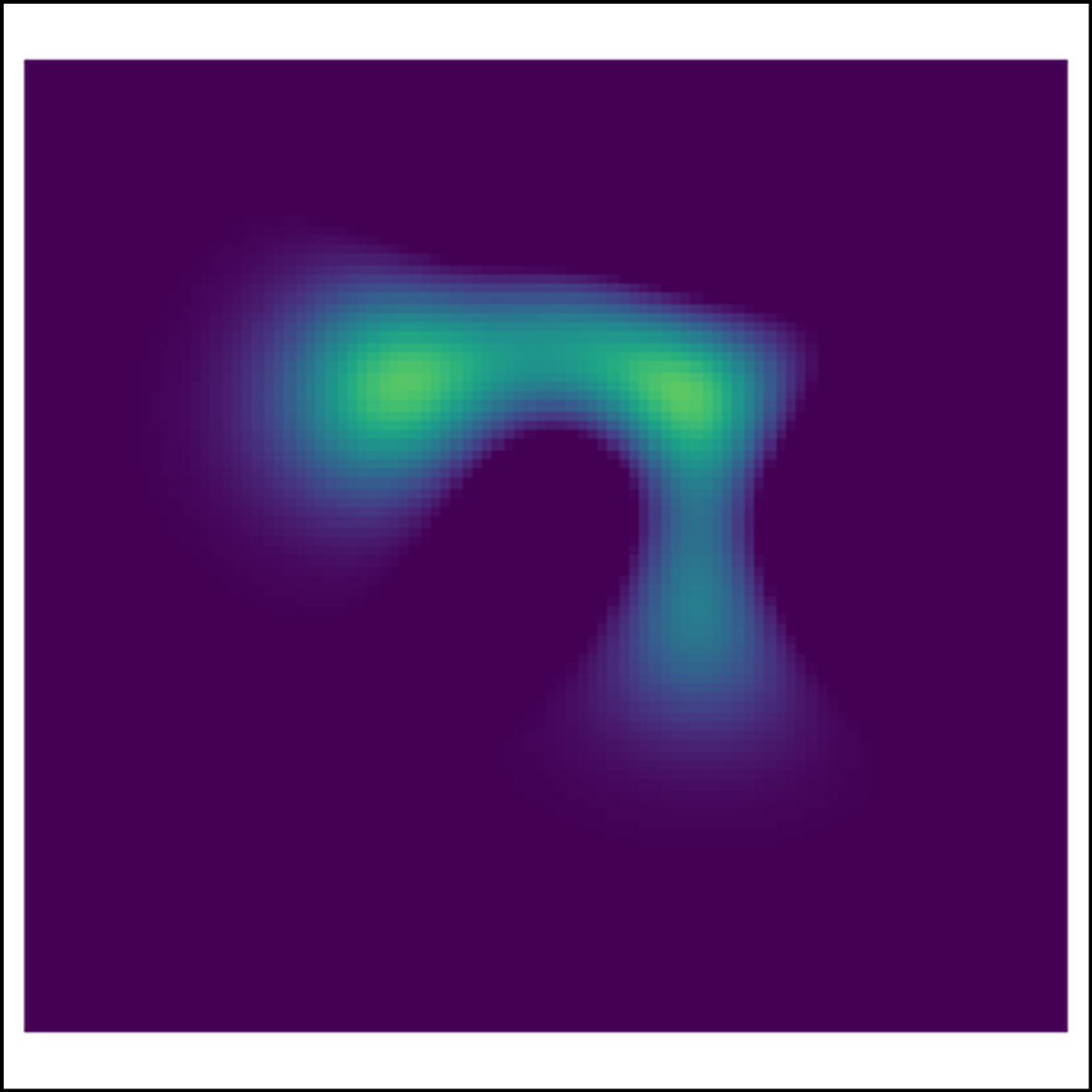}} &
		\frame{\includegraphics[width=0.13\linewidth,trim={45 40 35 40},clip]{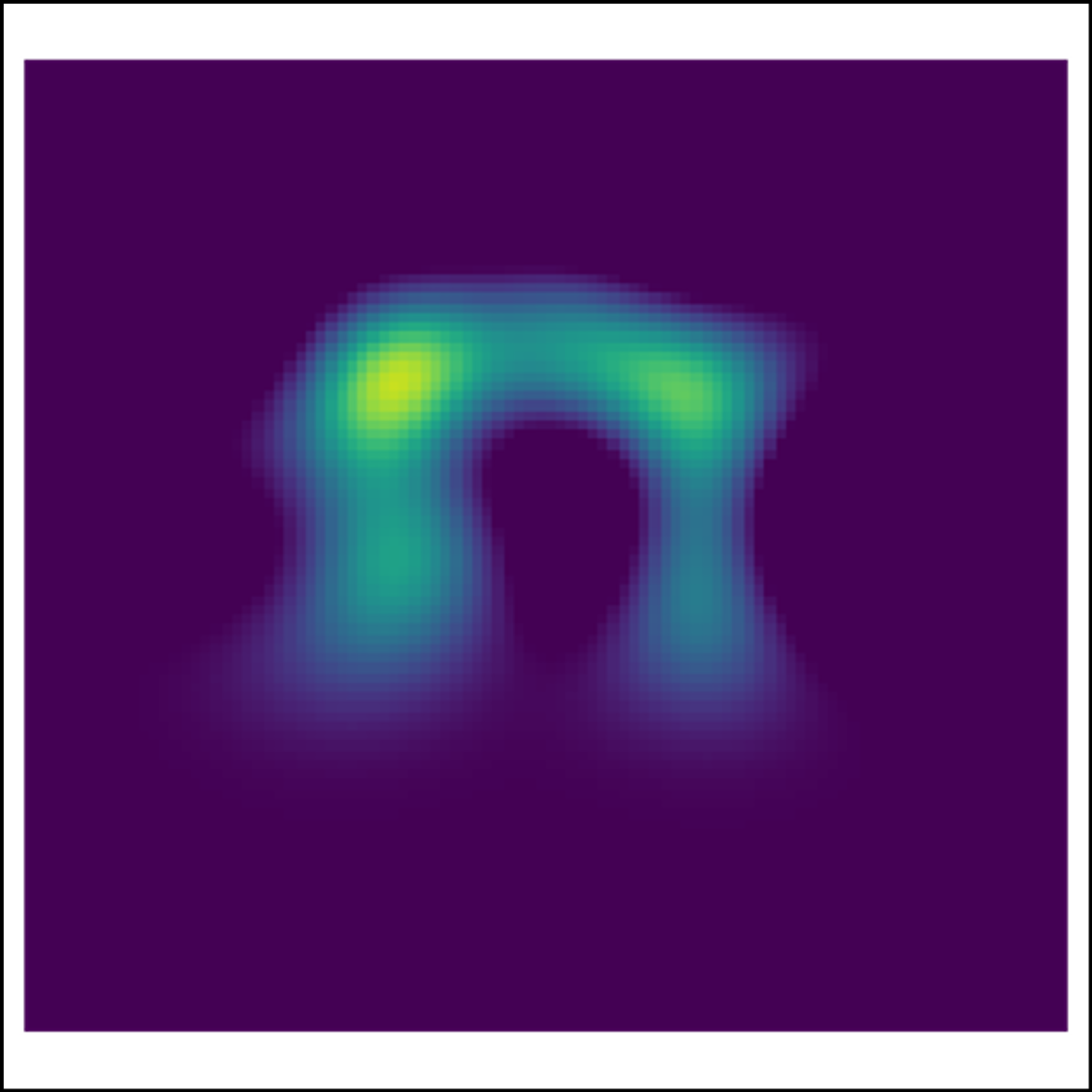}} &
		\frame{\includegraphics[width=0.13\linewidth,trim={45 40 35 40},clip]{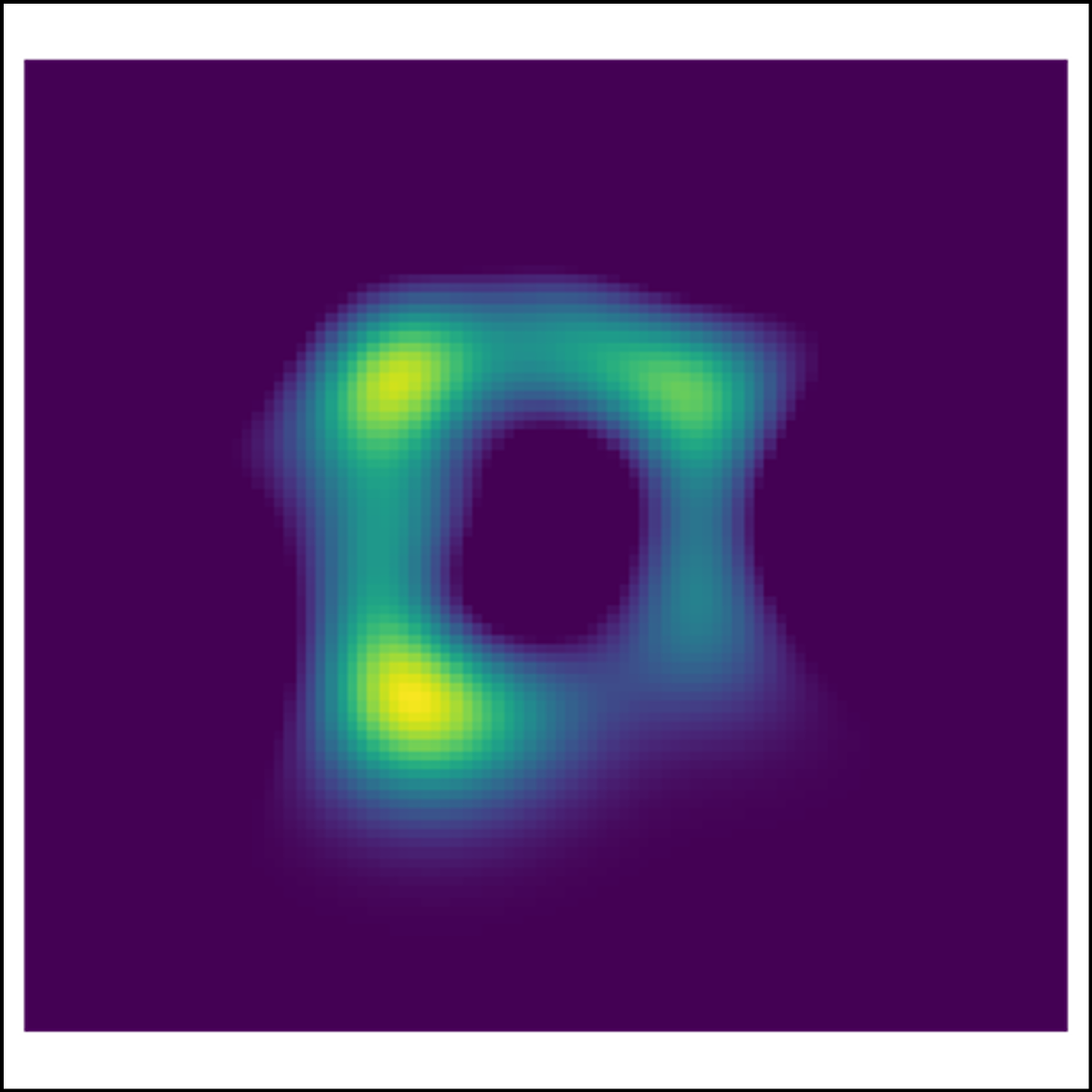}} &
		\frame{\includegraphics[width=0.13\linewidth,trim={45 40 35 40},clip]{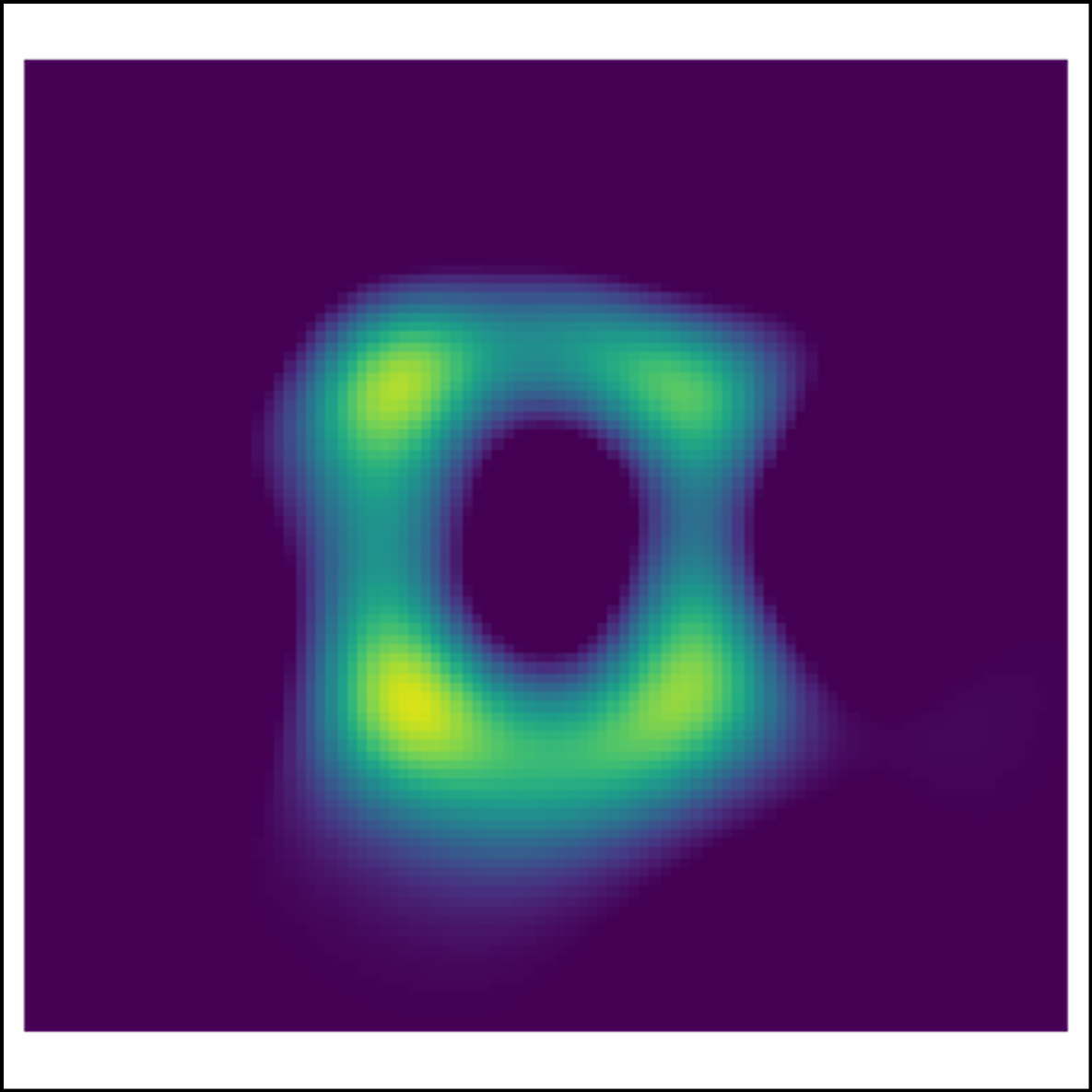}}
	\end{tabular}
	\caption{An example of the autonomous exploration of circular, triangle, and square mid-air haptic shapes, with a selection of steps taken by the system. The top rows show the data collected, and the bottom rows show the global mapping of the shapes at that current state. All plots cover a 60\,mm$\times$60\,mm square area.}
	\label{fig_results_exploration}
\end{figure*}

\subsection{Autonomous haptic exploration}

An initial testing of autonomous haptic exploration with our tactile robot was carried out on the haptic circle, triangle, and square, as regular shapes that typify the type of mid-air haptic exploration one would want with some variations.

The autonomous tactile robot was able to efficiently explore the shapes and produce detailed visualizations (Fig. \ref{fig_results_exploration}). With the perception method used in the algorithm, it is able to identify enough local information about the stimulus to determine the direction it should move to. This lets the system collect data only at the positions in space where there is stimulus. This method builds up a shape gradually as it explores the space. Fig. \ref{fig_results_exploration} shows this process, highlighting the sequence of motions the robot takes with the data it has gathered at each step, and the global perception of the shape it has built up with all of the data it has gathered so far. Overall, the tactile robot traces around the entire haptic shape and generates a representation of the stimulus with only 15 data samples for the circle and square and 13 data samples for the triangle, compared with 81 samples for the systematic mapping method.

This sampling procedure makes the autonomous exploration method more efficient than the first method for systematic mapping, in which every position in the grid space is sampled.
This algorithm does rely on some conditions for efficient haptic exploration, such as a high enough tactile signal which can be used to localize the stimulus and detect its orientation. It also requires the shapes to have closed, single continuous curves. A more robust exploration algorithm can be developed in the future which utilizes more sophisticated methods, as has been demonstrated with the tactile sensor for physical contour following \cite{leporaSoftBiomimeticOptical2021} which could overcome these limitations.
Overall, it shows the promise of real-time control as a component of the tactile robotic  system to map haptic shapes more efficiently and more like how humans would explore the shapes.

\section{Discussion}

This paper presented a tactile robotic system for evaluating virtual touch: combining a low-cost, accessible desktop robot arm with a 3D-printed biomimetic tactile sensor with a systematic haptic mapping procedure. This platform allowed us to successfully map mid-air haptic stimuli of various shapes. Additional control modes can be included within this platform, such as autonomous haptic exploration to efficiently sample only those regions of space where there may be an appreciable tactile stimulation. 

Various methods have been used to test and characterize mid-air haptics. Quantitative methods, such as using a Laser Doppler Vibrometer (LDV) to measure the deflection of materials caused by the acoustic radiation force of the mid-air haptic stimuli, or microphones to measure the sound pressure of the stimulus, can give detailed information of the sensations. However, both these methods are are limited to collecting data at single points with every measurement; a large number of data samples is needed to scan points in a two-dimensional plane to generate spatial data. With our method, just one single measurement of data can give us spatial information about the sensation. Since the sensor we use, the TacTip, captures an image of 127 marker-tipped pins, each marker gives information about the deflection of the sensor's skin at that point. 
%
Qualitative methods which have been used, such as projecting the sensations on an oil bath, lead to a faster identification of the shape of the stimulus than with our method, so they can have their advantage when a quick assessment of the shape is needed. However, as they are a qualitative method, they do not give the detailed quantification of the stimulus intensity provided by our method, or the capability to explore and interact with the stimulus more similarly to humans.

Since we use a robotic system, we can utilize different exploration and control methods to sample the data more intelligently, which we introduced with a method for autonomous haptic exploration. When humans interact with physical objects, they utilize contour following to move their hands along the most salient contact features of an object to determine the overall shape they are feeling. This has been identified as an exploratory procedure that can be used to determine various aspects such as shape and volume of handled objects \cite{ledermanHandMovementsWindow1987}. While it is not yet known which specific exploratory procedures are used by humans to interact with mid-air haptics, user studies have suggested that active touch could help people to distinguish between static mid-air haptic shapes more accurately \cite{hajasMidAirHapticRendering2020}.  This opens up a way to compare the manner in which robots sample data with how humans interact with objects via their sense of touch, to understand better the nature of human haptic perception and interaction.

{\em Acknowledgements:} We thank Andrew Stinchcombe for technical support and the rest of the Tactile Robotics group for their help.


\bibliographystyle{IEEEtran}
\bibliography{references}

\begin{thebibliography}{10}
\providecommand{\url}[1]{#1}
\csname url@samestyle\endcsname
\providecommand{\newblock}{\relax}
\providecommand{\bibinfo}[2]{#2}
\providecommand{\BIBentrySTDinterwordspacing}{\spaceskip=0pt\relax}
\providecommand{\BIBentryALTinterwordstretchfactor}{4}
\providecommand{\BIBentryALTinterwordspacing}{\spaceskip=\fontdimen2\font plus
\BIBentryALTinterwordstretchfactor\fontdimen3\font minus
  \fontdimen4\font\relax}
\providecommand{\BIBforeignlanguage}[2]{{%
\expandafter\ifx\csname l@#1\endcsname\relax
\typeout{** WARNING: IEEEtran.bst: No hyphenation pattern has been}%
\typeout{** loaded for the language `#1'. Using the pattern for}%
\typeout{** the default language instead.}%
\else
\language=\csname l@#1\endcsname
\fi
#2}}
\providecommand{\BIBdecl}{\relax}
\BIBdecl

\bibitem{carterUltraHapticsMultipointMidair2013}
T.~Carter, S.~A. Seah, B.~Long, B.~Drinkwater, and S.~Subramanian,
  ``{{UltraHaptics}}: Multi-point mid-air haptic feedback for touch surfaces,''
  in \emph{Proceedings of the 26th Annual ACM Symposium on User Interface
  Software and Technology (UIST)}, 2013, pp. 505--514.

\bibitem{rakkolainenSurveyMidAirUltrasound2021}
I.~Rakkolainen, E.~Freeman, A.~Sand, R.~Raisamo, and S.~Brewster, ``A
  {{Survey}} of {{Mid}}-{{Air Ultrasound Haptics}} and {{Its Applications}},''
  \emph{IEEE Trans. Haptics}, vol.~14, no.~1, pp. 2--19, Jan. 2021.

\bibitem{leporaSoftBiomimeticOptical2021}
N.~F. Lepora, ``Soft biomimetic optical tactile sensing with the tactip: A
  review,'' \emph{IEEE Sensors Journal}, vol.~21, no.~19, pp. 21\,131--21\,143,
  2021.

\bibitem{ward-cherrierTacTipFamilySoft2018}
B.~{Ward-Cherrier}, N.~Pestell, L.~Cramphorn, B.~Winstone, M.~E. Giannaccini,
  J.~Rossiter, and N.~F. Lepora, ``The {{TacTip Family}}: {{Soft Optical
  Tactile Sensors}} with {{3D}}-{{Printed Biomimetic Morphologies}},''
  \emph{Soft Robotics}, vol.~5, no.~2, pp. 216--227, Apr. 2018.

\bibitem{ledermanHandMovementsWindow1987}
S.~J. Lederman and R.~L. Klatzky, ``Hand movements: {{A}} window into haptic
  object recognition,'' \emph{Cognitive Psychology}, vol.~19, no.~3, pp.
  342--368, Jul. 1987.

\bibitem{frierUsingSpatiotemporalModulation2018}
W.~Frier, D.~Ablart, J.~Chilles, B.~Long, M.~Giordano, M.~Obrist, and
  S.~Subramanian, ``Using {{Spatiotemporal Modulation}} to {{Draw Tactile
  Patterns}} in {{Mid}}-{{Air}},'' in \emph{Proceedings of the International
  Conference on Human Haptic Sensing and Touch Enabled Computer Applications},
  2018, pp. 270--281.

\bibitem{chillesLaserDopplerVibrometry2019}
J.~Chilles, W.~Frier, A.~Abdouni, M.~Giordano, and O.~Georgiou, ``Laser
  {{Doppler Vibrometry}} and {{FEM Simulations}} of {{Ultrasonic Mid}}-{{Air
  Haptics}},'' in \emph{Proceedings of the {{IEEE World Haptics Conference}}
  ({{WHC}})}, Jul. 2019, pp. 259--264.

\bibitem{kappusSpatiotemporalModulationMidair2018}
B.~Kappus and B.~Long, ``Spatiotemporal modulation for mid-air haptic feedback
  from an ultrasonic phased array,'' \emph{The Journal of the Acoustical
  Society of America}, vol. 143, no.~3, p. 1836, Mar. 2018.

\bibitem{longRenderingVolumetricHaptic2014}
B.~Long, S.~A. Seah, T.~Carter, and S.~Subramanian, ``Rendering volumetric
  haptic shapes in mid-air using ultrasound,'' \emph{ACM Transactions on
  Graphics}, vol.~33, no.~6, pp. 1--10, Nov. 2014.

\bibitem{ruttenInvisibleTouchHow2019}
I.~Rutten, W.~Frier, L.~{Van den Bogaert}, and D.~Geerts, ``Invisible
  {{Touch}}: {{How Identifiable}} are {{Mid}}-{{Air Haptic Shapes}}?'' in
  \emph{Proceedings of the Extended Abstracts on Human Factors in Computing
  Systems}, 2019, pp. 1--6.

\bibitem{takahashiTactileStimulationRepetitive2020}
R.~Takahashi, K.~Hasegawa, and H.~Shinoda, ``Tactile {{Stimulation}} by
  {{Repetitive Lateral Movement}} of {{Midair Ultrasound Focus}},'' \emph{IEEE
  Transactions on Haptics}, vol.~13, no.~2, pp. 334--342, Apr. 2020.

\bibitem{sakiyamaEvaluationMultiPointDynamic2019}
E.~Sakiyama, D.~Matsumoto, M.~Fujiwara, Y.~Makino, and H.~Shinoda, ``Evaluation
  of {{Multi}}-{{Point Dynamic Pressure Reproduction Using Microphone}}-{{Based
  Tactile Sensor Array}},'' in \emph{Proceedings of the {{IEEE International
  Symposium}} on {{Haptic}}, {{Audio}} and {{Visual Environments}} and
  {{Games}} ({{HAVE}})}, Oct. 2019, pp. 1--6.

\bibitem{sakiyamaMidairTactileReproduction2020}
E.~Sakiyama, A.~Matsubayashi, D.~Matsumoto, M.~Fujiwara, Y.~Makino, and
  H.~Shinoda, ``Midair {{Tactile Reproduction}} of {{Real Objects}},'' in
  \emph{Proceedings of the International Conference on Human Haptic Sensing and
  Touch Enabled Computer Applications}, 2020, pp. 425--433.

\bibitem{martinez-hernandezActiveSensorimotorControl2017}
U.~{Martinez-Hernandez}, T.~J. Dodd, M.~H. Evans, T.~J. Prescott, and N.~F.
  Lepora, ``Active sensorimotor control for tactile exploration,''
  \emph{Robotics and Autonomous Systems}, vol.~87, pp. 15--27, Jan. 2017.

\bibitem{leporaExploratoryTactileServoing2017}
N.~F. Lepora, K.~Aquilina, and L.~Cramphorn, ``Exploratory {{Tactile Servoing
  With Active Touch}},'' \emph{IEEE Robotics and Automation Letters}, vol.~2,
  no.~2, pp. 1156--1163, Apr. 2017.

\bibitem{leporaPixelsPerceptsHighly2019}
N.~F. Lepora, A.~Church, C.~{de Kerckhove}, R.~Hadsell, and J.~Lloyd, ``From
  {{Pixels}} to {{Percepts}}: {{Highly Robust Edge Perception}} and {{Contour
  Following Using Deep Learning}} and an {{Optical Biomimetic Tactile
  Sensor}},'' \emph{IEEE Robotics and Automation Letters}, vol.~4, no.~2, pp.
  2101--2107, Apr. 2019.

\bibitem{liControlFrameworkTactile2013}
Q.~Li, C.~Sch{\"u}rmann, R.~Haschke, and H.~J. Ritter, ``A control framework
  for tactile servoing.'' \emph{Robotics: Science and systems}, 2013.

\bibitem{kappassovTouchDrivenController2020}
Z.~Kappassov, J.-A. Corrales, and V.~Perdereau, ``Touch driven controller and
  tactile features for physical interactions,'' \emph{Robotics and Autonomous
  Systems}, vol. 123, p. 103332, Jan. 2020.

\bibitem{pestell2022a}
N.~Pestell and N.~F. Lepora, ``Artificial sa-i, ra-i and ra-ii/vibrotactile
  afferents for tactile sensing of texture,'' \emph{Journal of The Royal
  Society Interface}, vol.~19, no. 189, 2022.

\bibitem{pestell2022b}
N.~Pestell, T.~Griffith, and N.~F. Lepora, ``Artificial sa-i and ra-i afferents
  for tactile sensing of ridges and gratings,'' \emph{Journal of The Royal
  Society Interface}, vol.~19, no. 189, 2022.

\bibitem{frierSimulatingAirborneUltrasound2022}
W.~Frier, A.~Abdouni, D.~Pittera, O.~Georgiou, and R.~Malkin, ``Simulating
  airborne ultrasound vibrations in human skin for haptic applications,''
  \emph{IEEE Access}, vol.~10, pp. 15\,443--15\,456, 2022.

\bibitem{alakhawandSensingUltrasonicMidAir2020}
N.~Alakhawand, W.~Frier, K.~M. Freud, O.~Georgiou, and N.~F. Lepora, ``Sensing
  {{Ultrasonic Mid}}-{{Air Haptics}} with~a {{Biomimetic Tactile Fingertip}},''
  in \emph{Proceedings of the International Conference on Human Haptic Sensing
  and Touch Enabled Computer Applications}, 2020, pp. 362--370.

\bibitem{cramphornVoronoiFeaturesTactile2018}
L.~Cramphorn, J.~Lloyd, and N.~F. Lepora, ``Voronoi {{Features}} for {{Tactile
  Sensing}}: {{Direct Inference}} of {{Pressure}}, {{Shear}}, and {{Contact
  Locations}},'' in \emph{Proceedings of the {{IEEE International Conference}}
  on {{Robotics}} and {{Automation}} ({{ICRA}})}, May 2018, pp. 2752--2757.

\bibitem{hajasMidAirHapticRendering2020}
D.~Hajas, D.~Pittera, A.~Nasce, O.~Georgiou, and M.~Obrist, ``Mid-{{Air Haptic
  Rendering}} of {{2D Geometric Shapes With}} a {{Dynamic Tactile Pointer}},''
  \emph{IEEE Transactions on Haptics}, vol.~13, no.~4, pp. 806--817, Oct. 2020.

\end{thebibliography}

\end{document}